\documentclass[10pt,twocolumn,letterpaper]{article}

\usepackage[pagenumbers]{iccv} 

\newcommand{\bm}{\mathbf{m}}

\makeatletter
\DeclareRobustCommand\onedot{\futurelet\@let@token\@onedot}
\def\@onedot{\ifx\@let@token.\else.\null\fi\xspace}

\makeatother

\newcommand{\xdownarrow}[1]{%
  {\left\downarrow\vbox to #1{}\right.\kern-\nulldelimiterspace}
}

\newcommand{\xuparrow}[1]{%
  {\left\uparrow\vbox to #1{}\right.\kern-\nulldelimiterspace}
}

\newcommand{\boldparagraph}[1]{\vspace{0.15cm}\noindent{\bf #1.} }

\definecolor{First}{HTML}{BDE6CD}
\definecolor{Second}{HTML}{E2EEBC}
\definecolor{Third}{HTML}{FFF8C5}

\newcommand{\fst}[1]{\cellcolor{First}#1}
\newcommand{\snd}[1]{\cellcolor{Second}#1}
\newcommand{\trd}[1]{\cellcolor{Third}#1}

\usepackage[dvipsnames]{xcolor}
\usepackage{colortbl}
\usepackage{booktabs}
\usepackage{multirow}
\usepackage{bm}
\usepackage{overpic}
\usepackage[export]{adjustbox}

\definecolor{iccvblue}{rgb}{0.21,0.49,0.74}
\usepackage[pagebackref,breaklinks,colorlinks,allcolors=iccvblue]{hyperref}

\def\paperID{AAAA}
\def\confName{AAAA}
\def\confYear{AAAA}

\title{WarpRF: Multi-View Consistency 
for Training-Free Uncertainty Quantification and Applications in Radiance Fields}

\author{Sadra Safadoust\textsuperscript{1} \quad Fabio Tosi\textsuperscript{2} \quad Fatma Güney\textsuperscript{1} \quad Matteo Poggi\textsuperscript{2}\\
\textsuperscript{1}Koç University \quad \textsuperscript{2}University of Bologna \\
\textbf{\url{https://kuis-ai.github.io/WarpRF/}}
}

\begin{document}

\twocolumn[{
\renewcommand\twocolumn[1][]{#1}
\maketitle
\begin{center} 
    \vspace{-4mm}
    \centering
    \begin{overpic}[width=1.0\linewidth]{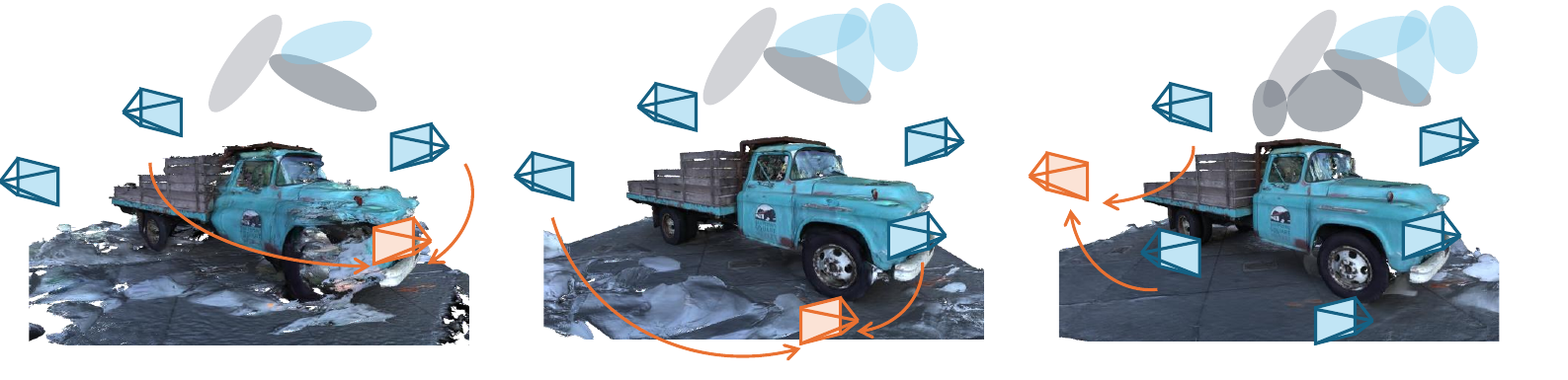} 
    \put (8,20) {\small\textcolor{black}{\textbf{3DGS$_{t0}$}}}
    \put (38,20) {\small\textcolor{black}{\textbf{3DGS$_{t1}$}}}
    \put (74,20) {\small\textcolor{black}{\textbf{3DGS$_{t2}$}}}
    \put (27,6) {\small\textcolor{orange}{\textbf{$\mathcal{U}_0$}}}
    \put (54,1) {\small\textcolor{orange}{\textbf{$\mathcal{U}_1$}}}
    \put (64,11) {\small\textcolor{orange}{\textbf{$\mathcal{U}_2$}}}
    \put (98,1) {\small\textcolor{black}{\textbf{...}}}
    \put (98,20) {\small\textcolor{black}{\textbf{...}}}
    \end{overpic}\vspace{-0.7cm}
    \
    \captionof{figure}{\textbf{Active mapping through WarpRF uncertainty quantification.} Given a radiance field-based surface reconstruction framework trained on an initial set of images -- e.g. the 3D Gaussian Splatting 3DGS$_{t0}$ (left) trained on the blue viewpoints -- WarpRF estimates the best next view (in orange in the figure) by quantifying the rendering uncertainty $\mathcal{U}_0$ associated to it through warping. This is added to the training set and used to fit a more accurate 3DGS$_{t1}$ (center), from where we can identify a new next best view with maximum uncertainty $\mathcal{U}_1$, add it to the training set and obtain 3DGS$_{t2}$ (right), and so on and so forth.  } 
    \label{fig:teaser} 
\end{center}
}
]

\newcommand{\sadra}[1]{ \noindent {\color{blue} {#1}} } 

\begin{abstract}

We introduce WarpRF, a training-free general-purpose framework for quantifying the uncertainty of radiance fields. 
Built upon the assumption that photometric and geometric consistency should hold among images rendered by an accurate model, WarpRF quantifies its underlying uncertainty from an unseen point of view by leveraging backward warping across viewpoints, projecting reliable renderings to the unseen viewpoint and measuring the consistency with images rendered there. 
WarpRF is simple and inexpensive, does not require any training, and can be applied to any radiance field implementation for free. WarpRF excels at both uncertainty quantification and downstream tasks, e.g., active view selection and active mapping, outperforming any existing method tailored to specific frameworks. 

\end{abstract}    
\section{Introduction}
\label{sec:intro}

The advent of Neural Radiance Fields (NeRF) \cite{mildenhall2020nerf} reshaped image rendering and novel view synthesis in the last five years, as well as several other applications of multi-view geometry and computer vision. By casting a 3D scene into a \textit{radiance field}, as an implicit representation embedded within the weights of a multi-layer perceptron, this family of \textit{network-based} models allow for rendering new arbitrary views of the scene with unprecedented realism. More recently, a new family of \textit{point-based} approaches gained attention, by neglecting the use of neural networks: among them, 3D Gaussian Splatting (3DGS) \cite{kerbl3Dgaussians} is by far the most popular, thanks to its fast rasterizer and high rendering quality. 
As both can encode the geometry of the scene very accurately, the two frameworks have been revised for dense 3D reconstruction, or \textit{mapping} \cite{neus,sugar}. 
Although both have their pros and cons, both require a significant number of images to achieve either optimal view synthesis or mapping, ranging from a few hundred for object-centric scenes to several thousand for large-scale and outdoor environments. If the training set does not meet this requirement, the quality of the final results severely drops. 

Among the latest advances in this field, uncertainty quantification has recently \cite{gawlikowski2023survey,kendall2017uncertainties} attracted the attention of the radiance fields community, on the one hand, as a tool to identify the inability of a trained NeRF or 3DGS to render novel images of satisfactory quality.
On the other hand, the ability to reliably measure the model uncertainty when rendering an image from a particular viewpoint may also serve as a criterion for \textit{active view selection} or \textit{active mapping} --. i.e., identifying the unseen views that are most informative for the model to add to the training set and improve its internal scene representation. From a practical perspective, this means finding the best next viewpoint to move the camera to capture additional images and maximize scene coverage.

Different strategies have been proposed to quantify the uncertainty in radiance fields \cite{lee2022uncertainty,yan2023activeIO,zhan2022activermap,CF-NeRF,shen2021snerf,sünderhauf2022densityaware,goli2023,lyu2024manifold,Jiang2024FisherRF}, with different degrees of complexity and generality: some approaches require modifying the internal structure of the radiance fields and performing ad hoc optimization during training, such as implementing Bayesian models \cite{CF-NeRF,shen2021snerf} or custom perturbation fields \cite{goli2023}; others allow seamless uncertainty estimation from pre-trained models, such as from the distribution of densities predicted along rays during the NeRF rendering process \cite{lee2022uncertainty}.
Nonetheless, all of the solutions developed so far are customized to a radiance field implementation or family, most of them being suited to NeRFs \cite{lee2022uncertainty,yan2023activeIO,zhan2022activermap,CF-NeRF,shen2021snerf,sünderhauf2022densityaware,goli2023} and some of the latest specifically targeting 3DGS \cite{lyu2024manifold,Jiang2024FisherRF}.

We argue that none of the aforementioned approaches take into account the simplest cue available to measure how faithfully a radiance field can approximate the observed scene: \textit{multi-view consistency}. Given a source camera and one or more target cameras, this is measured by projecting either images or depth maps from the target viewpoints onto the source, depending on the depth map associated with the latter. The smaller the difference between the source and projected images or depth maps, the higher the multi-view consistency. Over the years, this has been a popular cue to measure the confidence of stereo matching algorithms \cite{poggi2022confidence}, to filter out noisy depth points during multi-view reconstruction \cite{furukawa2009accurate}, or even for self-supervised learning of single-image depth estimation \cite{monodepth2}.

In this paper, we unveil the surprising effectiveness of multi-view consistency in quantifying the uncertainty of modern radiance fields, and present \textbf{WarpRF}. By treating the depth maps rendered by a radiance field as a proxy for the geometry it encodes, we use it to reproject either rendered images or depth maps from training viewpoints and measure multi-view consistency. This solution is simple, yet effective and general -- as it can be seamlessly implemented on top of any radiance field without having access to its internal structure. Extensive experimental results show how our approach outperforms existing solutions in several downstream tasks that exploit uncertainty, such as active view selection and active mapping, the latter of which is shown in Fig. \ref{fig:teaser}.

Our main contributions can be summarized as follows: 

\begin{itemize}
    \item We introduce WarpRF, a novel framework for inexpensive uncertainty quantification in radiance field that exploits multi-view consistency for the first time for this purpose.
    \item WarpRF is both \textit{training-free} and \textit{general}, making it applicable to any radiance field model, pre-trained or not.
    \item Compared to existing frameworks for uncertainty quantification \cite{Jiang2024FisherRF}, WarpRF demonstrates consistently superior performance across several applications.
    \item Given its generality, WarpRF is the first method enabling active view selection with the cutting-edge SVRaster framework \cite{Sun2024SVR}.
\end{itemize}
Our code will be made publicly available.

\section{Related Work}
\label{sec:rw}

We briefly review the literature concerning uncertainty quantification in radiance fields and active view selection.

\textbf{Uncertainty in Neural Fields.} With the latest advances in novel view synthesis \cite{mildenhall2020nerf,kerbl3Dgaussians}, quantifying the uncertainty related to the rendering process has acquired increasing attention over the years. Starting with network-based frameworks such as NeRF, the first attempts \cite{lee2022uncertainty,yan2023activeIO,zhan2022activermap} focused on deriving uncertainty from the distribution of densities predicted along rays passing through each pixel. 
More recent approaches \cite{CF-NeRF,shen2021snerf} recast NeRF as a Bayesian model to extract predictive uncertainty during rendering, use ray termination probability \cite{sünderhauf2022densityaware} to measure the uncertainty in unobserved parts of the scene, or introduce a perturbation field \cite{goli2023} from which uncertainty is queried and rendered through ray casting as done for color.
Recent works \cite{klasson2024sources} provide a taxonomy of different uncertainty sources in 3D reconstruction, while others address specific challenges such as occluded regions and visibility \cite{shen2024estimating,xue2024nvf},  handling dynamic distractors in in-the-wild captures \cite{ren2024nerf}.
Lately, following the increasing popularity of 3DGS, uncertainty has been modeled in a differentiable manner as a low-dimensional manifold in the space of the 3DGS parameters \cite{lyu2024manifold}, integrated with 4D Gaussian Splatting for dynamic scenes \cite{kim20254d}, and extended to SLAM applications \cite{hu2024cg}.
FisherRF \cite{Jiang2024FisherRF} introduces a general approach for uncertainty quantification in point-based rendering models such as 3DGS or plenoxels, based on Fisher information.
In this work, we further pursue generality and prove how a simple strategy based on multi-view consistency is more than enough for effective uncertainty quantification with both network-based (NeRF) and point-based (3DGS) approaches.

\textbf{Active View Selection and Mapping.} One of the practical applications enabled by uncertainty is the ability to actively select the images to be used to train either NeRF or 3DGs. ActiveNeRF \cite{pan2022activenerf} is the pioneer in this context, predicting mean and variance as outputs of vanilla NeRF and using the latter as uncertainty to select the next best view during training for the final quality of novel view synthesis.
NeurAR \cite{Ran2023neurar} follows a similar path to plan the optimal trajectory to perform object-centric 3D reconstructions, while Yan et al. \cite{yan2023active-neural-mapping} approximate uncertainty from the output of neighboring points and FisherRF \cite{Jiang2024FisherRF} applies its own uncertainty quantified from Fisher information.
Beyond the scope of our work, SCONE and MARACONS \cite{guedon2022scone,guedon2023macarons} tackled active view selection for 3D reconstruction with feed-forward networks.
We will prove that our simple yet effective approach to quantifying uncertainty also allows for optimal active view selection for both the novel view synthesis and 3D reconstruction tasks.

\section{Methodology}
\label{sec:method}
In this section, we describe how multi-view consistency, i.e. the projection of either depth or images from training views to target views, can be used as an effective method for measuring uncertainty in radiance fields.

\subsection{Background}

First, we introduce the fundamentals concerning the two most popular frameworks designed for modeling radiance fields, respectively, NeRF and 3DGS.

\textbf {Neural Radiance Field (NeRF).} Formally, NeRF is expressed as $f(\mathbf{x}, \mathbf{d}) \rightarrow (\mathbf{c}, \sigma)$, where a Multi-Layer Perceptron (MLP) with weights $\mathit{\Theta}$, denoted as $f_\mathit{\Theta}$, approximates a 5D function mapping 3D coordinates $\mathbf{x} = (x,y,z)$ and viewing direction $\mathbf{d} = (\theta,\phi)$ to color $\mathbf{c} = (r,g,b)$ and volume density $\sigma$. The representation ensures multi-view consistency by making density dependent only on position, while color depends on both position and viewing direction.

For novel view synthesis, NeRF casts camera rays through the scene and employs volume rendering techniques. The color $\mathcal{C}(\mathbf{r})$ along a camera ray $\mathbf{r}(t) = \mathbf{o} + t\mathbf{d}$ is computed as:

\begin{equation}
\mathcal{C}(\mathbf{r}) = \sum_{i=1}^{N} \alpha_i T_i\mathbf{c}_i, \quad T_i = \exp\left(-\sum_{j=1}^{i-1}\sigma_j\delta_j\right)
\end{equation}

where $\delta_i$ is the distance between adjacent samples, $\sigma_i$ and $\mathbf{c}_i$ are the density and color at sample point $i$, and $\alpha_i = (1-\exp(-\sigma_i\delta_i))$ represents the opacity from alpha compositing. $T_i$ is the accumulated transmittance along the ray up to point $i$. Additionally, NeRF provides depth estimation along rays using:

\begin{equation}
{\mathcal{D}}(\mathbf{r}) = \sum_{i=1}^{N} \alpha_i t_i T_i
\end{equation}

The entire model is optimized using a simple photometric loss between rendered and ground truth pixels:
\begin{equation}
\mathcal{L} = \sum_{r \in R} \| \hat{C}(\mathbf{r}) - C_{gt}(\mathbf{r}) \|^2_2
\end{equation}

where $C_{gt}(\mathbf{r})$ is the ground truth color for the pixel corresponding to ray $\mathbf{r}$, and $R$ represents the batch of rays used during training. According to this formulation, NeRF can be easily extended to deal with surface reconstruction \cite{neus}, by predicting a Signed Distance Function (SDF) through the MLP as $f(\mathbf{r}(t))$ and replacing $\alpha$ with $\rho(t)$, derived from the SDF as follows:

\begin{equation}
    \rho(t) = \max \big( \frac{-\frac{d\Phi}{dt}(f(\mathbf{r}(t)))}{\Phi(f(\mathbf{r}(t)))}, 0 \big)
\end{equation}
with $\Phi$ being the sigmoid function and $\frac{d\Phi}{dt}$ its derivative. 

\textbf {3D Gaussian Splatting (3DGS).} 3D Gaussian Splatting~\cite{kerbl3Dgaussians} represents a paradigm shift in novel view synthesis as an explicit radiance field technique. Unlike NeRF's implicit neural representation, 3DGS models scenes through differentiable 3D Gaussian primitives, enabling real-time rendering via a tile-based rasterizer.

Formally, 3DGS learns a set $\mathcal{G} = \{g_1, g_2, \dots, g_N\}$ of 3D Gaussians from multi-view images with known camera poses. Each Gaussian $g_i$ is characterized by a center position $\bm{\mu}_i \in \mathbb{R}^3$, a covariance matrix $\mathbf{\Sigma}_i \in \mathbb{R}^{3 \times 3}$, an opacity $o_i \in [0,1]$, and a view-dependent color $\mathbf{c}_i$ represented using spherical harmonics. The spatial influence of each Gaussian is defined as:

\begin{equation}
    g_i(\mathbf{x}) = e^{-\frac{1}{2} (\mathbf{x}-\bm{\mu}_i)^\top \mathbf{\Sigma}_i^{-1} (\mathbf{x}-\bm{\mu}_i)}
\end{equation}

For rendering, 3DGS projects these 3D Gaussians onto a 2D image plane (``splatting''), transforming the 3D covariance to 2D as $\mathbf{\Sigma}' = \mathbf{JW \Sigma W^T J^T}$ and the center as $\bm{\mu}' = \mathbf{JW}\bm{\mu}$. The final pixel color $\mathcal{C}$ is computed by alpha-blending overlapping Gaussians:

\begin{equation}
    \mathcal{C} = \sum_{i \in \mathcal{N}} \mathbf{c}_i \alpha_i \prod_{j=1}^{i-1} (1 - \alpha_j)
\end{equation}

where the opacity $\alpha_i$ combines the learned opacity with the projected Gaussian. Similarly, depth is rendered as:

\begin{equation}
    \mathcal{D} = \sum_{i \in \mathcal{N}} d_i \alpha_i \prod_{j=1}^{i-1} (1 - \alpha_j)
\end{equation}

with $d_i$ representing the depth of the $i$-th Gaussian's center. 3DGS is optimized through SGD using a combination of L1 and D-SSIM losses, with adaptive densification periodically adjusting the representation by adding points in regions with significant gradients and removing low-opacity points. Although some attempts to extract dense meshes from 3DGS exist \cite{sugar}, so far most approaches \cite{chen2024pgsr} reconstruct surfaces by rendering depth maps and fusing them through TSDF-fusion.

\begin{figure*}[t]
    \centering
    \begin{overpic}[clip,trim=0cm 1cm 20cm 0cm,width=\linewidth]{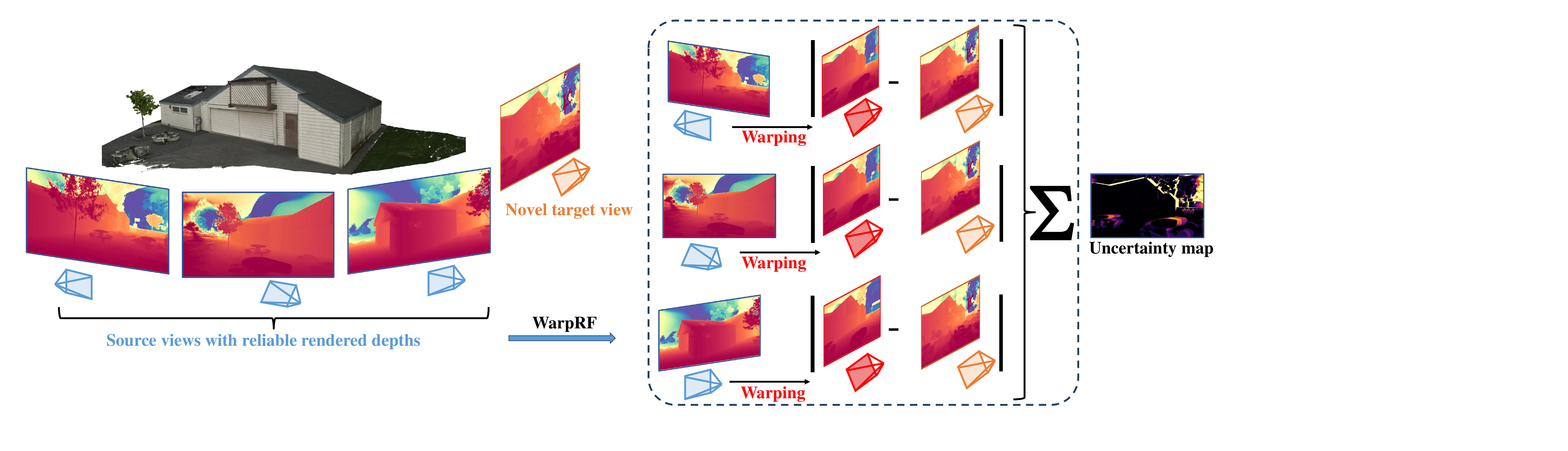}
    \put (15,5) {\small\textcolor{RoyalBlue}{
    $\{\mathcal{S}_0, \mathcal{S}_1, \mathcal{S}_2, \ldots, \mathcal{S}_S\}$}
    }
    \put (46,16) {\small\textcolor{orange}{$\mathcal{T}$}}
    \put (93,13) {\small\textcolor{black}{$\mathcal{U}_\mathcal{T}$}}
    \end{overpic}
    \vspace{-0.8cm}
    \caption{\textbf{Uncertainty quantification with WarpRF.} Given an initial set of source viewpoints 
    $\mathcal{S}=\{\mathcal{S}_0, \mathcal{S}_1, \mathcal{S}_2, \ldots, \mathcal{S}_S\}$ available for training, a radiance field is optimized over images $\{\mathcal{I}_{\mathcal{S}_0}, \mathcal{I}_{\mathcal{S}_1}, \mathcal{I}_{\mathcal{S}_2}, \ldots, \mathcal{I}_{\mathcal{S}_S}\}$. Then, upon selecting a novel target viewpoint $\mathcal{T}$, uncertainty can be quantified by rendering depth $\mathcal{D}_\mathcal{T}$ and use it to project either images $\mathcal{I}_{\mathcal{S}_n}$ or depth maps $\mathcal{D}_{\mathcal{S}_n}$ -- in this figure, we show an example of the latter -- into ${\mathcal{I}}_{\mathcal{S}_n \rightarrow \mathcal{T}}$ or ${\mathcal{D}}_{\mathcal{S}_n \rightarrow \mathcal{T}}$, according to their known camera poses. Then, uncertainty $\mathcal{U}_\mathcal{T}$ is measured by computing the per-pixel absolute difference between projected images/depths and those rendered from $\mathcal{T}$, averaged in case of multiple source views. }
    \label{fig:warprf}
\end{figure*}

\subsection{WarpRF}

We now introduce WarpRF, our framework for quantifying the uncertainty of radiance fields, as shown in Fig. \ref{fig:warprf}. At the core of our approach is the fact that any radiance fields -- whether NeRF, 3DGS or others -- are optimized based on multi-view geometry principles. As such, basic assumptions of photometric and geometric consistency should hold across images rendered by an accurate radiance field model -- with exceptions in cases of sudden illumination changes or moving objects, which are known challenges for these frameworks. We therefore propose that by measuring these properties and, more generally, the multi-view consistency exhibited by a radiance field model, we can obtain a reliable means of quantifying its underlying uncertainty in specific regions of the scene.
Depending on the application for which the neural field is deployed, we may be interested in quantifying either the uncertainty related to the rendering capabilities of the model or the uncertainty affecting the 3D geometry encoded within the neural field itself, which is used for surface reconstruction and mesh generation.
In both cases, the accuracy of the depth maps rendered by the model serves as a vital proxy for the quality of the underlying scene structure learned by the radiance field. Given evidence from the literature \cite{kangle2021dsnerf,roessle2022depthpriorsnerf,safadoust2024BMVC} that the fidelity of scene structure is closely related to rendering quality, we build our uncertainty quantification framework upon rendered depth and multi-view consistency. Specifically, we assume that both images and depth maps are accurately rendered in correspondence of the training viewpoints; therefore, we measure the consistency between these and the images and depth maps rendered from novel, unseen viewpoints. As such, in case of high unconsistency between the two viewpoints, we can assume that rendered images and depth maps from the unseen viewpoints are uncertain, as they are not as accurate as those obtained from the images used to train the radiance field itself.

\subsubsection{Pixel-wise Uncertainty}
\label{sec:unc_pixel}
Let $\mathcal{M}$ be any radiance field that, given a generic target viewpoint $\mathcal{T}$, can render the corresponding image $\mathcal{I}_\mathcal{T}$ and depth map $\mathcal{D}_\mathcal{T}$ at resolution H$\times$W. 
For instance, $\mathcal{M}$ can be a NeRF, 3DGS or even a newer model \cite{Sun2024SVR}. Our goal is to calculate pixel-wise uncertainty $\mathcal{U}_\mathcal{T} \in \mathbb{R}^{\text{H}\times \text{W}}$ assosiated with viewpoint $\mathcal{T}$.

Let $\mathcal{S}=\{\mathcal{S}_0, \mathcal{S}_1, \mathcal{S}_2, \ldots, \mathcal{S}_S\}$ denote the set of $S$ source views used to train the radiance field $\mathcal{M}$. For each training view $\mathcal{S}_n \in \mathcal{S}$, we first render the corresponding depth map $\mathcal{D}_{\mathcal{S}_n}$, and then project it into the target pose $\mathcal{T}$, denoting this new depth map as $\mathcal{D}_{{\mathcal{S}_n}\rightarrow \mathcal{T}}$. This can be easily implemented by applying the relative camera transformation between $\mathcal{S}_n$ and $\mathcal{T}$, followed by a backward warping process performed according to $\mathcal{D}_\mathcal{T}.$

Then, we define the pixel-wise uncertainty for the view $\mathcal{T}$ as the average $L1$ difference between the projected depths from multiple viewpoints $\mathcal{S}_n \in \mathcal{S}$ and the rendered depth. Specifically, the uncertainty at pixel $(i,j)$ is calculated as:
\begin{equation}
    \mathcal{U}_\mathcal{T}^{i,j} = \frac{1}{S} \sum_{n=0}^{S}{\left| \mathcal{D}_\mathcal{T}^{i,j} - \mathcal{D}_{\mathcal{S}_n\rightarrow \mathcal{T}}^{i,j} \right| }
    \label{eq:unc_pixel}
\end{equation}
where the superscript $i,j$ represents the pixel index. In practice, if the projection $D_{{\mathcal{S}_n}\rightarrow \mathcal{T}}^{i,j}$ results in a negative depth, we exclude it from the average.

\subsubsection{Image-level Uncertainty}
\label{sec:unc_image}
The uncertainty for an entire image can be simply defined using Eq. \eqref{eq:unc_pixel} by summing the pixel-wise uncertainties across the image. However, depending on the downstream task, we may find that warping different cues such as the color of rendered images may yield better results.

Therefore, given a pose $\mathcal{T}$, we calculate the image-level uncertainty $\mathcal{U}_\mathcal{T} \in \mathbb{R}$ as follows. For every pose $\mathcal{S}_n$ in the training view, we first render the corresponding image $\mathcal{I}_{\mathcal{S}_n}$, and then use the rendered depth $\mathcal{D}_\mathcal{S}$ to project it into the target pose $\mathcal{T}$ by means of backward warping, denoted as $\mathcal{I}_{{\mathcal{S}_n}\rightarrow \mathcal{T}}$. The image-level uncertainty $\mathcal{U}_\mathcal{T}$ is then computed as:
\begin{equation}
    \mathcal{U}_\mathcal{T} = \sum_{i,j}\min_{n} {\left|\mathcal{I}_\mathcal{T}^{i,j} - \mathcal{I}_{{\mathcal{S}_n}\rightarrow \mathcal{T}}^{i,j} \right|}
    \label{eq:unc_image}
\end{equation}
where the summation is taken over all pixels. Note that, in this case, we compute the per-pixel minimum reprojection difference rather than averaging, as this approach has been shown to be more robust to occlusions and disocclusions, and generally, parts of the image that are visible in some views but not others \cite{monodepth2}. Importantly, our approach does not require any ground-truth images or depths, as it relies solely on the rendered images and depths to compute uncertainties. It is completely training-free and does not depend on the specific neural field implementation.

\begin{table}[t]
\centering
\begin{tabular}{l|c|c}
\toprule
Method           & ScanNet++ & ETH3D \\ 
\midrule
NeRF + BayesRays \cite{goli2023} & 0.438     & 0.278 \\
\bf NeRF + WarpRF (ours)      & \fst 0.423     & \fst 0.250 \\ 
\midrule
3DGS + Manifold \cite{lyu2024manifold}  & 0.510     & 0.522 \\
3DGS + FisherRF \cite{Jiang2024FisherRF}  & 0.355     & 0.295 \\
\bf 3DGS + WarpRF (ours)      & \fst 0.337     & \fst 0.227 \\
\bottomrule
\end{tabular}\vspace{-0.2cm}
\caption{\textbf{AUSE evaluation $\downarrow$ on ScanNet++ and ETH3D.} Comparison with methods for NeRF (top) and 3DGS (bottom).}
\label{tab:ause}
\end{table}

\begin{figure}[t]
\centering
\renewcommand{\tabcolsep}{1pt}
\begin{tabular}{ccccc}
(a) & (b) & (c) & (d) & (e) \\
\includegraphics[width=.19\linewidth]{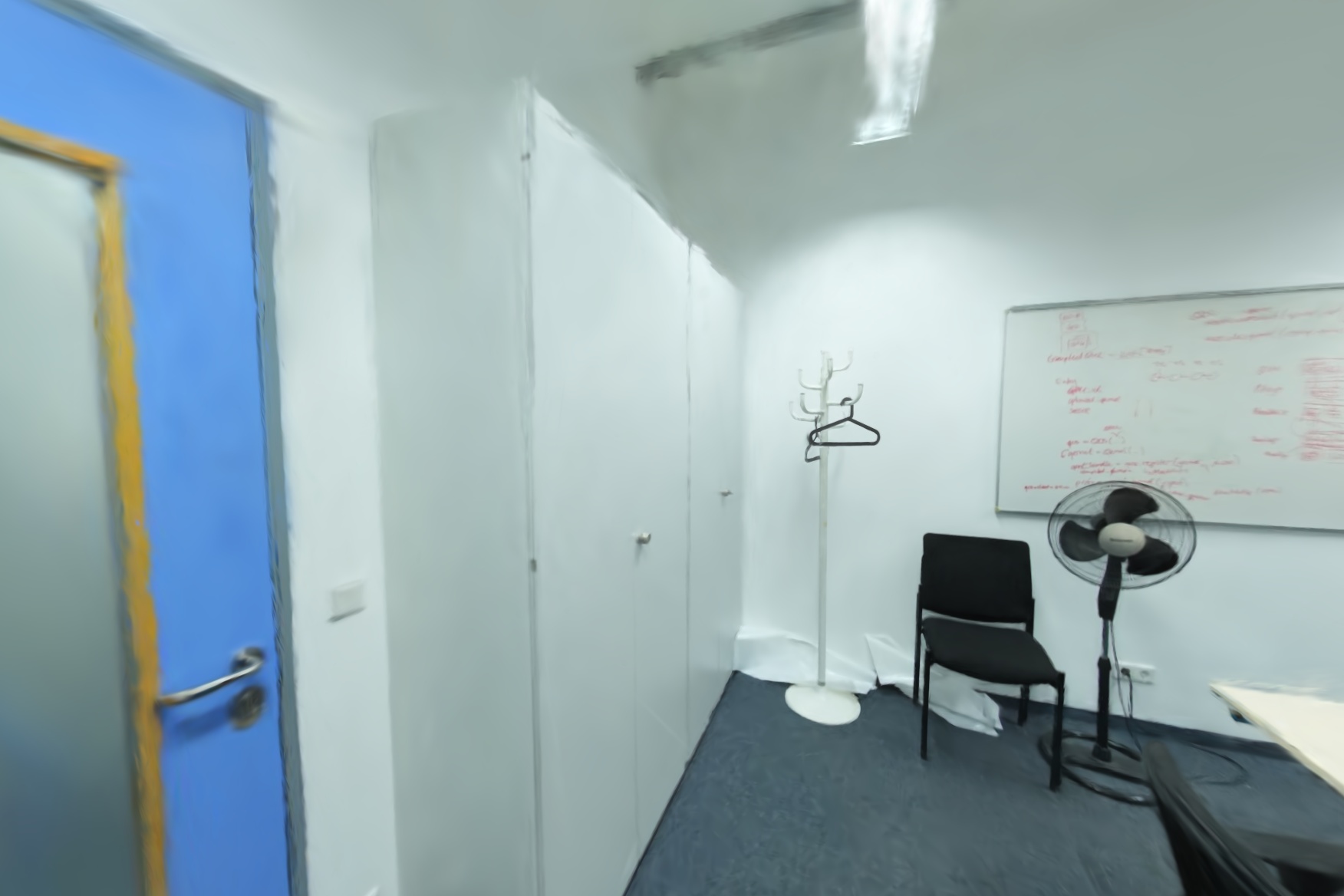} & 
\includegraphics[width=.19\linewidth]{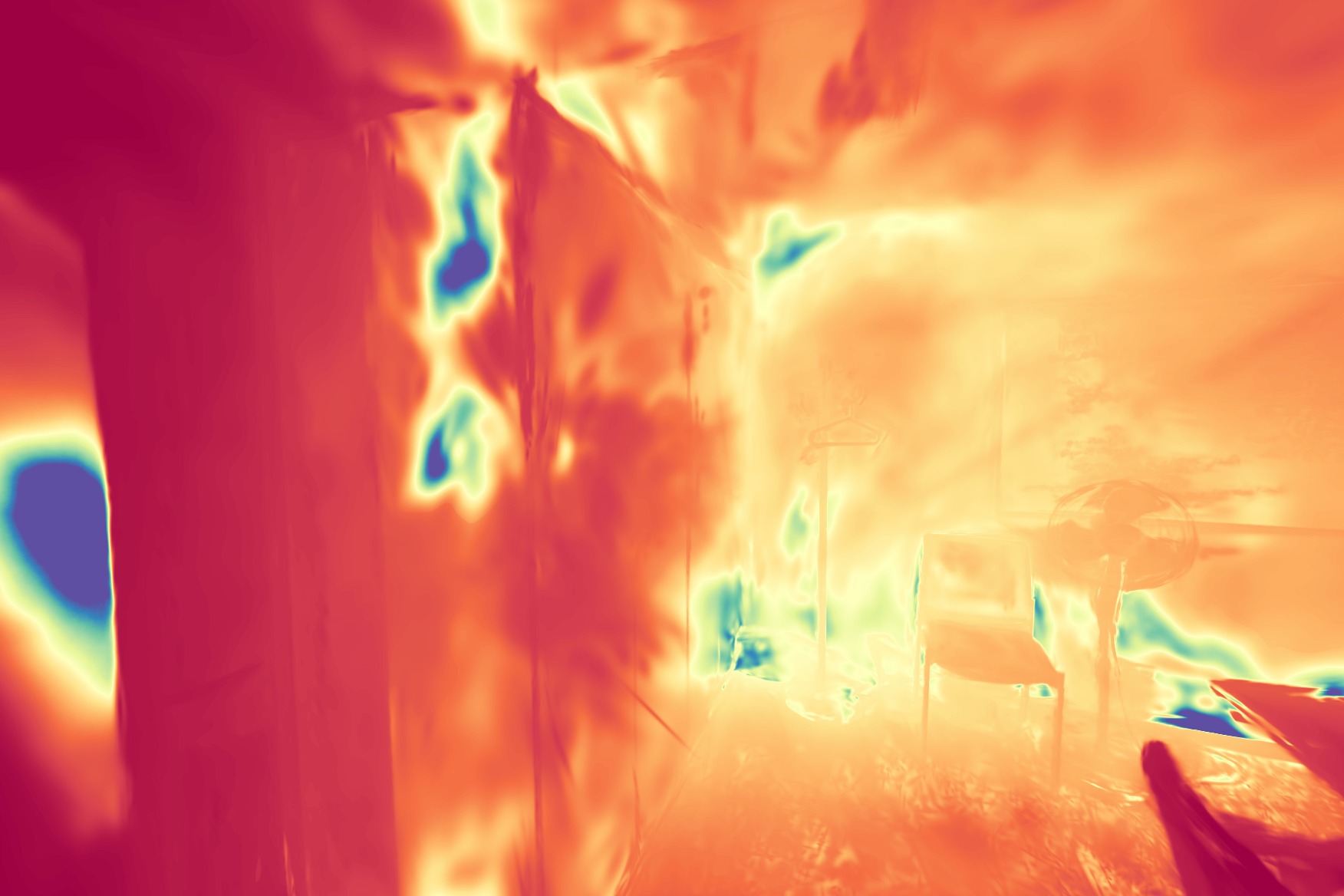}  & 
\includegraphics[width=.19\linewidth]{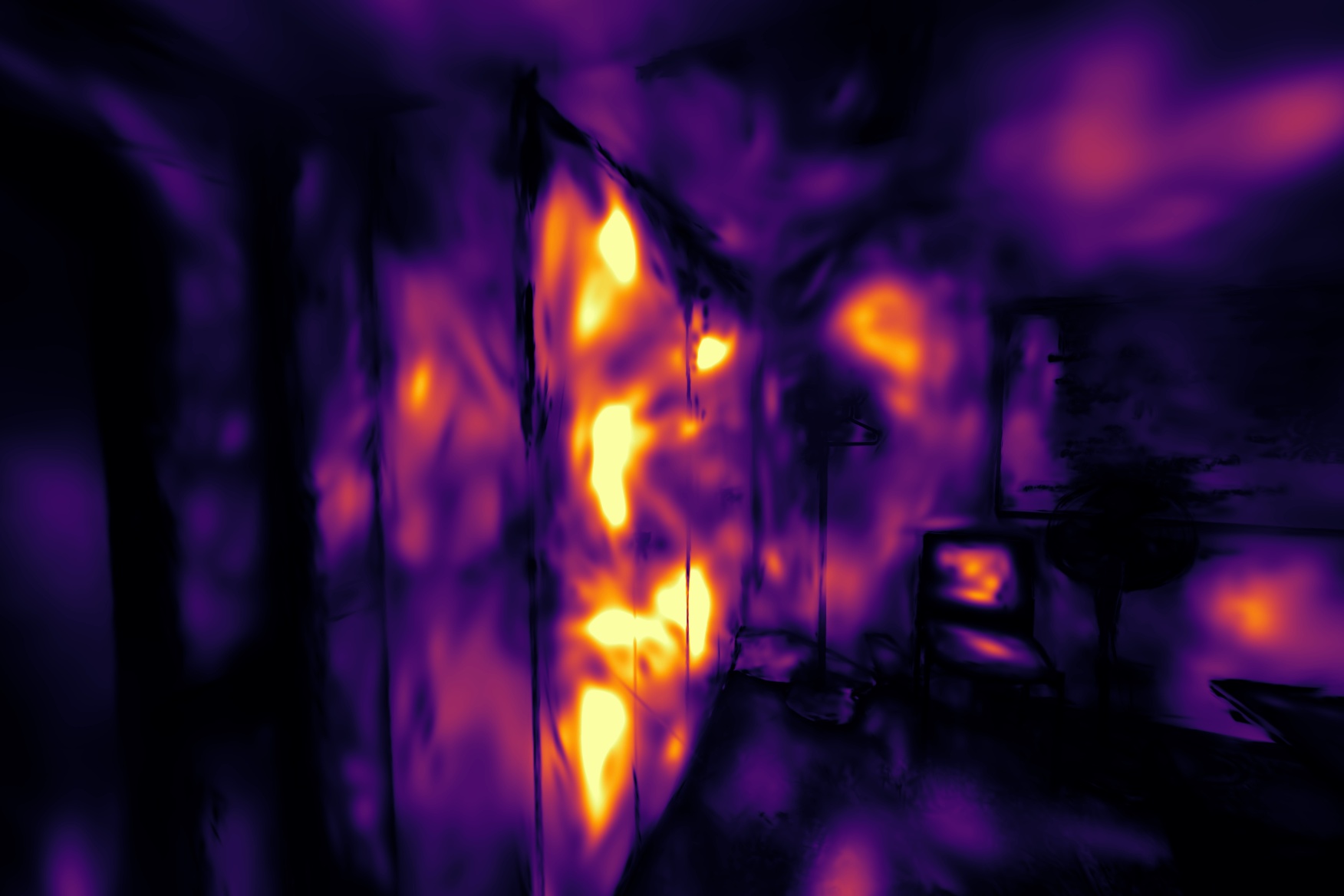} & 
\includegraphics[width=.19\linewidth]{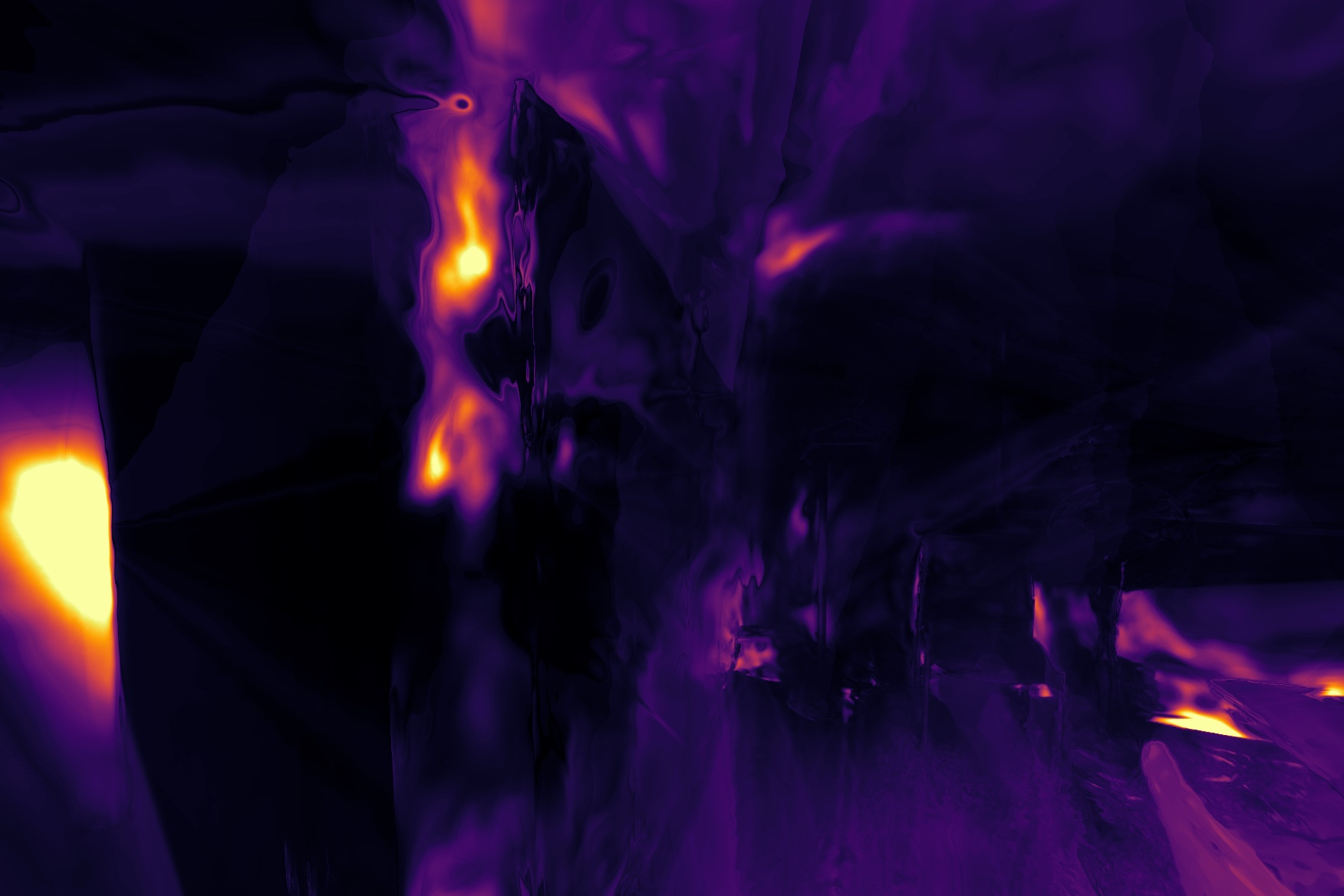} & 
\includegraphics[width=.19\linewidth]{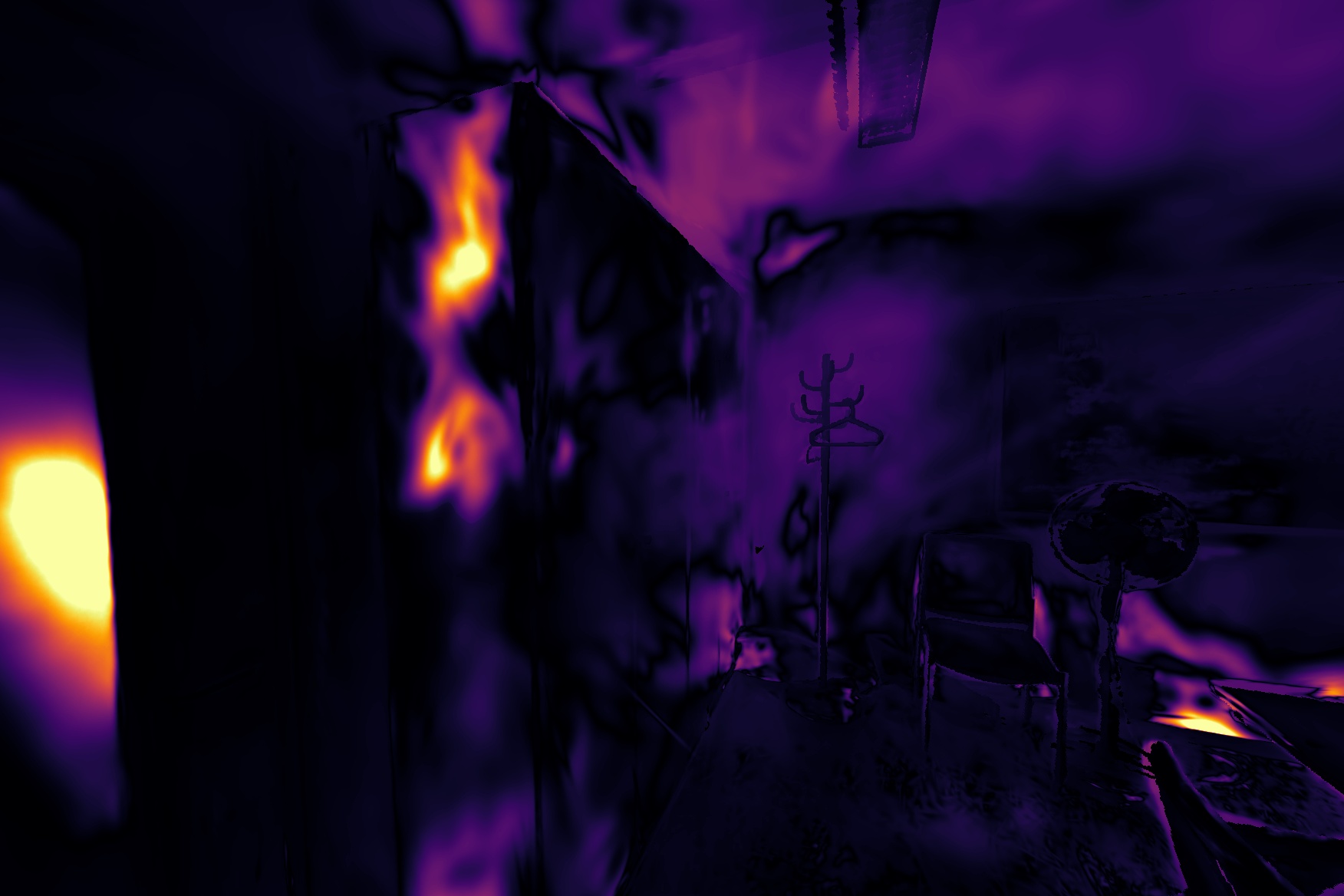} \\ 
\includegraphics[width=.19\linewidth]{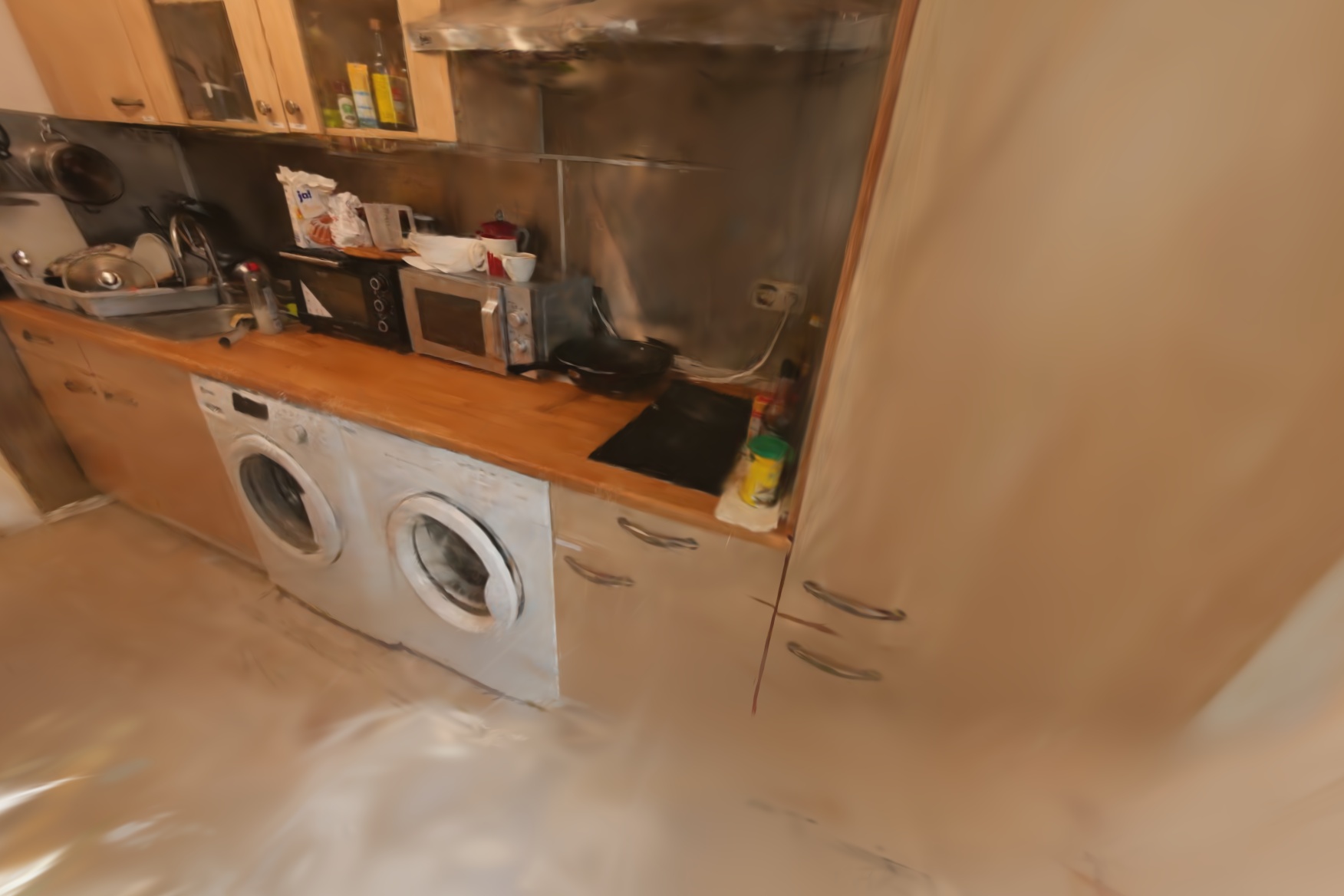} & 
\includegraphics[width=.19\linewidth]{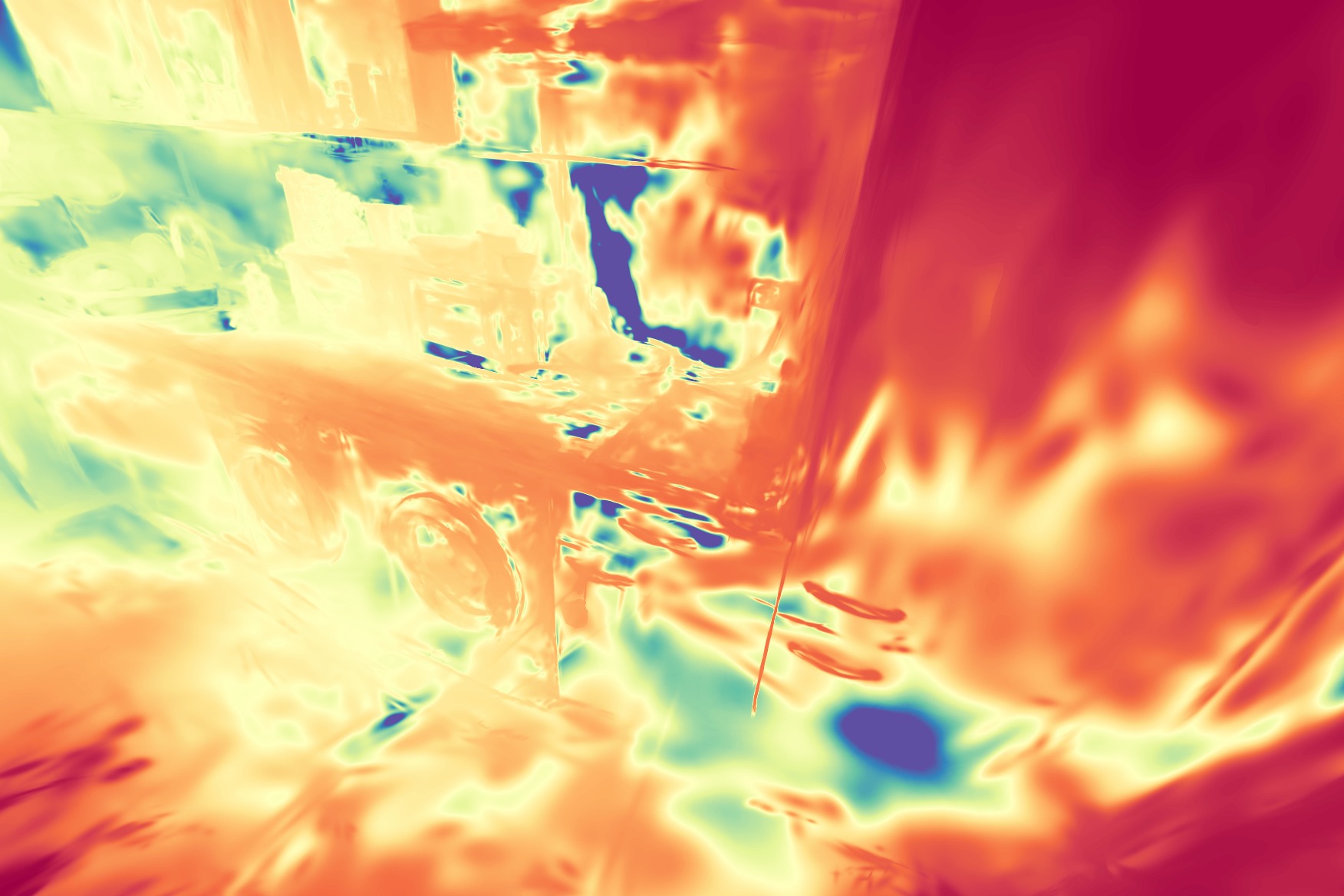} & 
\includegraphics[width=.19\linewidth]{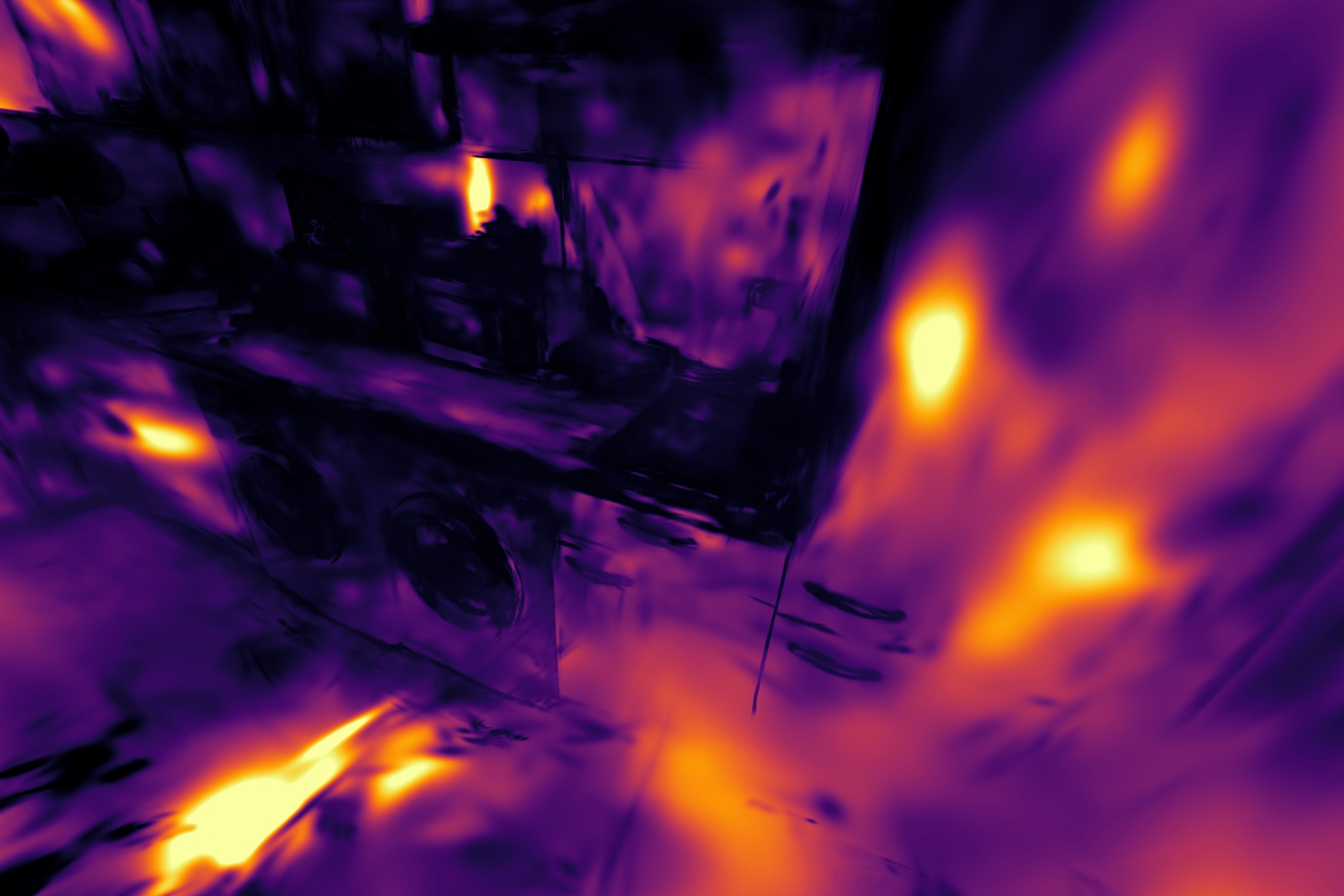} & 
\includegraphics[width=.19\linewidth]{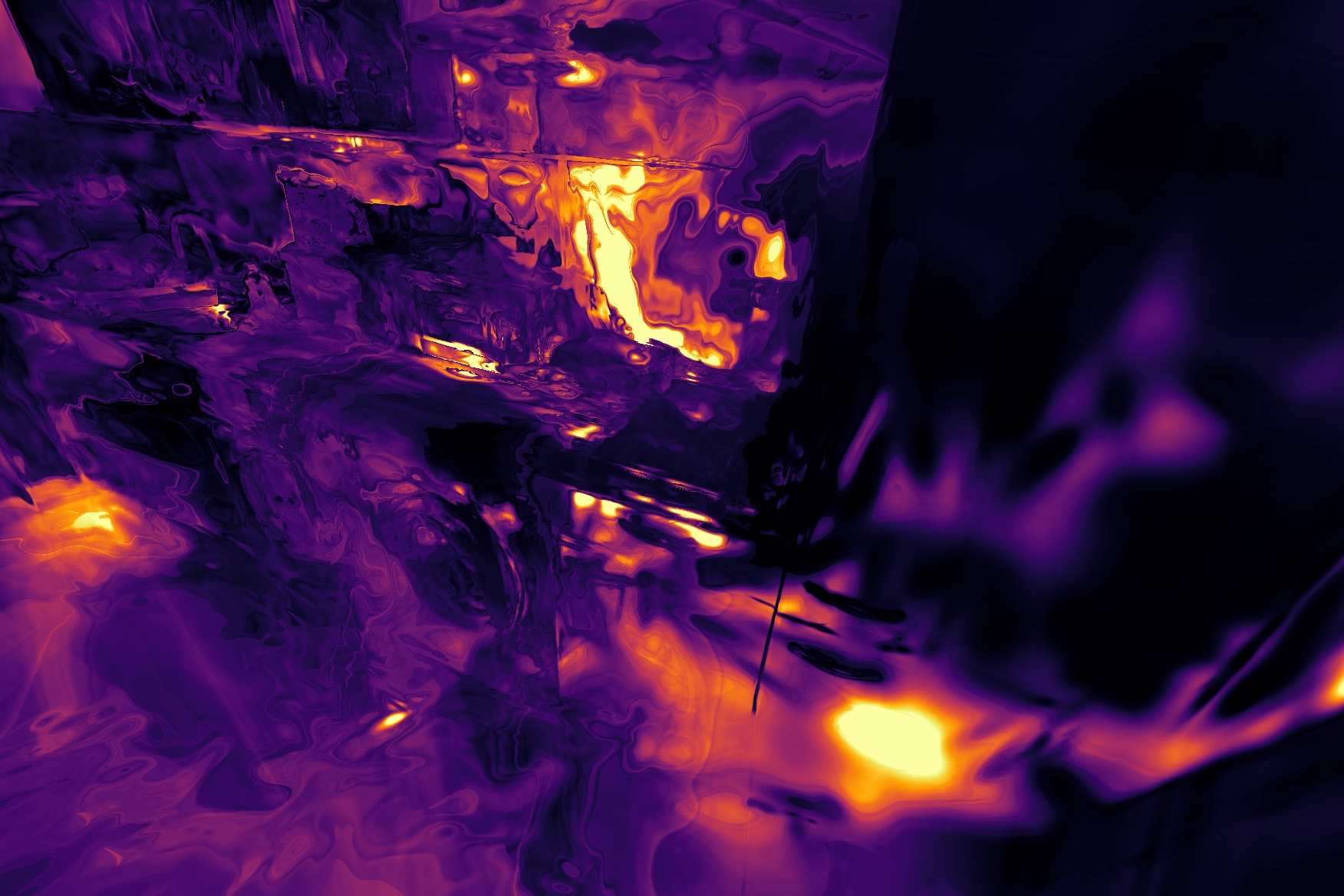} & 
\includegraphics[width=.19\linewidth]{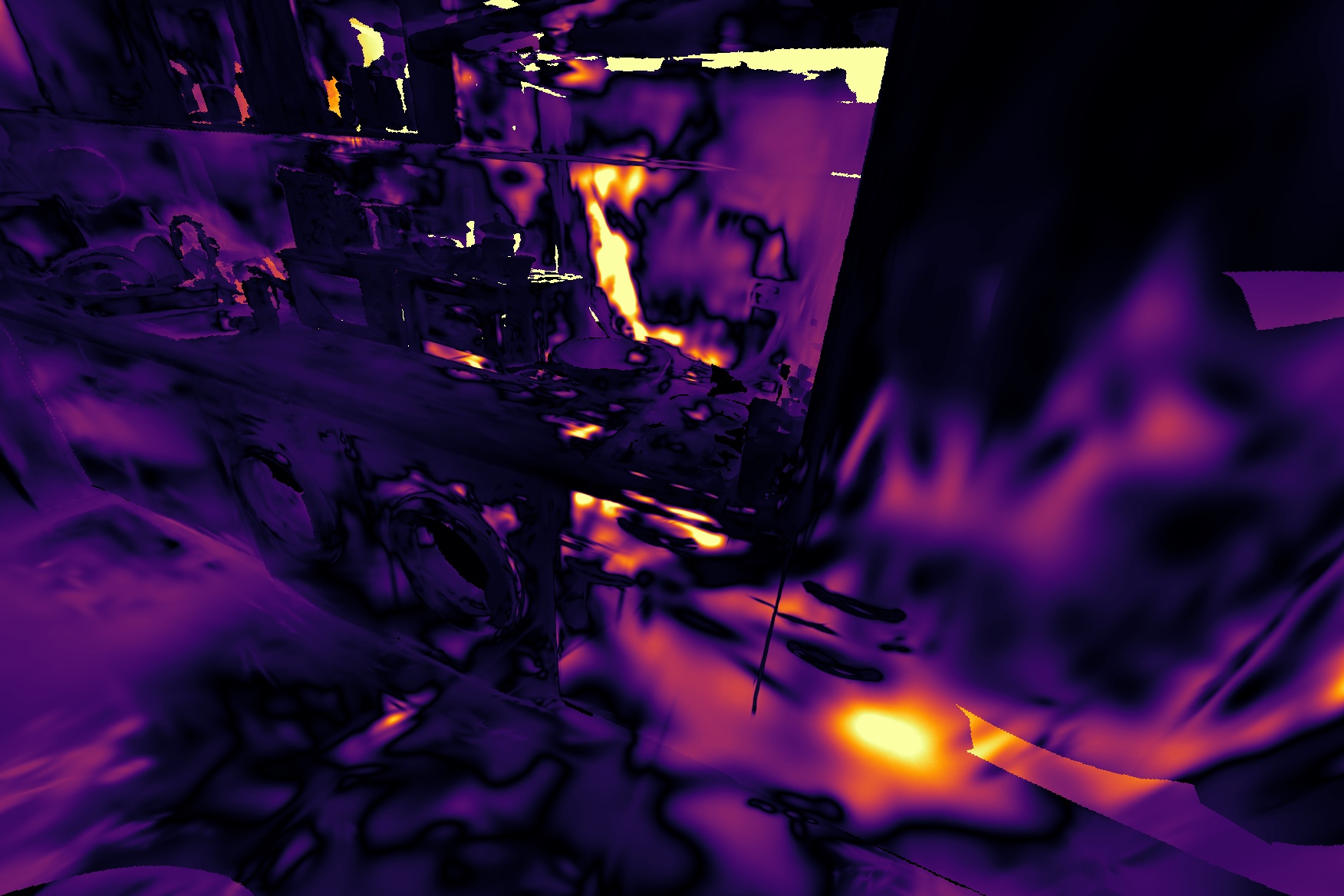} \\
\includegraphics[width=.19\linewidth]{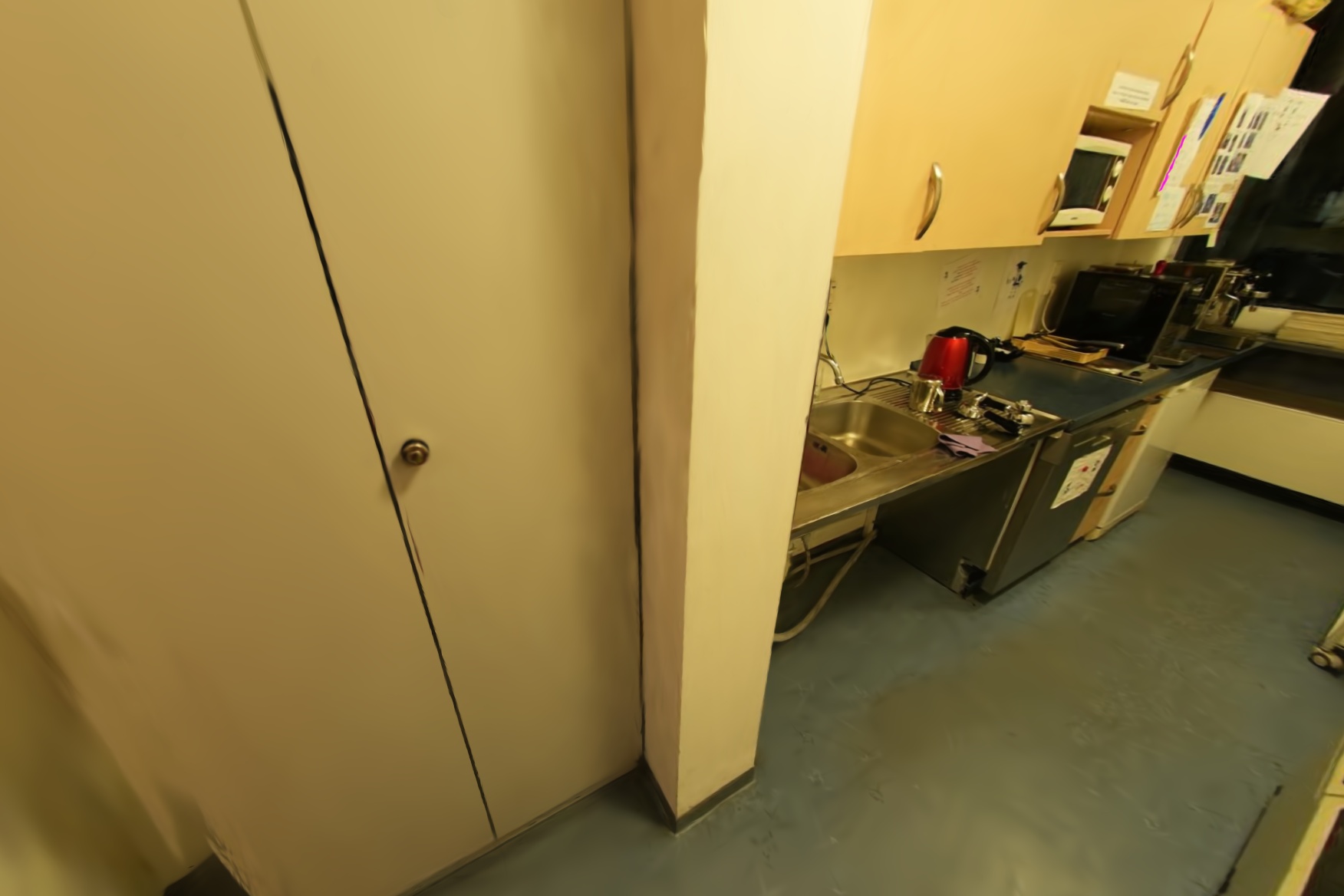} & 
\includegraphics[width=.19\linewidth]{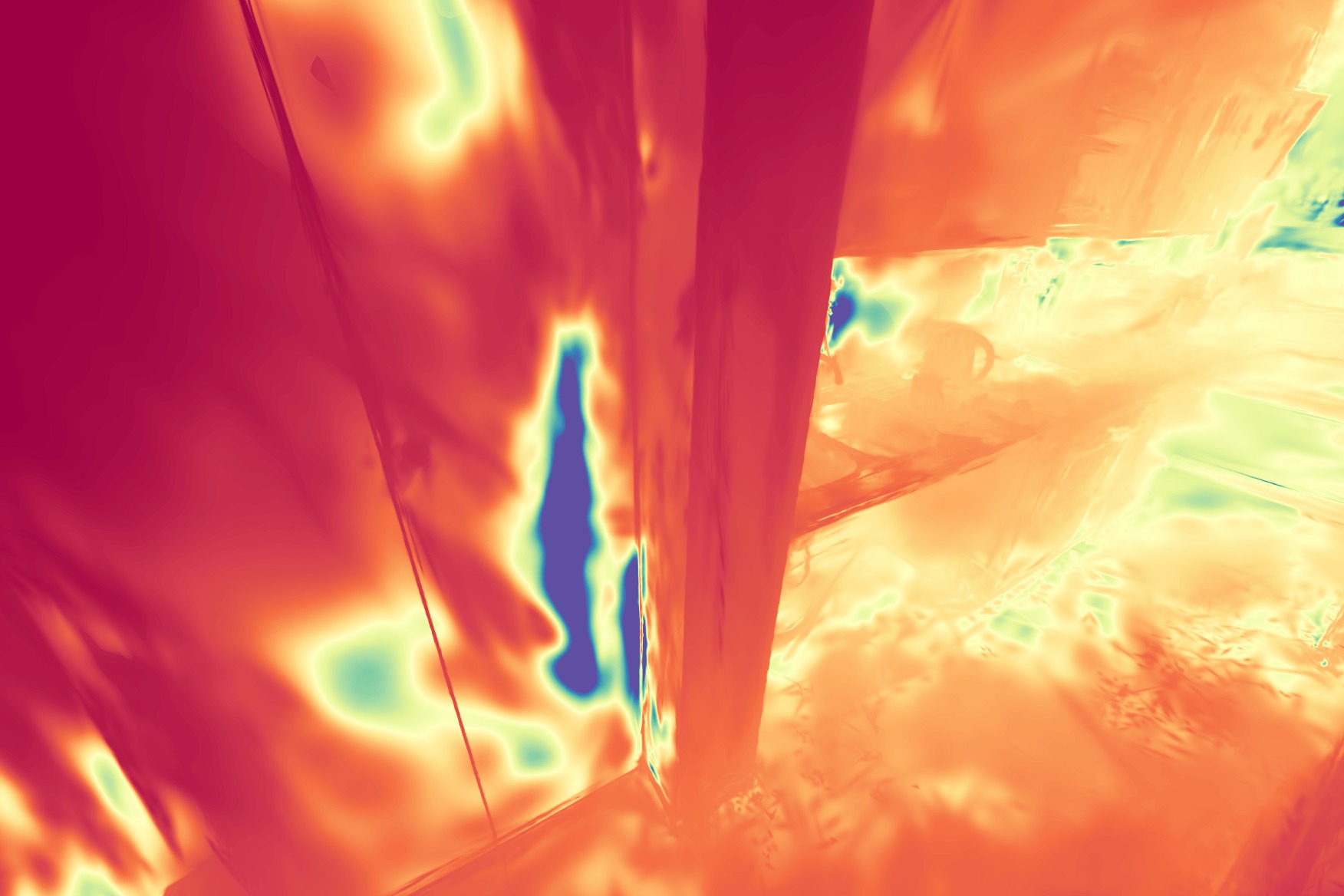}  & 
\includegraphics[width=.19\linewidth]{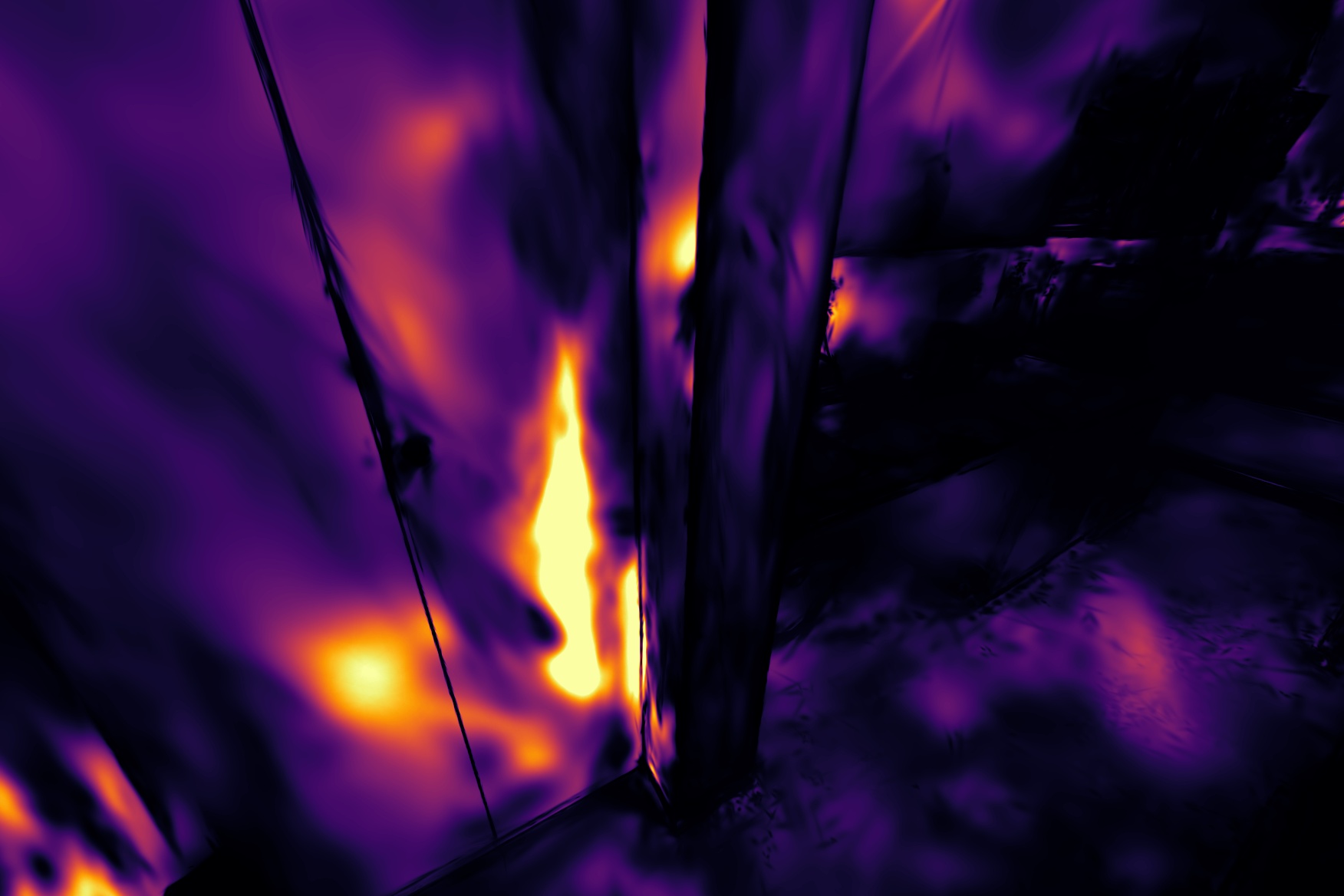} & 
\includegraphics[width=.19\linewidth]{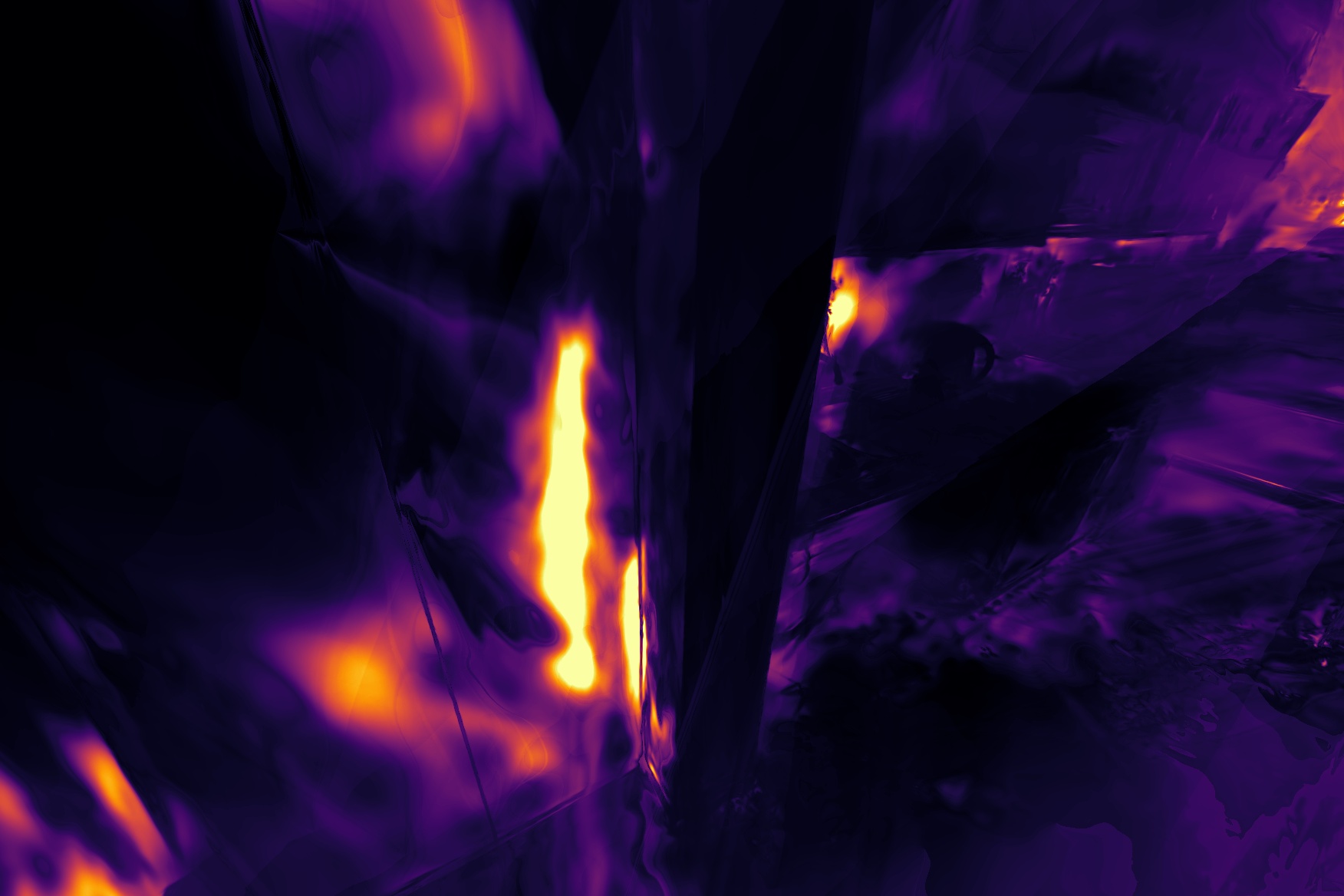} & 
\includegraphics[width=.19\linewidth]{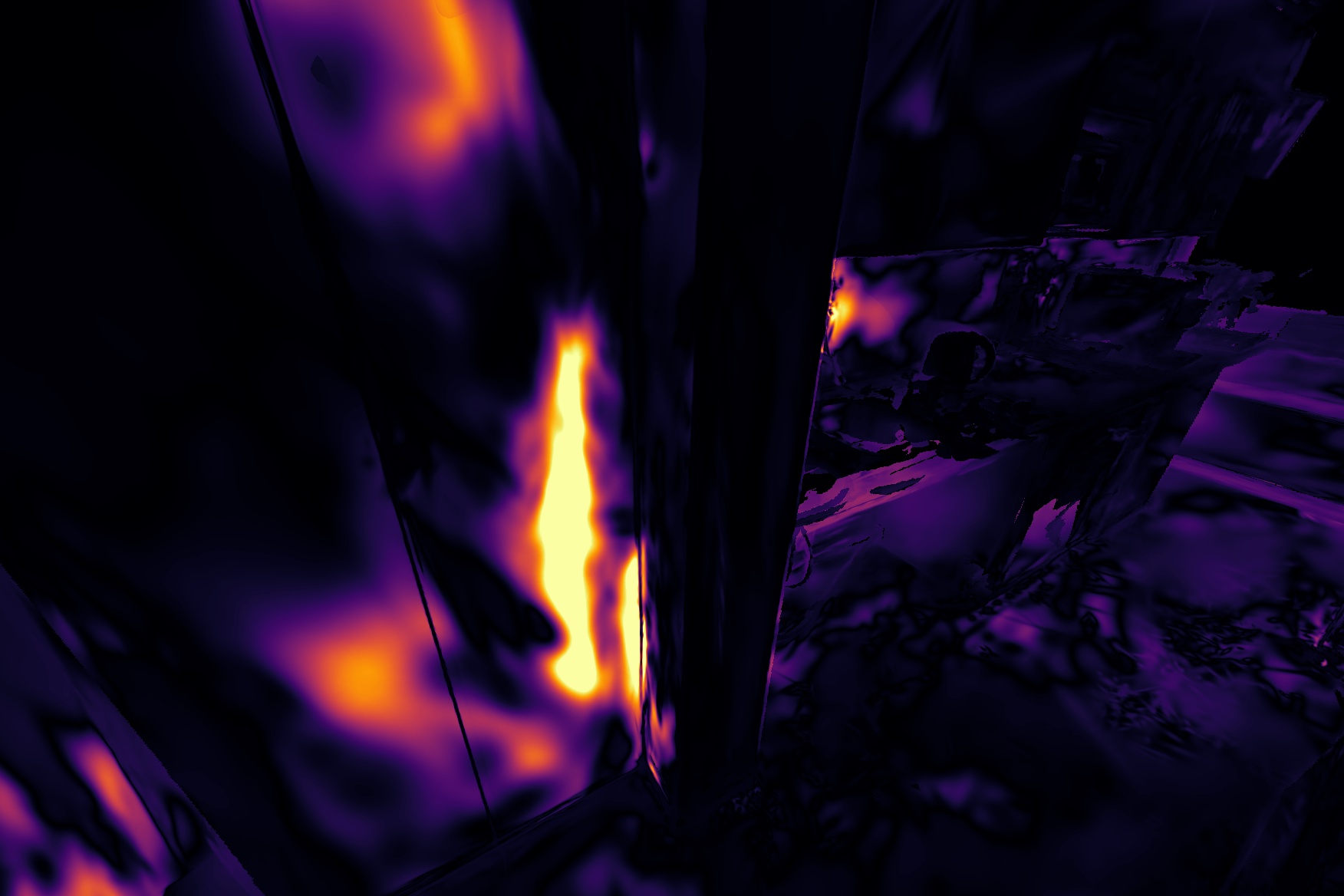} \\
\includegraphics[width=.19\linewidth]{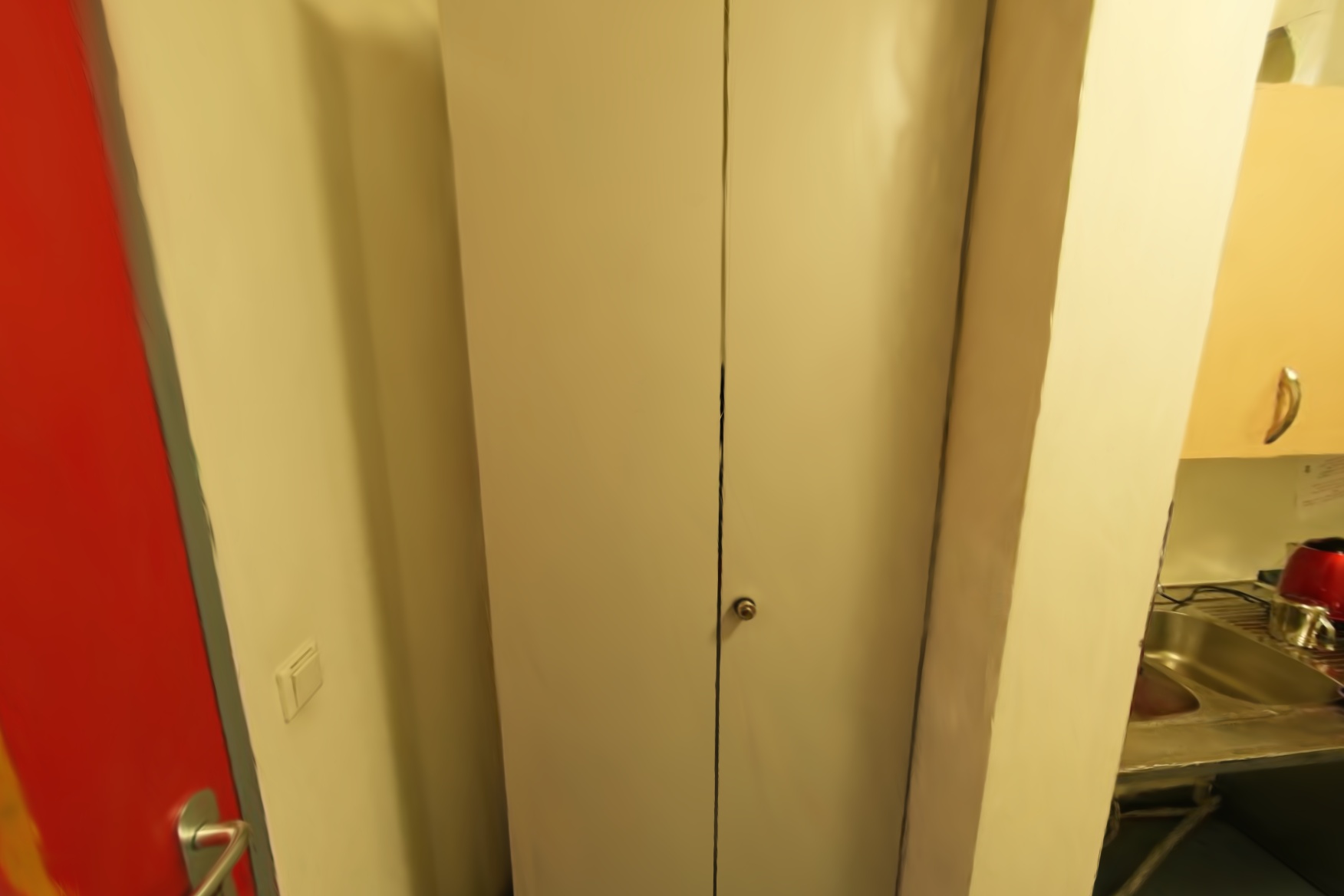} & 
\includegraphics[width=.19\linewidth]{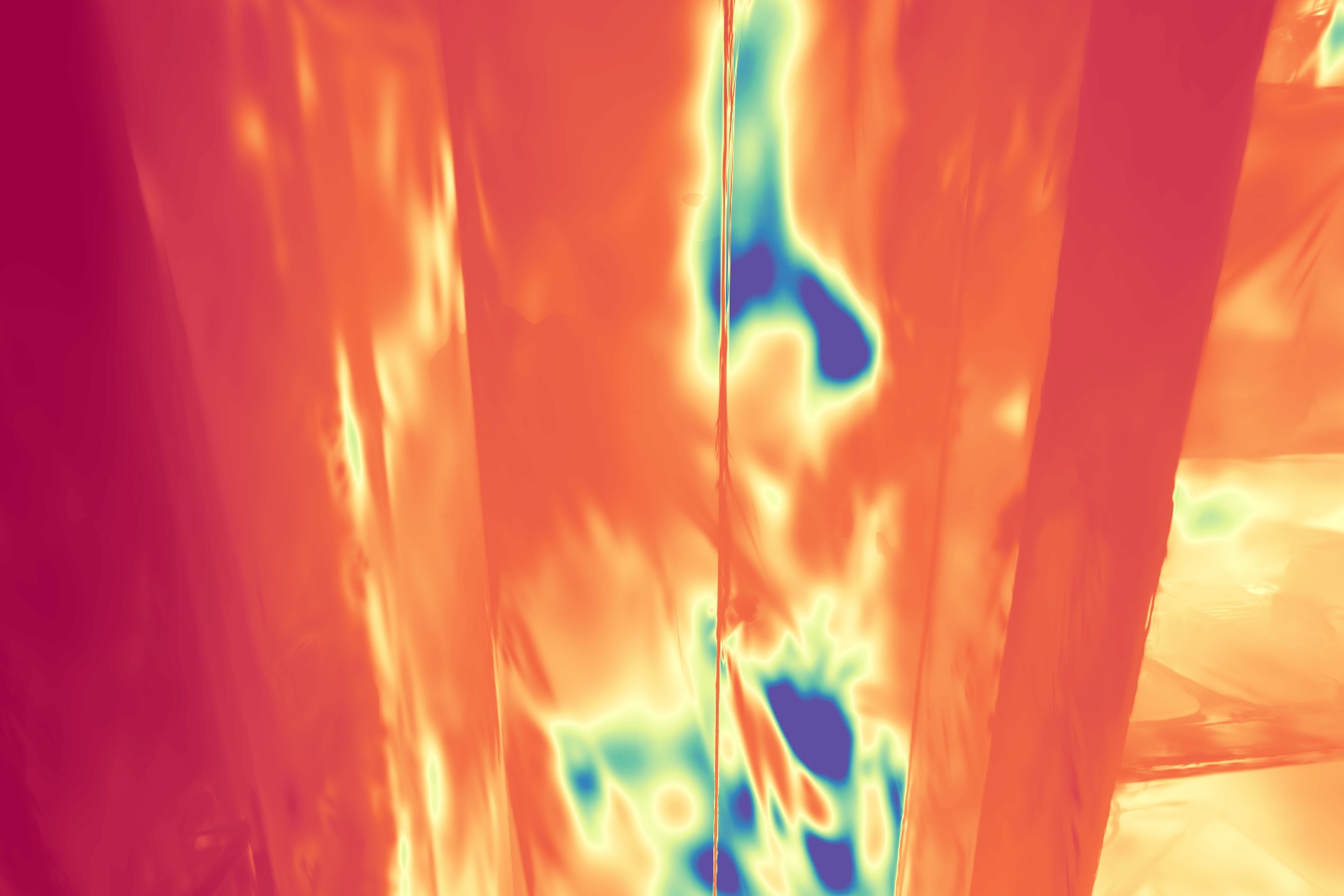}& 
\includegraphics[width=.19\linewidth]{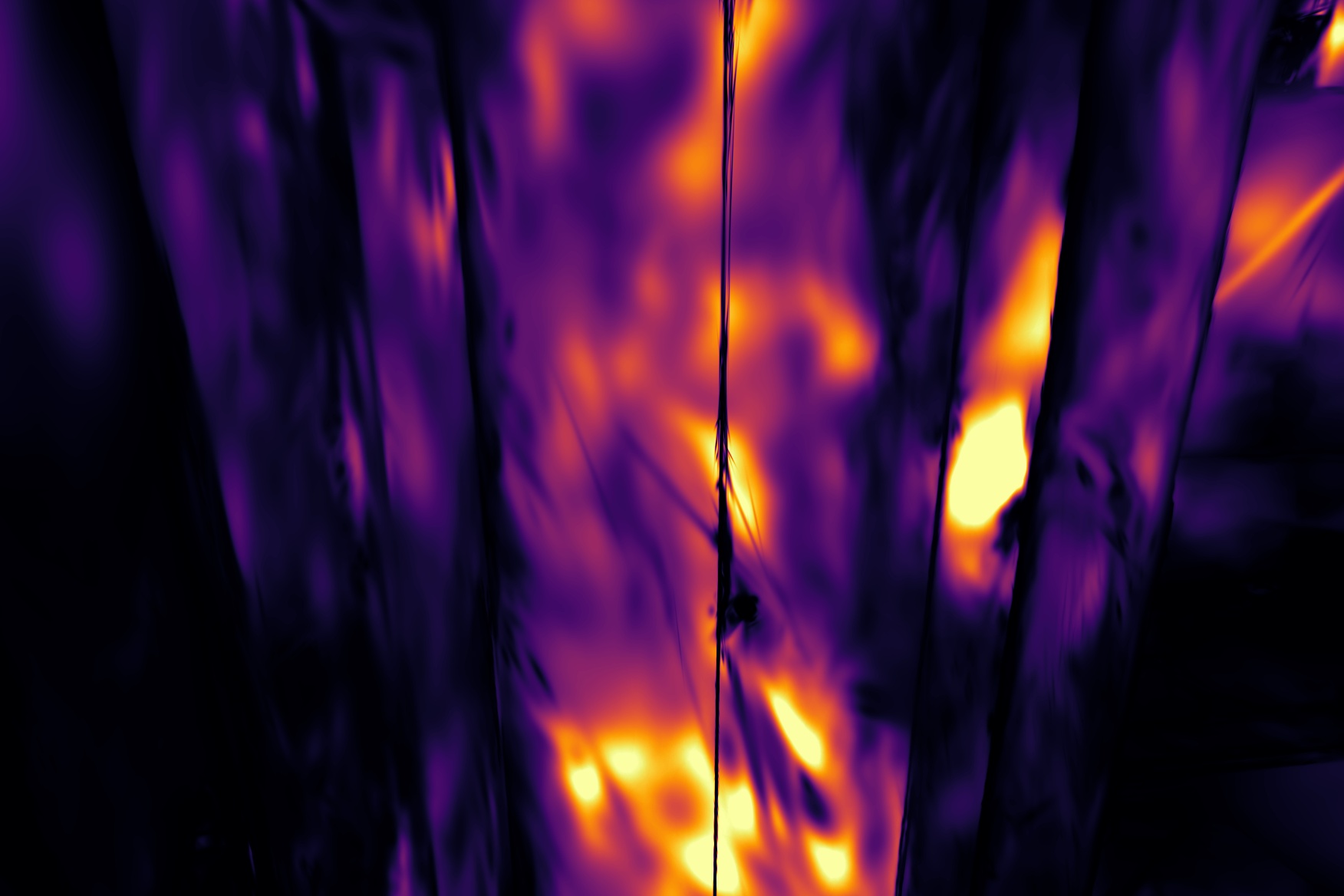} & 
\includegraphics[width=.19\linewidth]{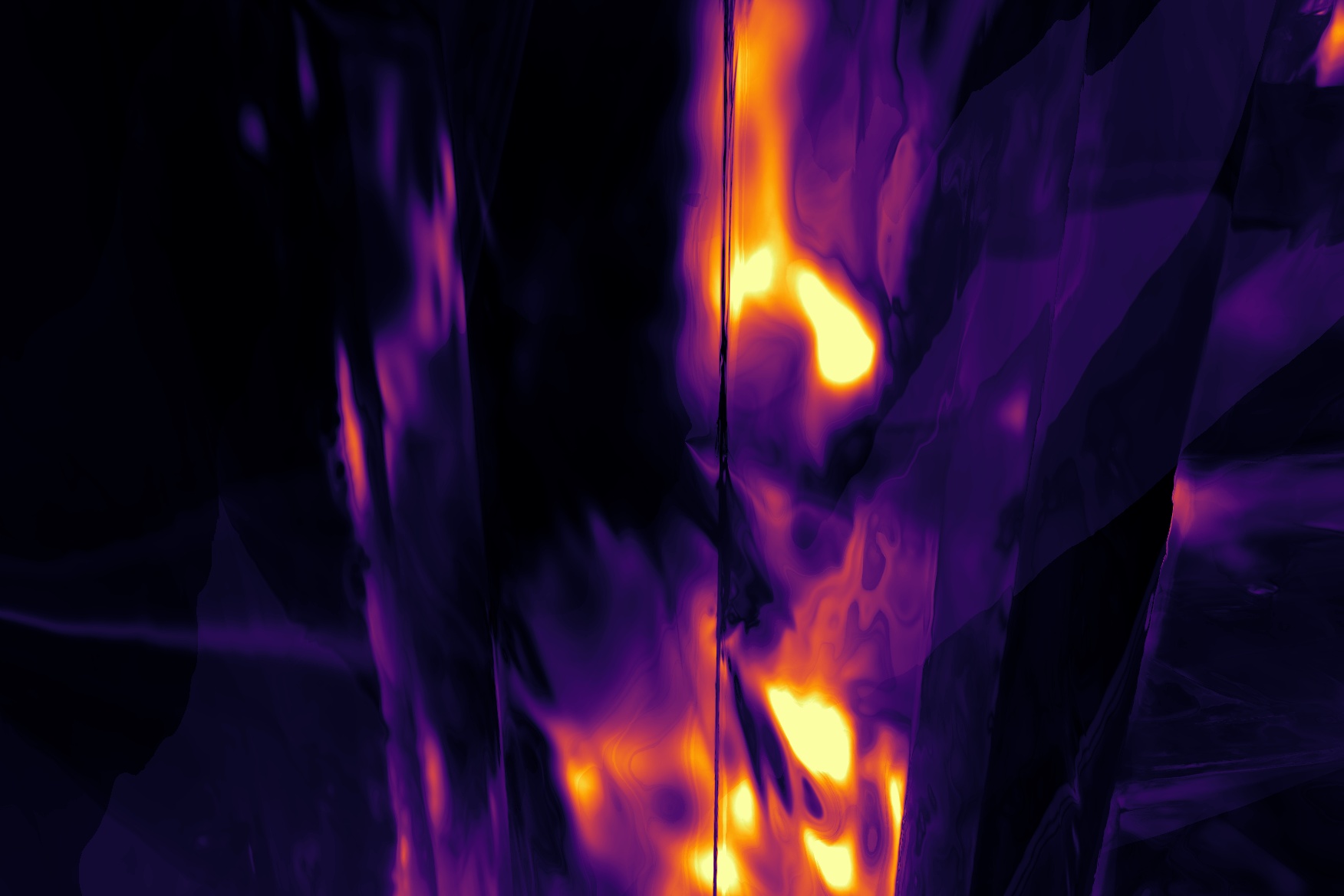} & 
\includegraphics[width=.19\linewidth]{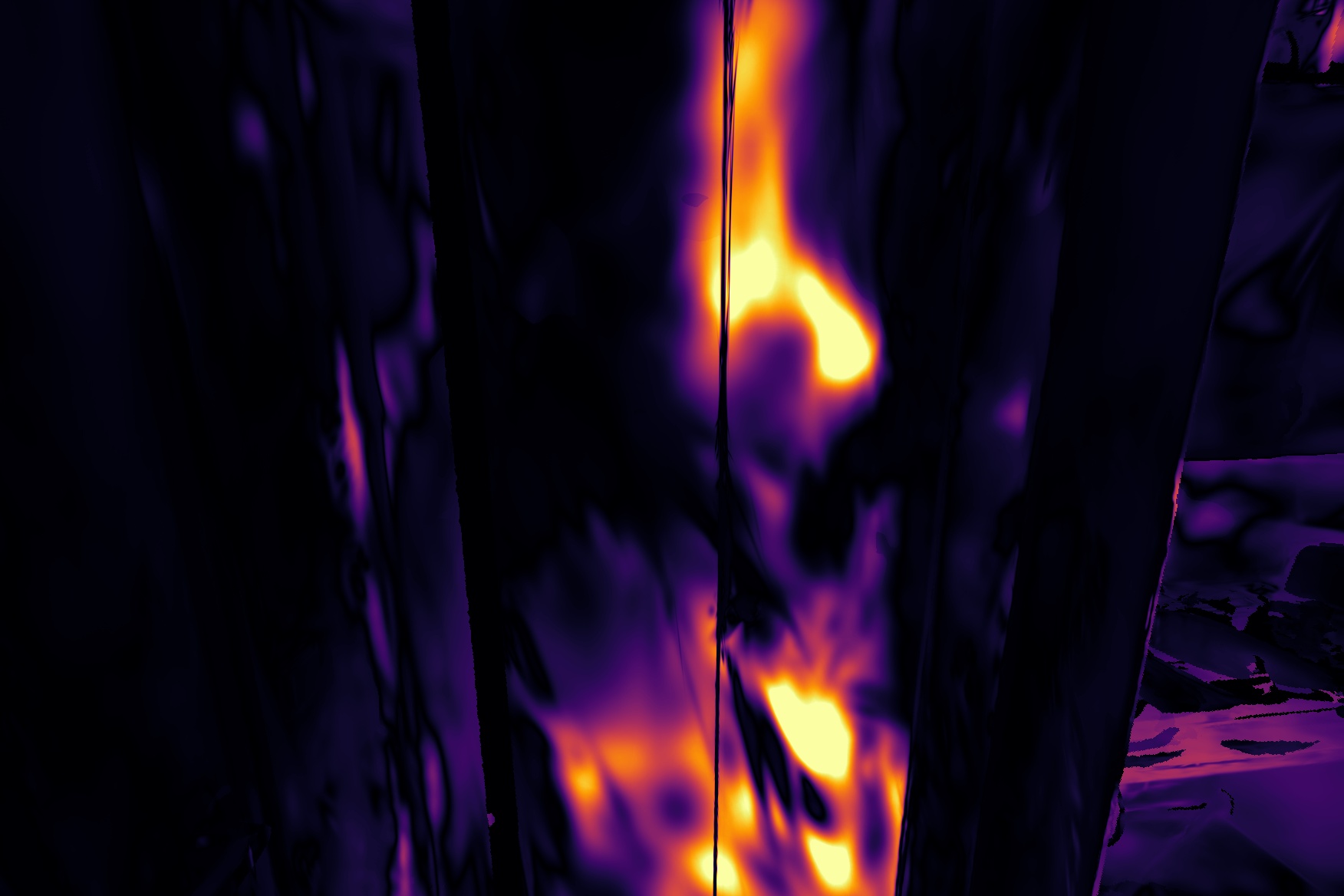}  \\
\includegraphics[width=.19\linewidth]{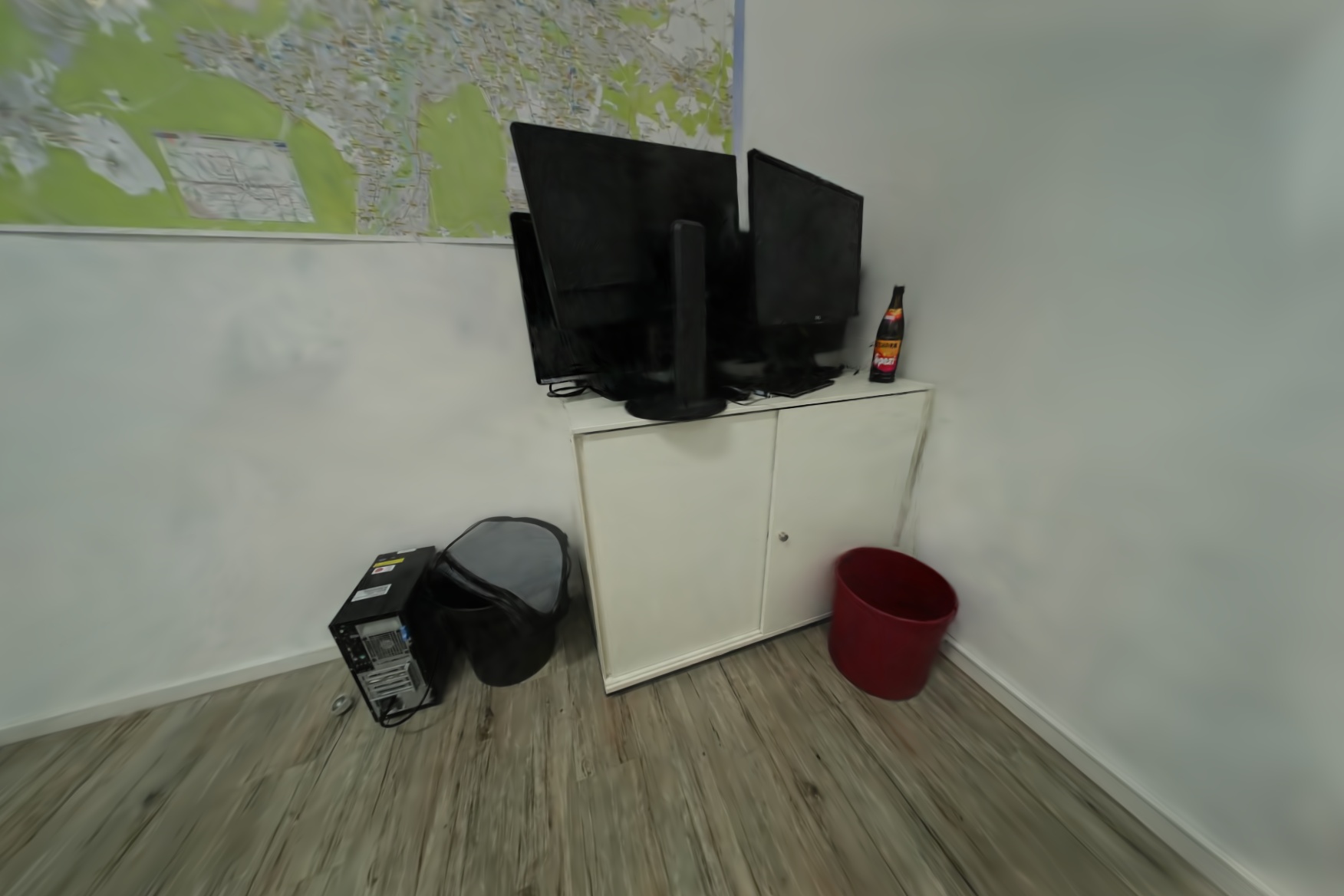} & 
\includegraphics[width=.19\linewidth]{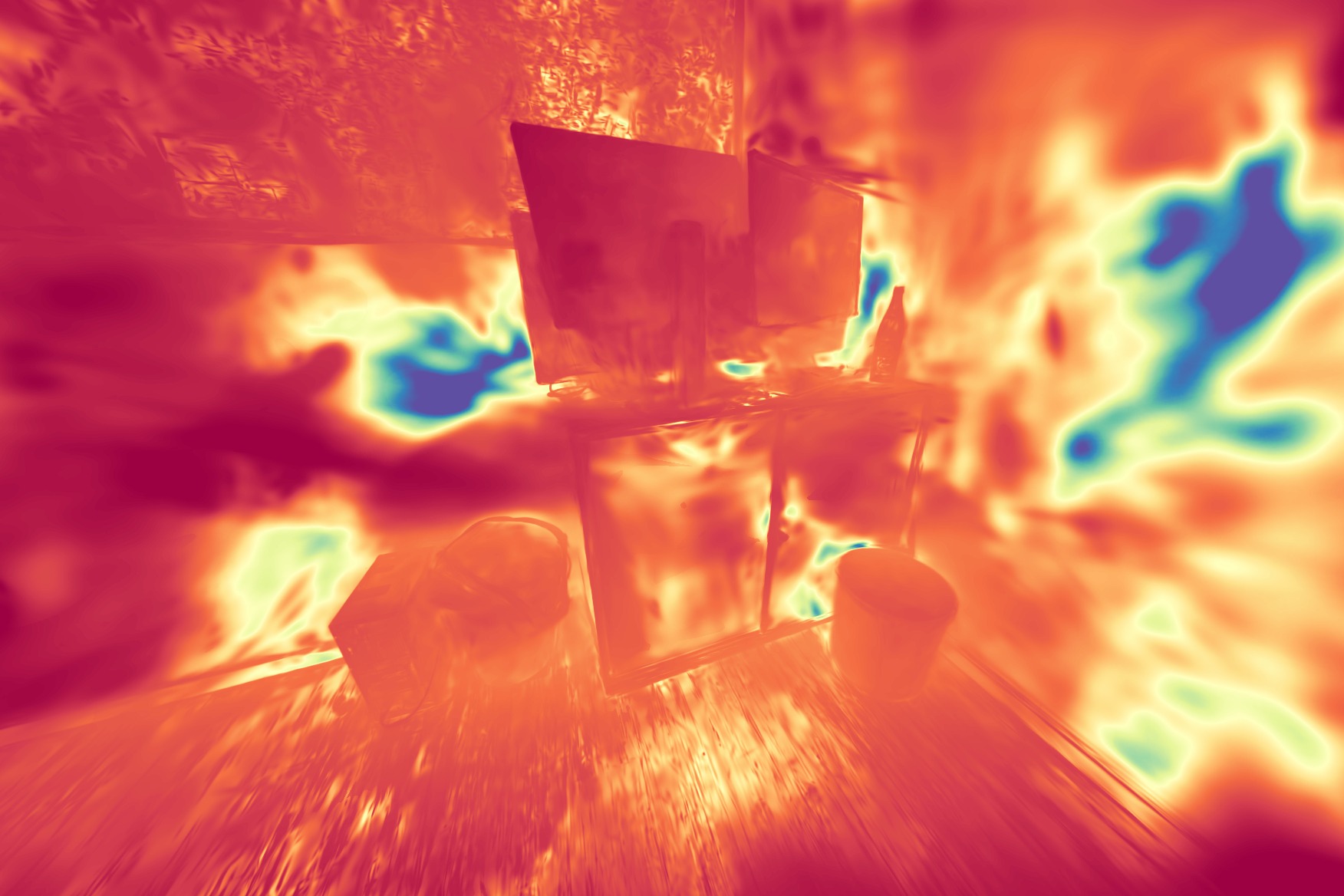}& 
\includegraphics[width=.19\linewidth]{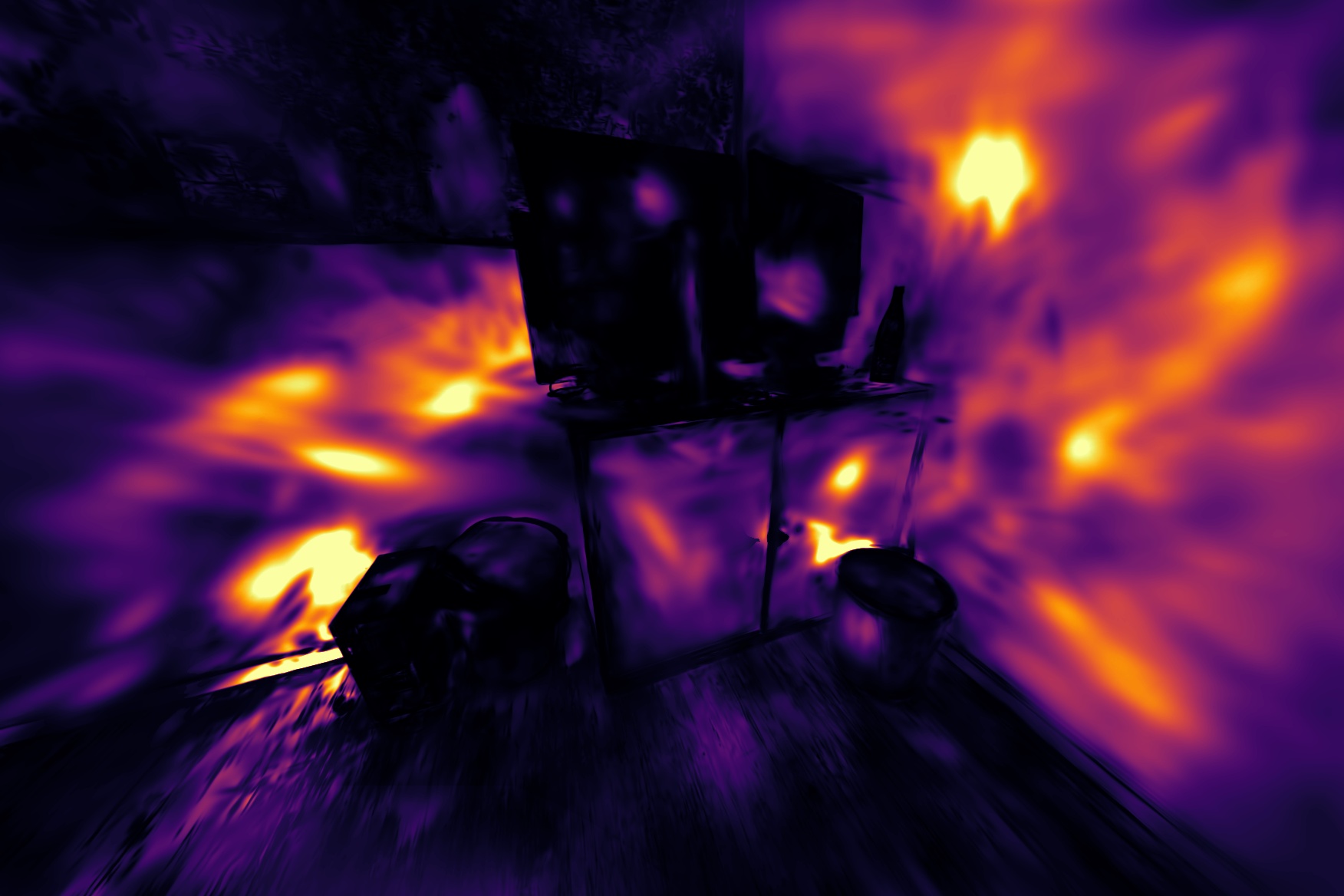} & 
\includegraphics[width=.19\linewidth]{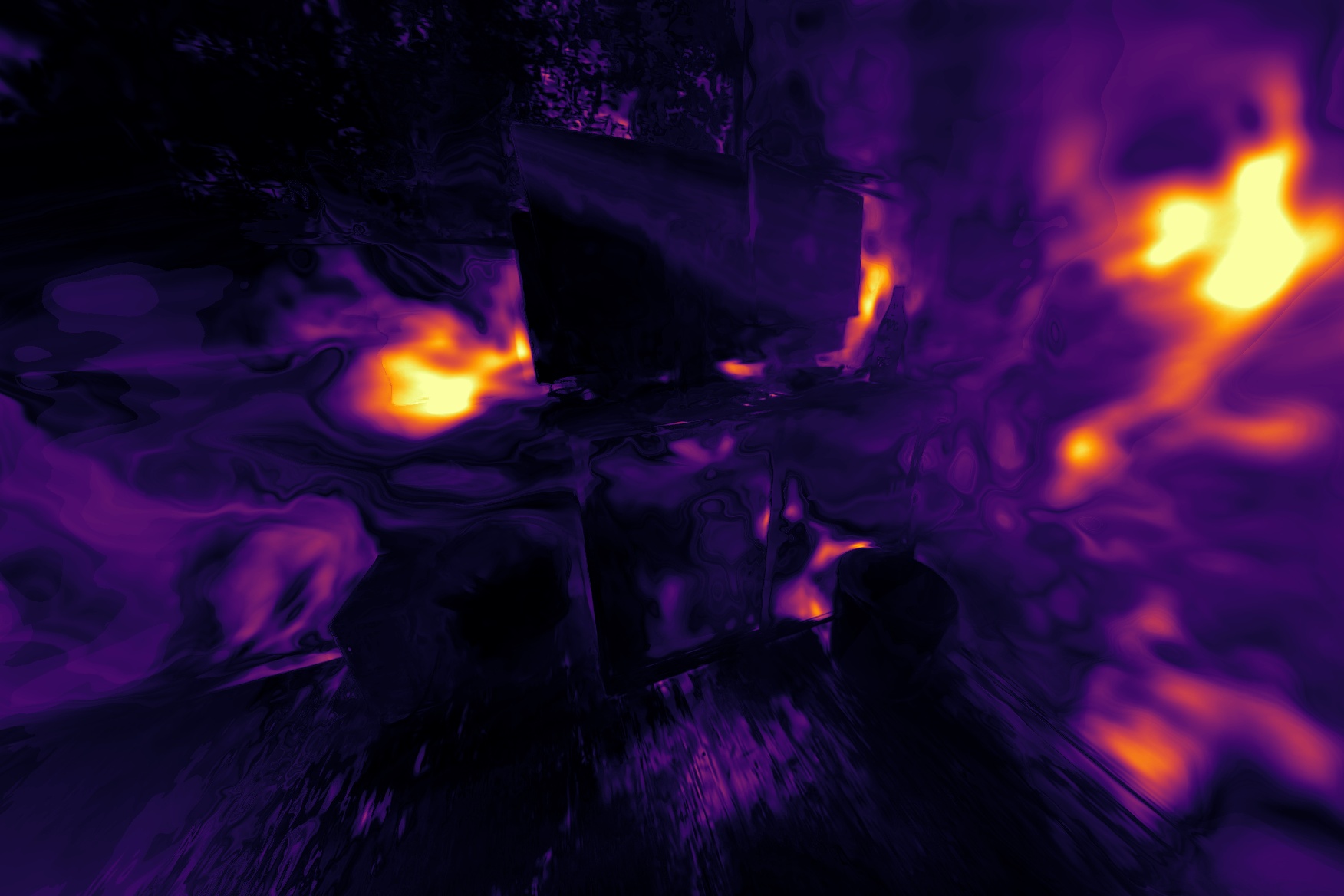}  & 
\includegraphics[width=.19\linewidth]{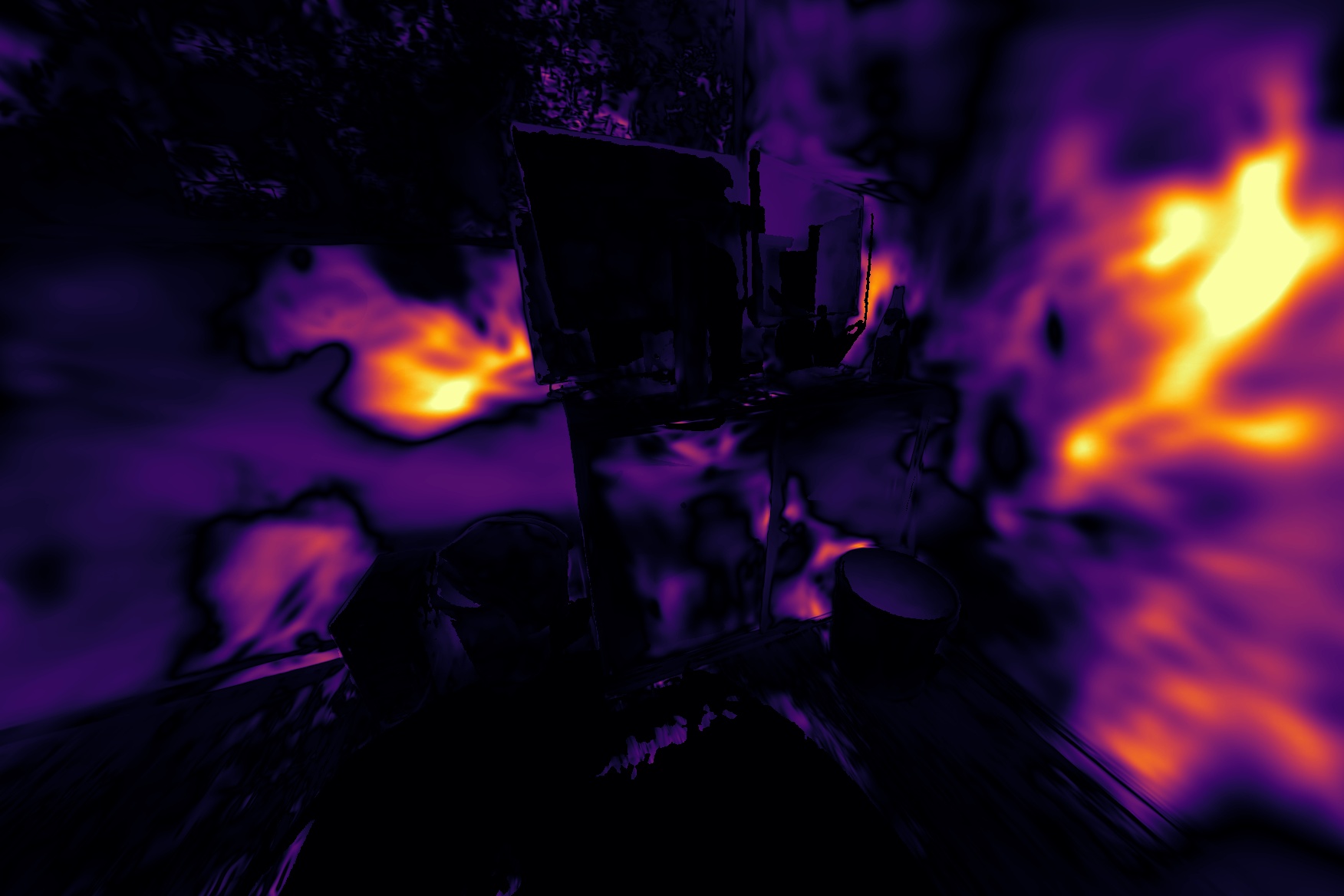} \\
\includegraphics[width=.19\linewidth]{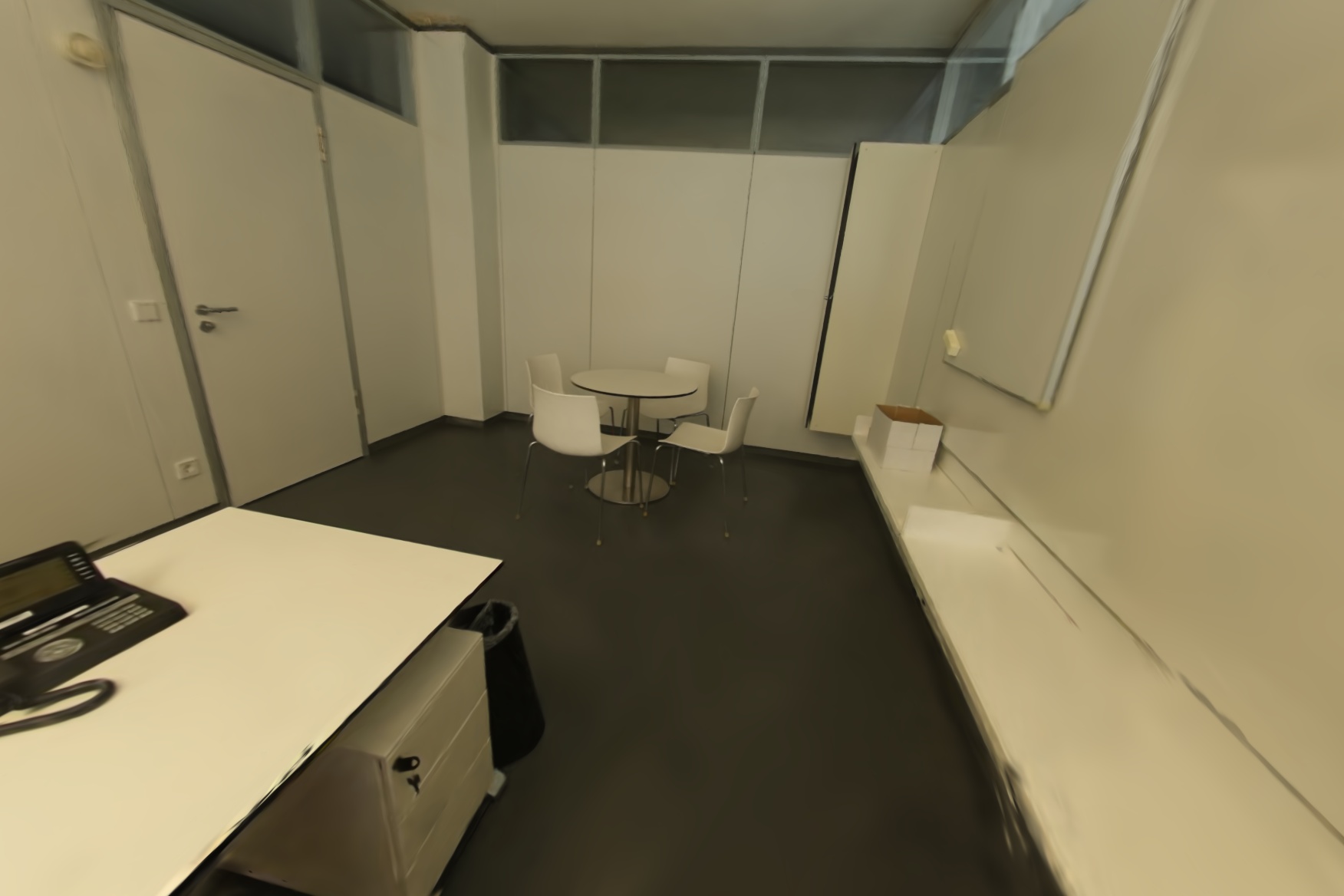} &
\includegraphics[width=.19\linewidth]{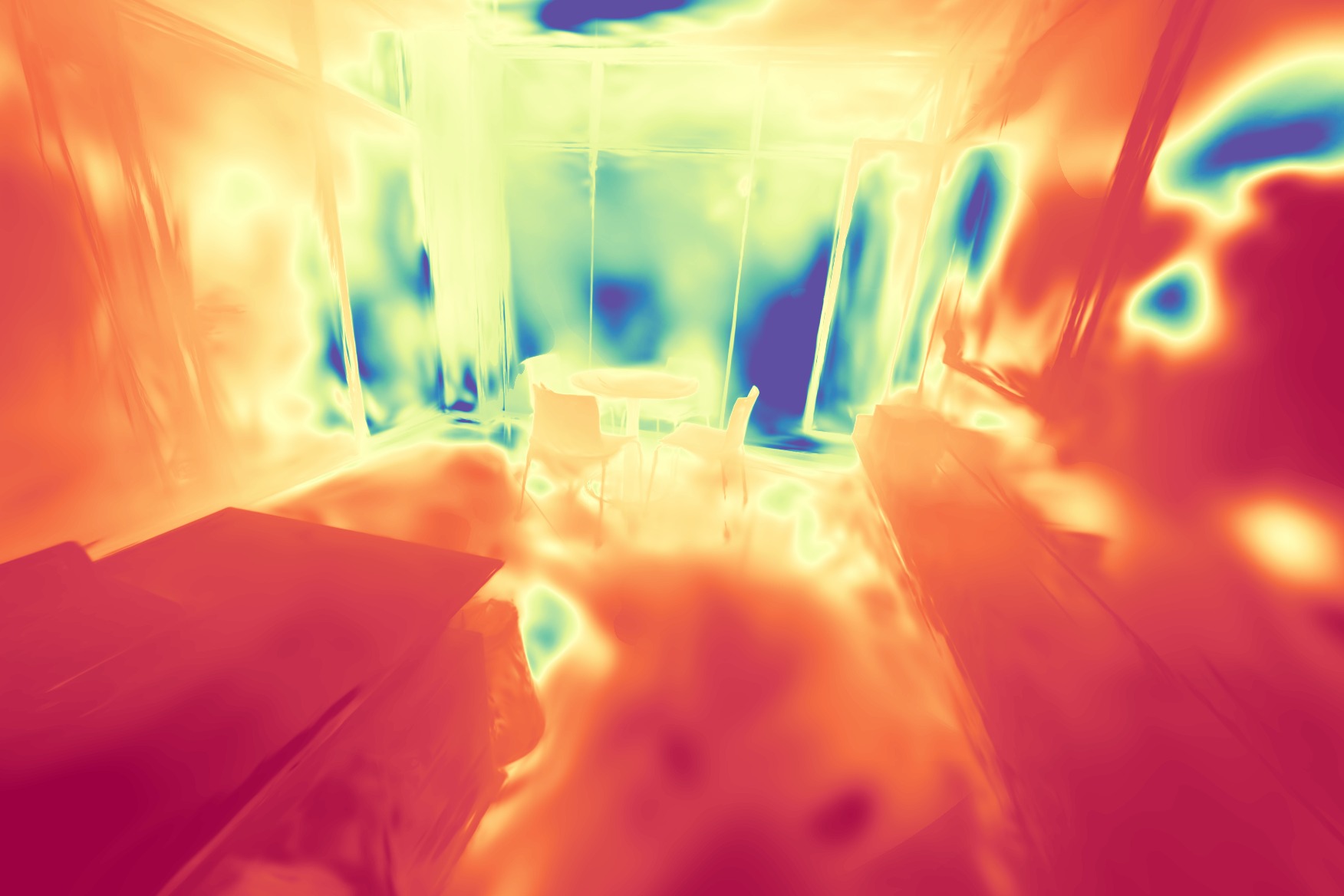} & 
\includegraphics[width=.19\linewidth]{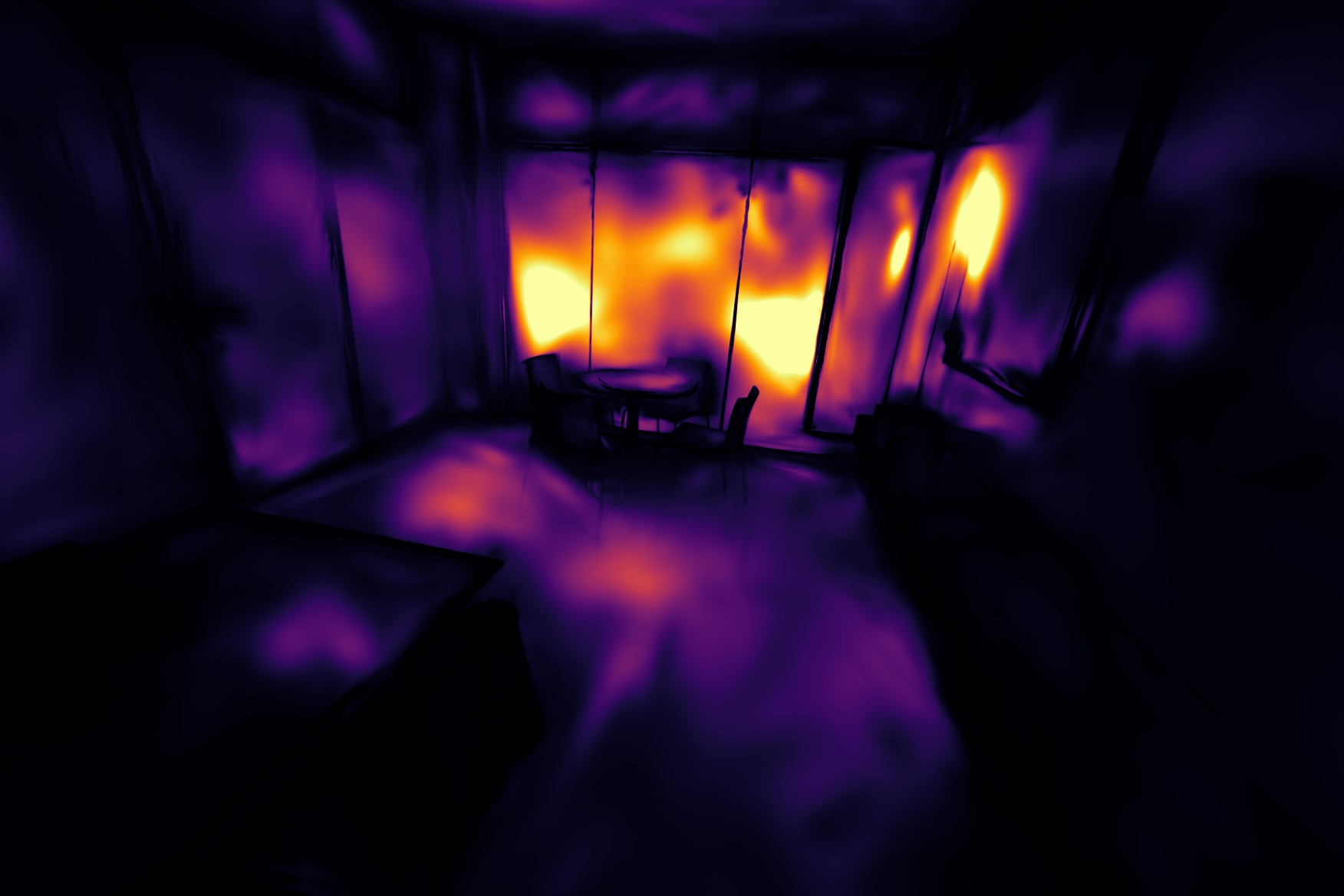} & 
\includegraphics[width=.19\linewidth]{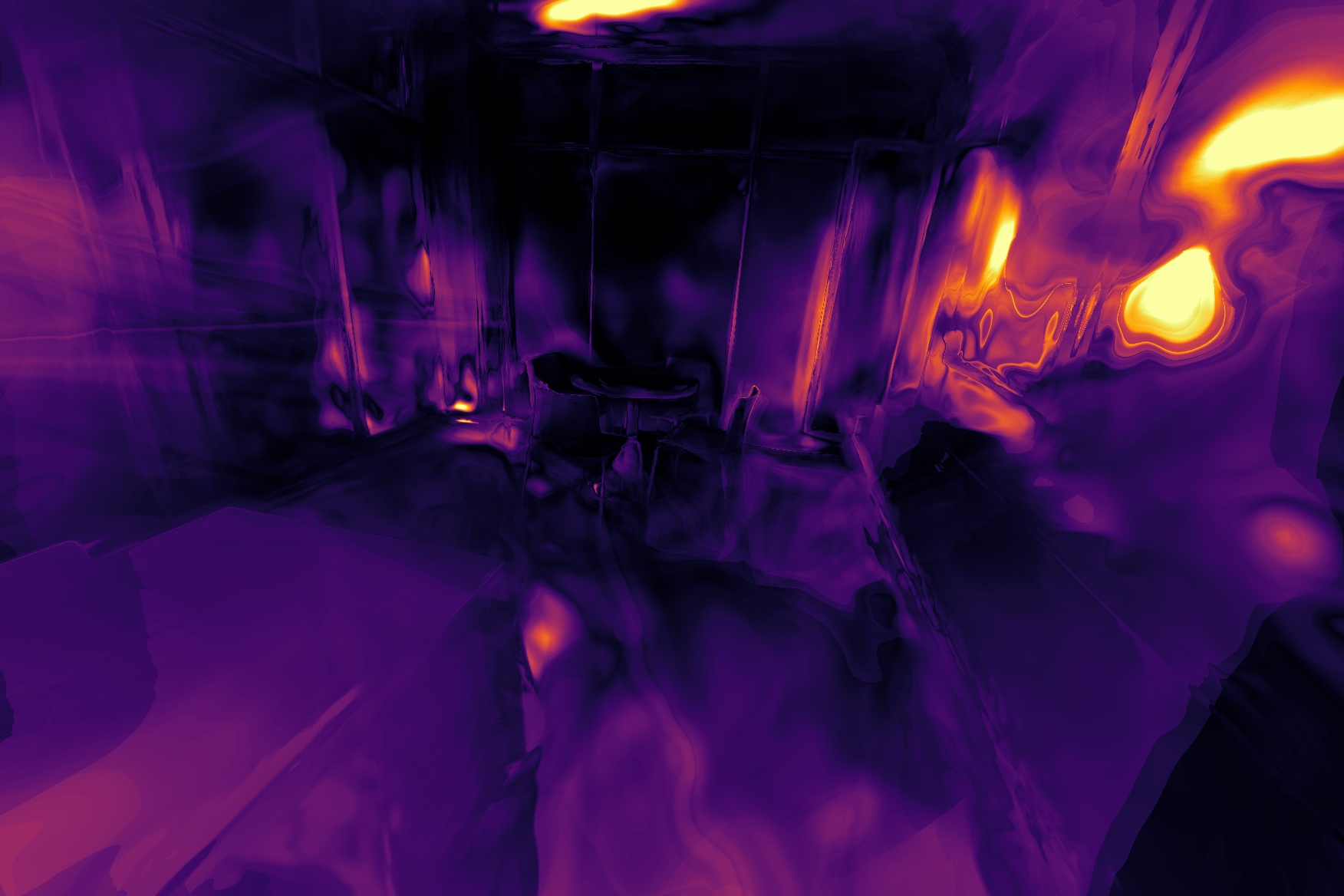} & 
\includegraphics[width=.19\linewidth]{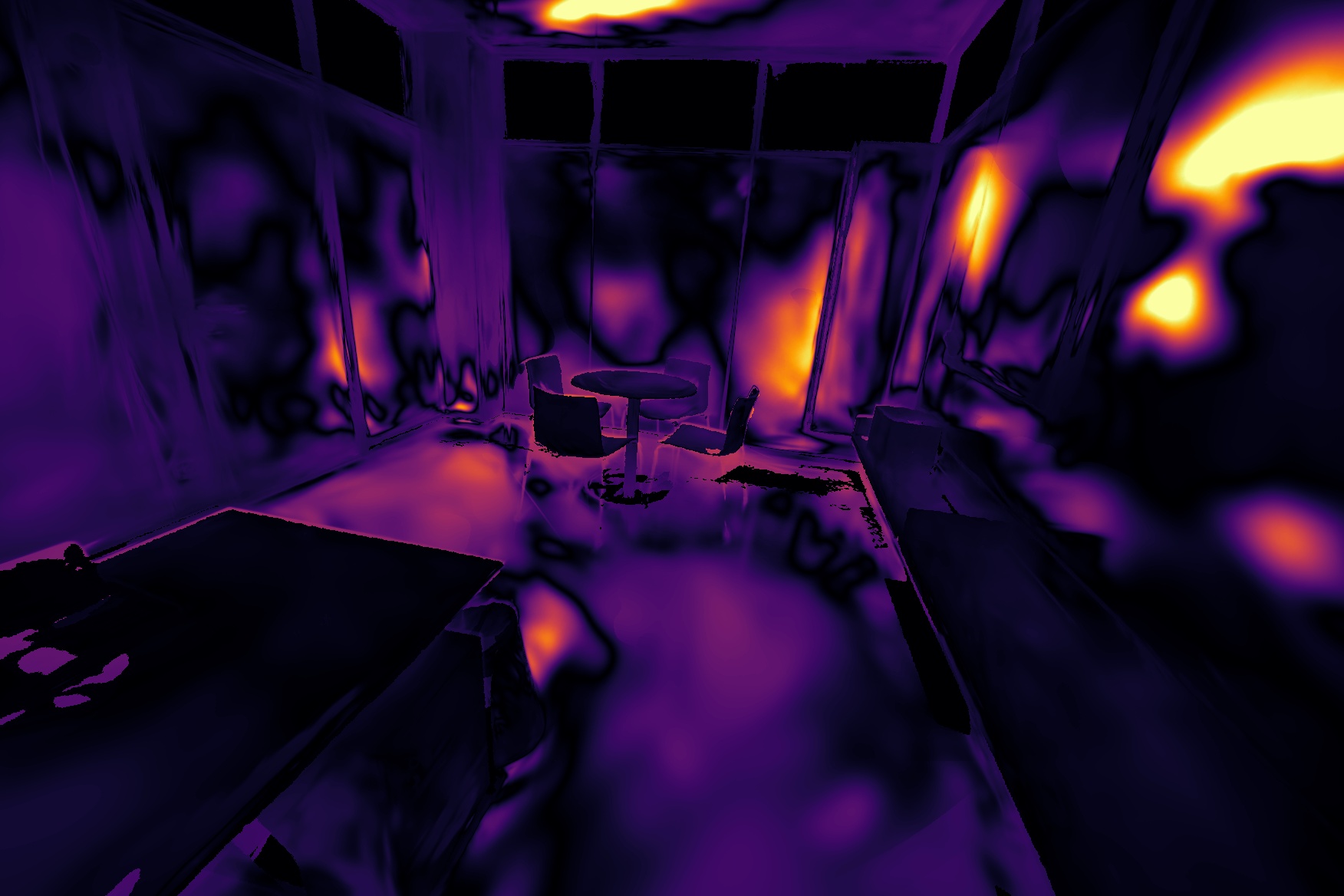} \\
\includegraphics[width=.19\linewidth]{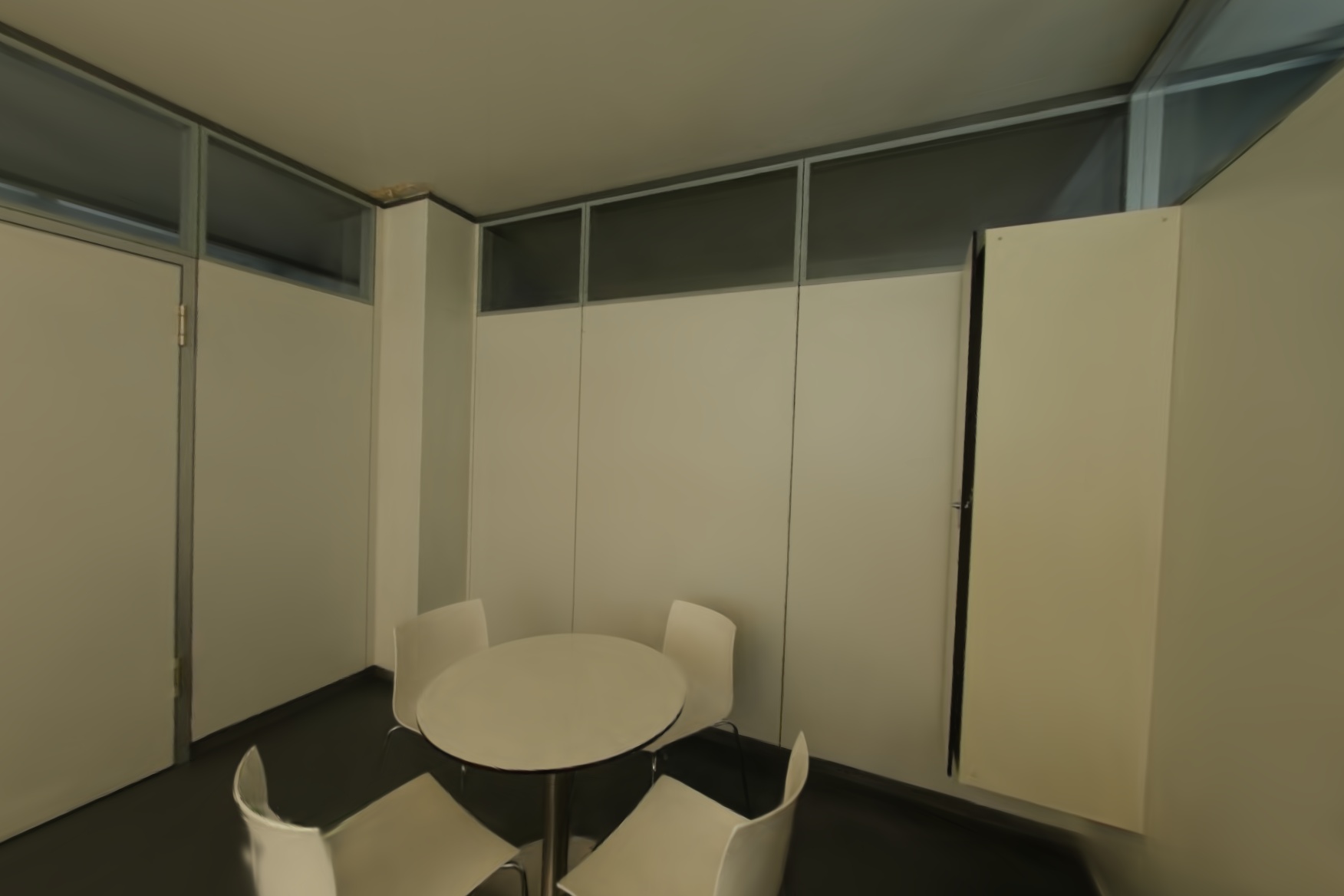} &
\includegraphics[width=.19\linewidth]{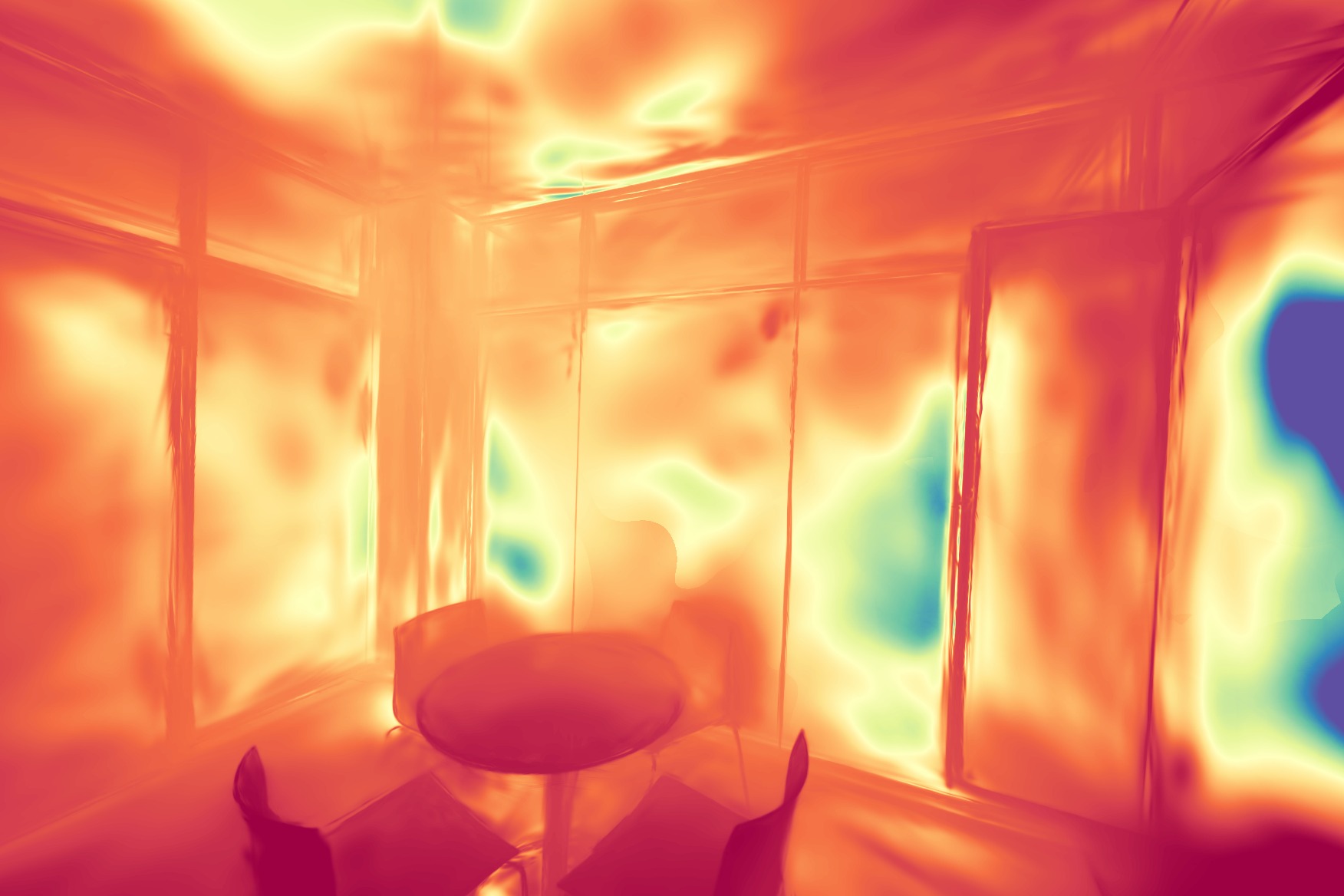}  & 
\includegraphics[width=.19\linewidth]{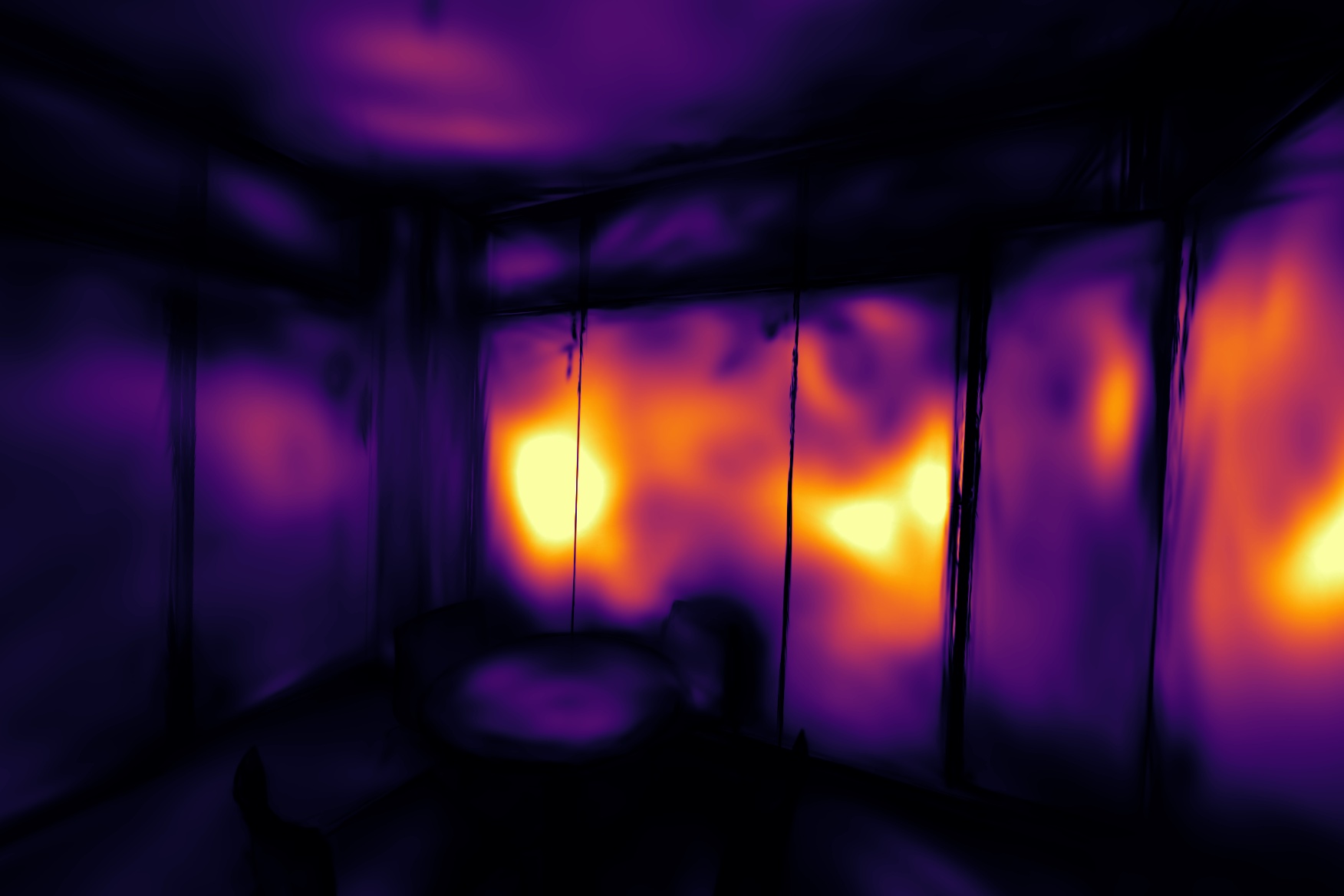} & 
\includegraphics[width=.19\linewidth]{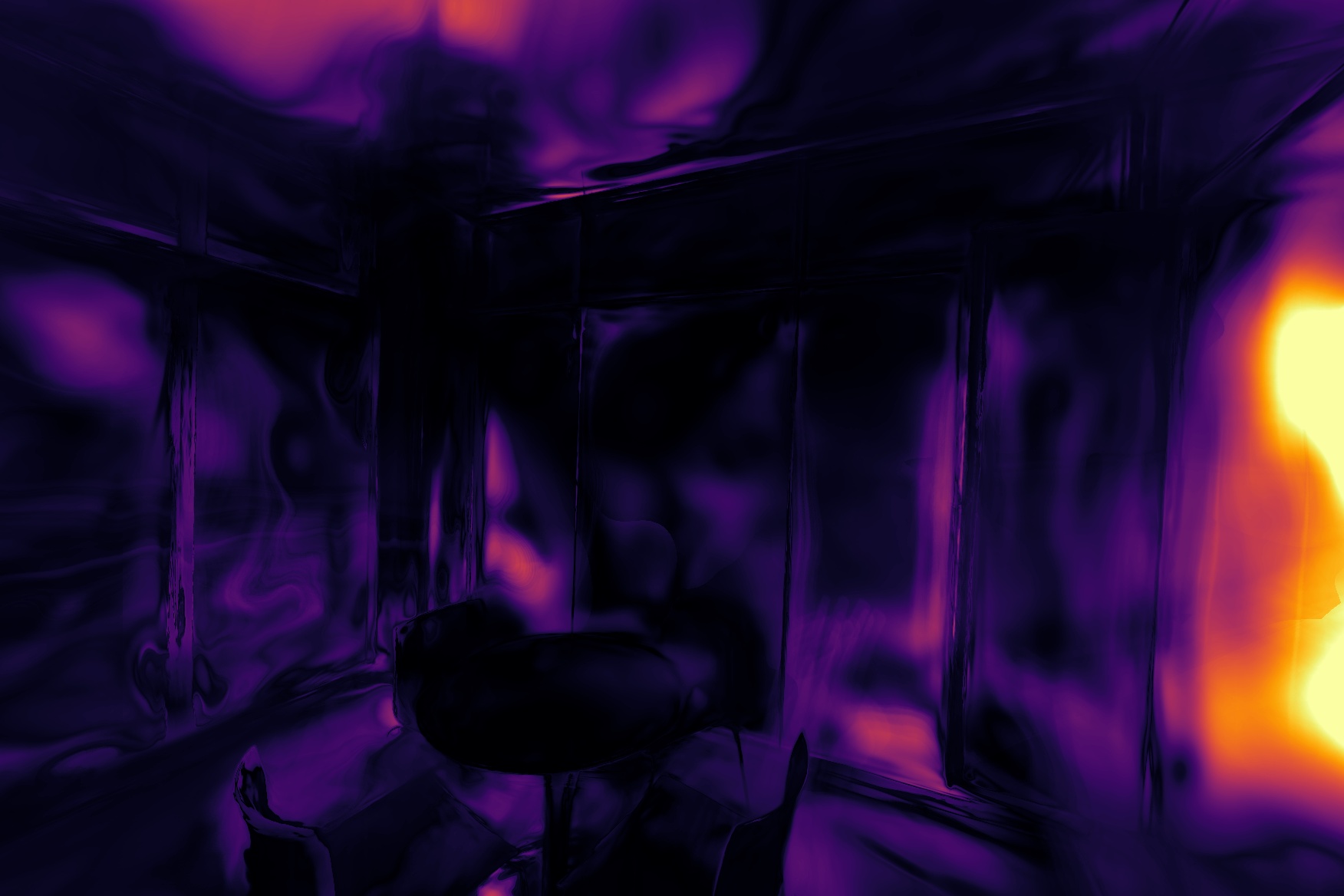} & 
\includegraphics[width=.19\linewidth]{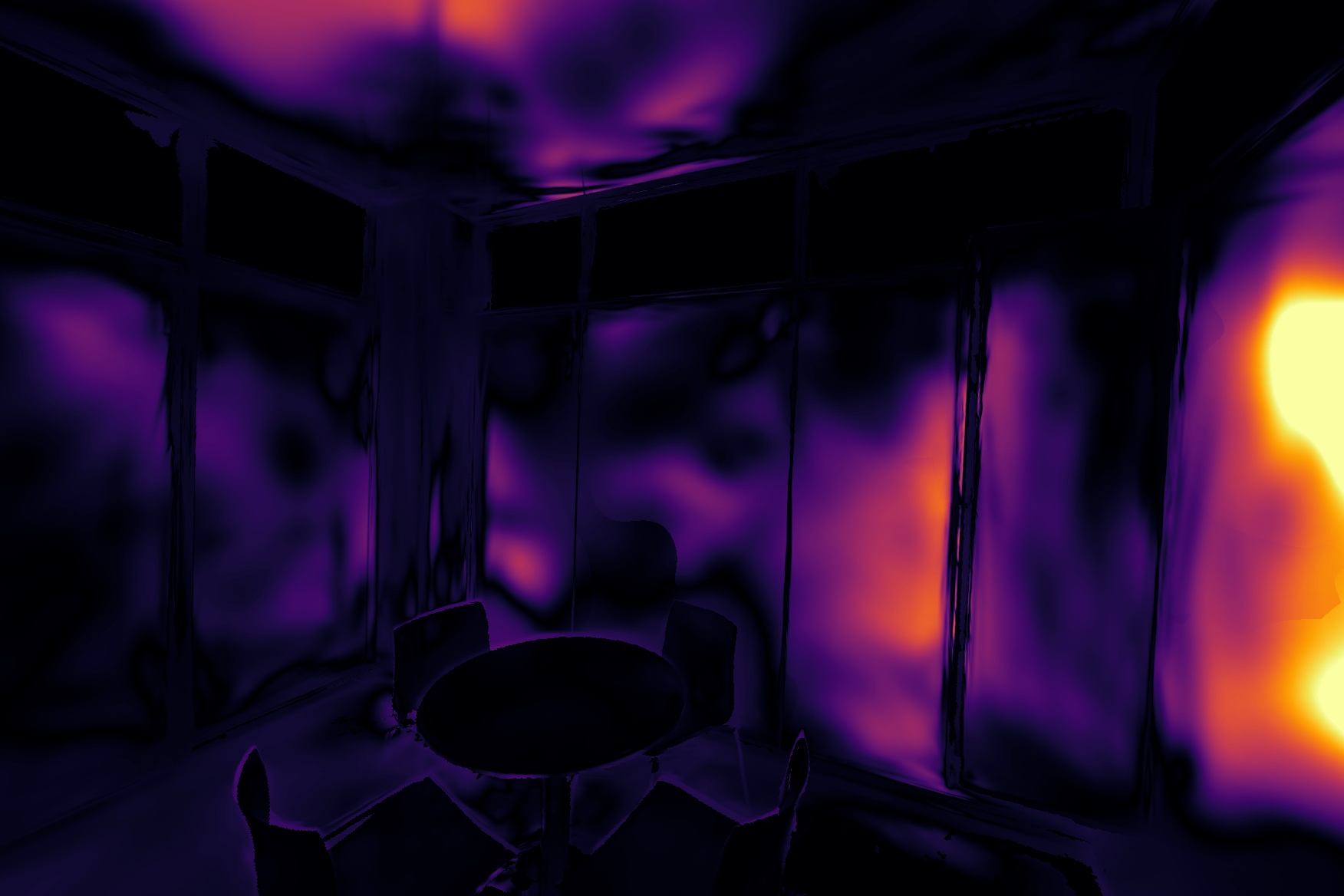} \\
\end{tabular}\vspace{-0.2cm}
\caption{\textbf{Uncertainty Visualization.} From left to right: (a) rendered image, (b) rendered depth, (c) FisherRF uncertainty, (d) WarpRF uncertainty, (e) depth error. 
}
 \label{fig:unc}
\end{figure}

\section{Experiments}
\label{sec:exp}

In our evaluation, we first measure the effectiveness of WarpRF at quantifying the uncertainty in rendered depth maps in  Sec. \ref{sec:ause}, following standard protocols in the literature. We then apply WarpRF to two downstream tasks: active view selection for novel-view synthesis (Sec. \ref{sec:active_rgb}) and active mapping for surface reconstruction (Sec \ref{sec:active_mapping}), where we actively expand the set of training views. The former task focuses on rendering images from novel views, while the latter aims at accurate 3D mesh reconstruction. 

We conduct experiments with two main radiance field frameworks, NeRF \cite{mildenhall2020nerf} and 3DGS \cite{kerbl3Dgaussians}, while also demonstrating WarpRF's generality by applying it to the brand-new SVRaster framework \cite{Sun2024SVR}. Throughout our experiments, we highlight the \colorbox{First}{best}, the \colorbox{Second}{second-best}, and the \colorbox{Third}{third-best} results.

\subsection{Uncertainty Quantification}
\label{sec:ause}
Following previous methods, we evaluate our per-pixel uncertainty predictions as described in Sec. \ref{sec:unc_pixel}, by correlating them with the errors in the rendered depths according to the Area Under Sparsification Error (AUSE) metric. 
The evaluation consists of progressively removing pixels based on predicted uncertainty, computing Mean Absolute Depth Error between the rendered and ground-truth depth at the remaining pixels coordinates, and plotting a sparsification curve with these errors. We assess how well the predicted uncertainty correlates  with actual depth error by computing the area between the previous curve and the one obtained by removing pixels based on depth error. 
While prior work commonly employs the Light Field (LF) dataset \cite{yucer2016eLF} for this evaluation, we found its ground truth depths to be highly inaccurate. Instead, we use the ScanNet++ \cite{yeshwanth2023scannet++} and ETH3D \cite{schops2017eth3d} datasets, which provide higher-quality depth annotations. 
For the ScanNet++ dataset, we select 6 scenes, including those used in 
\cite{turkulainen2024dnsplatter, safadoust2024BMVC}, while we use all the 13 high-resolution scenes from the ETH3D dataset. Please see supplementary for details about the scenes.

As WarpRF is not limited to a specific radiance field implementation, we apply it to both NeRF and 3DGS and compare our performance with the previous state-of-the-art for frameworks. While existing methods require specific modifications to the training procedure, WarpRF does not, allowing us to directly apply our method on radiance fields trained according to their original specifications.
In Table \ref{tab:ause}, we report the results achieved by BayesRays \cite{goli2023} on NeRF and by WarpRF applied to the very same trained model, while we compare with FisherRF \cite{Jiang2024FisherRF} on 3DGS to which we also apply WarpRF. We include Manifold \cite{lyu2024manifold} as a further reference.
As shown in the table, our method consistently outperforms previous approaches in both cases. 

Fig. \ref{fig:unc} shows qualitative comparisons between FisherRF (c) and WarpRF (d) uncertainties: we can observe that our uncertainty estimation is considerably closer to the actual depth error (e).

\begin{table}[]
\centering
\setlength{\tabcolsep}{4pt}
\begin{tabular}{l|lll}
\toprule
Method             & PSNR $\uparrow$   & SSIM $\uparrow$   & LPIPS $\downarrow$  \\ \midrule
3DGS + Random*      & 17.914 & 0.564-  & 0.430- \\
3DGS + ActiveNerf* \cite{pan2022activenerf} & 17.889 & 0.5326 & 0.4142 \\
3DGS + Manifold \cite{lyu2024manifold}   & \snd 20.654 & \snd 0.6163 & \fst 0.3478 \\
3DGS + FisherRF \cite{Jiang2024FisherRF}   & \trd 20.295 & \trd 0.6015 & \trd 0.3595 \\
\bf 3DGS + WarpRF (ours)        & \fst 20.715 & \fst 0.6174 & \snd 0.3487 \\ \bottomrule
\end{tabular}\vspace{-0.2cm}
\caption{\textbf{Novel view synthesis results with active view selection on Mip-NeRF360}. *: Numbers taken from \cite{Jiang2024FisherRF}.}
\label{tab:active_mip}
\end{table}
\begin{table}[]
\centering
\scalebox{0.9}{
\begin{tabular}{l|lll}
\toprule
Method                 & PSNR $\uparrow$   & SSIM $\uparrow$   & LPIPS $\downarrow$  \\ \midrule
NeRF + ActiveNeRF* \cite{pan2022activenerf}            & 26.240 & 0.8560 & 0.1240 \\ \hline
3DGS + Random*          & \trd 28.732 & \trd 0.9389 & \trd 0.0534 \\
3DGS + ActiveNeRF* \cite{pan2022activenerf}     & 25.854 & 0.9157 & 0.0766 \\ 
3DGS + Manifold \cite{lyu2024manifold}       & 28.148 & 0.9352 & 0.0567 \\
3DGS + FisherRF \cite{Jiang2024FisherRF}       & \snd 29.262 & \snd 0.9425 & \snd 0.0501 \\

\bf 3DGS + WarpRF (ours)            & \fst 29.269 & \fst 0.9442 & \fst 0.0482 \\ \midrule
SVRaster + Random & 26.382 & 0.9053 & 0.0879  \\
SVRaster + Farthest & 26.555 & 0.9117 & 0.0790  \\
\bf SVRaster + WarpRF (ours) & \fst 27.410 & \fst \fst 0.9201 & \fst 0.0730  \\ \bottomrule
\end{tabular}}\vspace{-0.2cm}
\caption{\textbf{Novel view synthesis results with active view selection on NeRF Synthetic with 20 views}. *: Numbers taken from \cite{Jiang2024FisherRF}.}
\label{tab:active_nerf_20}
\end{table}

\subsection{Active View Selection for NVS}
\label{sec:active_rgb}
We now apply WarpRF to active view selection for training radiance fields for novel view synthesis. Since our approach to quantifying uncertainty as described in Sec. \ref{sec:unc_image} is independent of the underlying 3D representation, we apply our method to both 3DGS and the very recent SVRaster \cite{Sun2024SVR}. 

Training begins with a limited set of initial views, and during the training, we actively select the next best views from a pool of candidates to be added to the training set. The selection criterion is based on maximizing uncertainty as defined in Sec. \ref{sec:unc_image}, ensuring that each added view contributes with the most information.
Once training is complete, we evaluate the model using standard novel view synthesis on held-out test views.

\begin{table}[]
\centering
\setlength{\tabcolsep}{3pt}
\begin{tabular}{l|ccc}
\toprule
Method                 & PSNR $\uparrow$   & SSIM $\uparrow$  & LPIPS $\downarrow$ \\ \midrule
3DGS + Random*          & 20.670  & 0.8242 & 0.2049 \\
3DGS + ActiveNeRF* \cite{pan2022activenerf}      & 22.979  & 0.8756 & 0.1109 \\
3DGS + Manifold \cite{lyu2024manifold}        & \trd 24.159  & \trd 0.8912 & \trd 0.0942 \\
3DGS + FisherRF \cite{Jiang2024FisherRF}        & \snd 24.943  & \snd 0.8962 & \snd 0.0867 \\
\bf 3DGS + WarpRF (ours)             & \fst 24.957 & \fst 0.9002 & \fst 0.0845 \\ \bottomrule
\end{tabular}\vspace{-0.2cm}
\caption{\textbf{Novel view synthesis results with active view selection on NeRF Synthetic with 10 views}. *: Numbers taken from \cite{Jiang2024FisherRF}.}
\label{tab:active_nerf_10}
\end{table}

\begin{table}[]
\centering
\scalebox{0.9}{
\begin{tabular}{l|lll}
\toprule
Method                 & PSNR $\uparrow$   & SSIM $\uparrow$  & LPIPS $\downarrow$ \\ \midrule
SVRaster + Random     & 16.633 & 0.5738 & 0.3578 \\
SVRaster + Farthest & 17.474 & 0.6117 & 0.3264 \\
\bf SVRaster + WarpRF (ours) & \fst 17.774 & \fst 0.6244 & \fst 0.3178 \\ \bottomrule
\end{tabular}}\vspace{-0.2cm}
\caption{\textbf{Novel view synthesis results with active view selection on Tanks and Tamples}. Follwoing \cite{kerbl3Dgaussians, Sun2024SVR} we use the two outdoor scenes `train' and `truck'.}
\label{tab:active_tandt}
\end{table}

\begin{figure}[t]
    \centering
    \renewcommand{\tabcolsep}{1pt}
    \begin{tabular}{rcccccc}        
    \rotatebox[origin=l]{90}{Manifold} & 
        \includegraphics[width=0.08\textwidth]{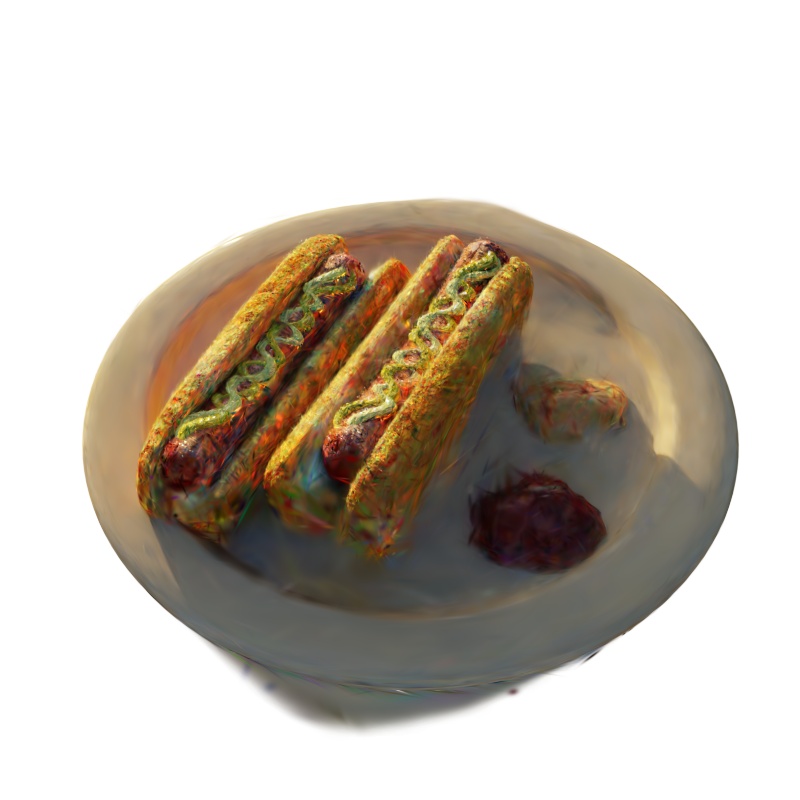} & 
        \includegraphics[width=0.08\textwidth]{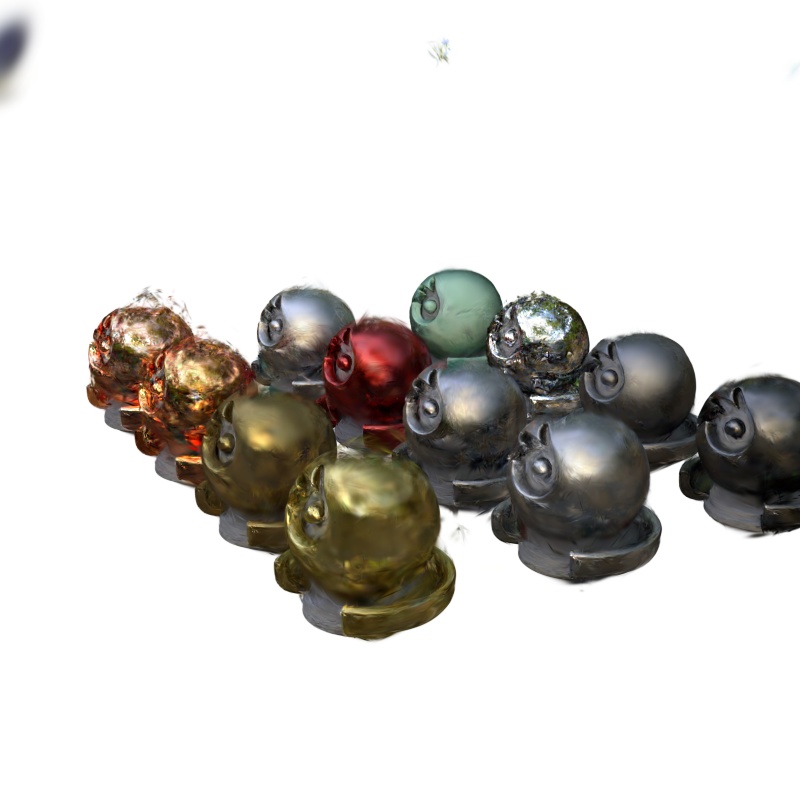} 
        & 
        \includegraphics[width=0.08\textwidth]{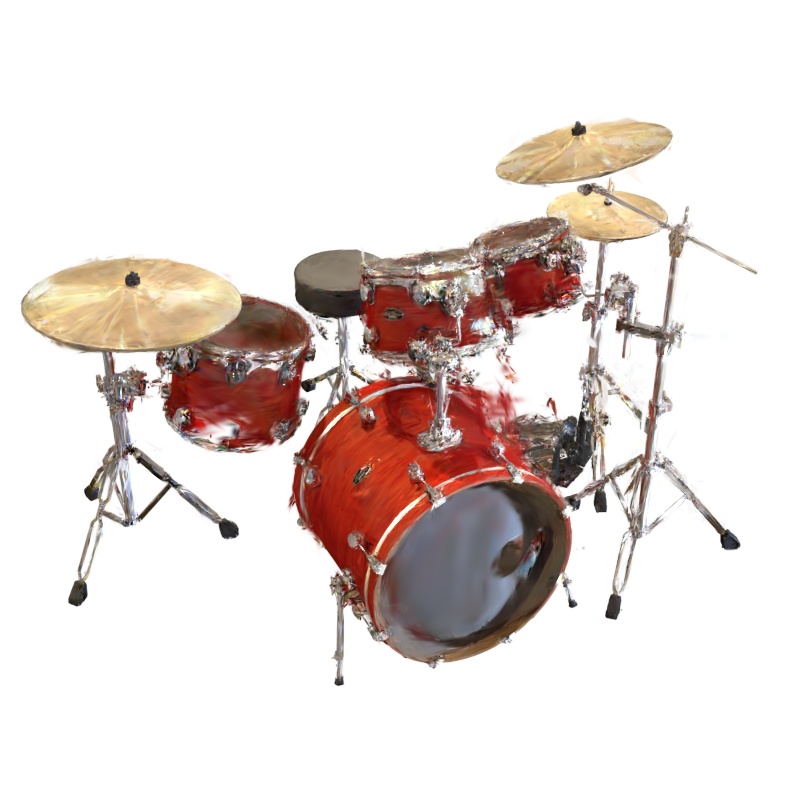}
        & 
        \includegraphics[width=0.08\textwidth]{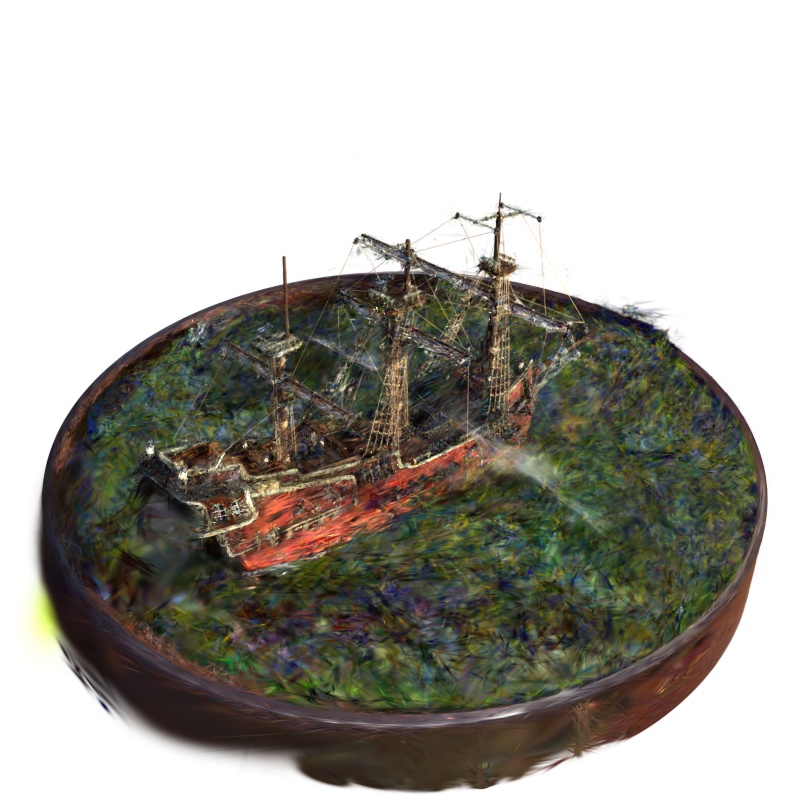} & 
        \includegraphics[width=0.08\textwidth]{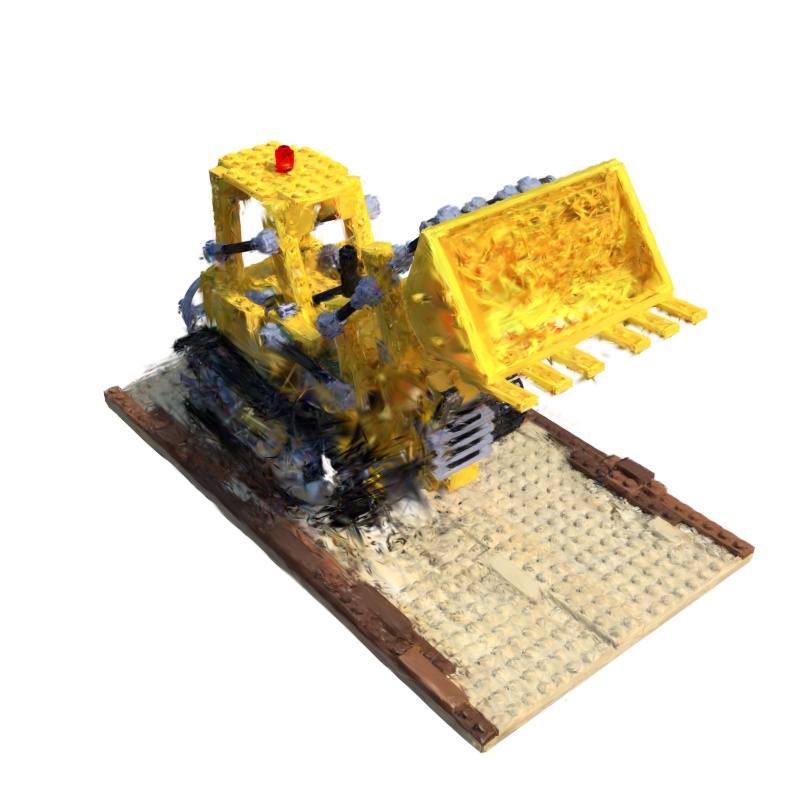}  \\
        \rotatebox[origin=l]{90}{FisherRF} &
        \includegraphics[width=0.08\textwidth]{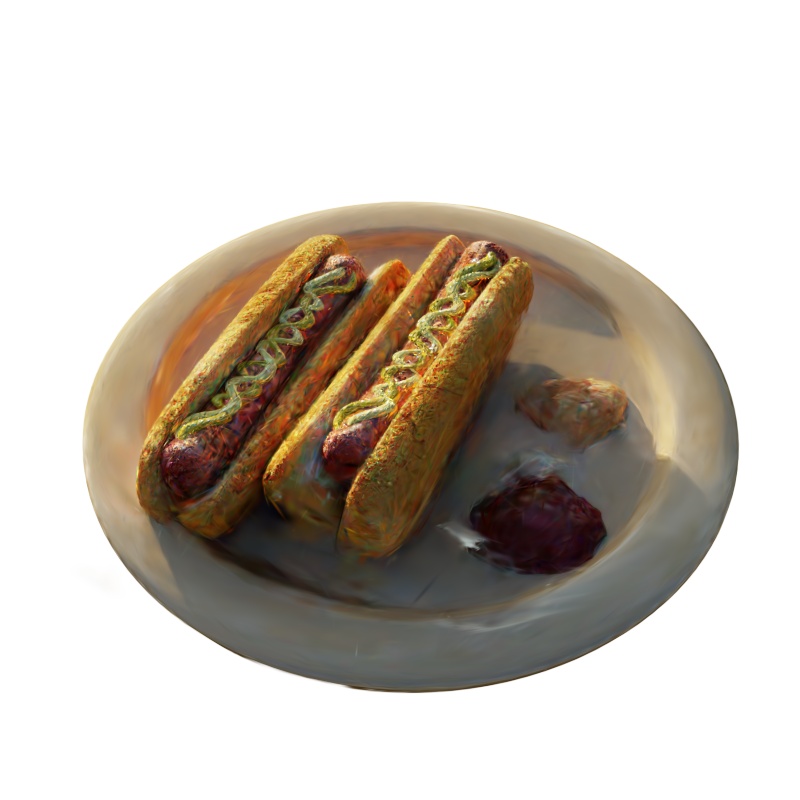} & 
        \includegraphics[width=0.08\textwidth]{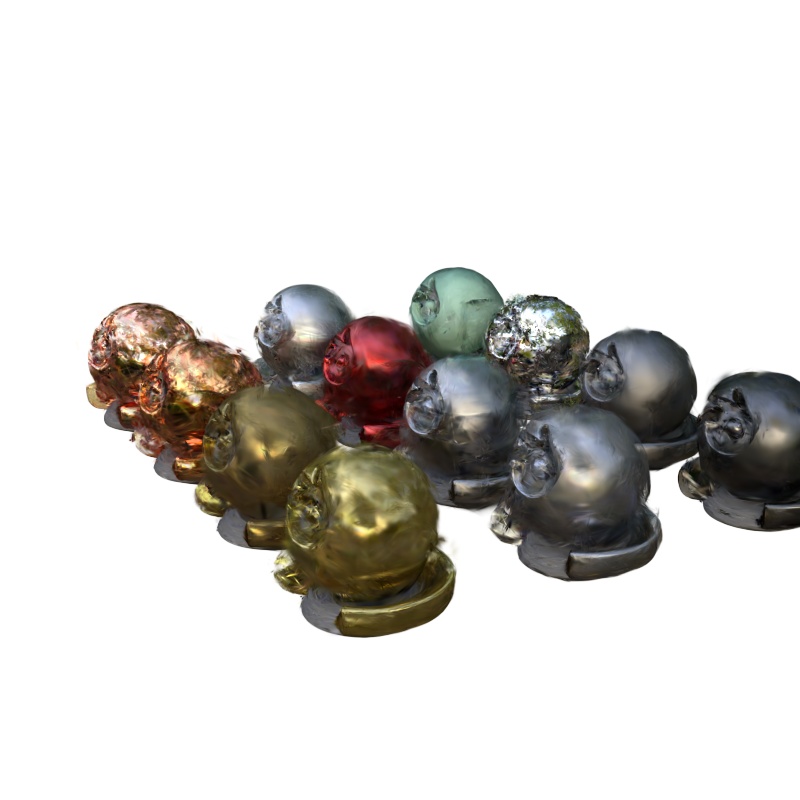} 
        & 
        \includegraphics[width=0.08\textwidth]{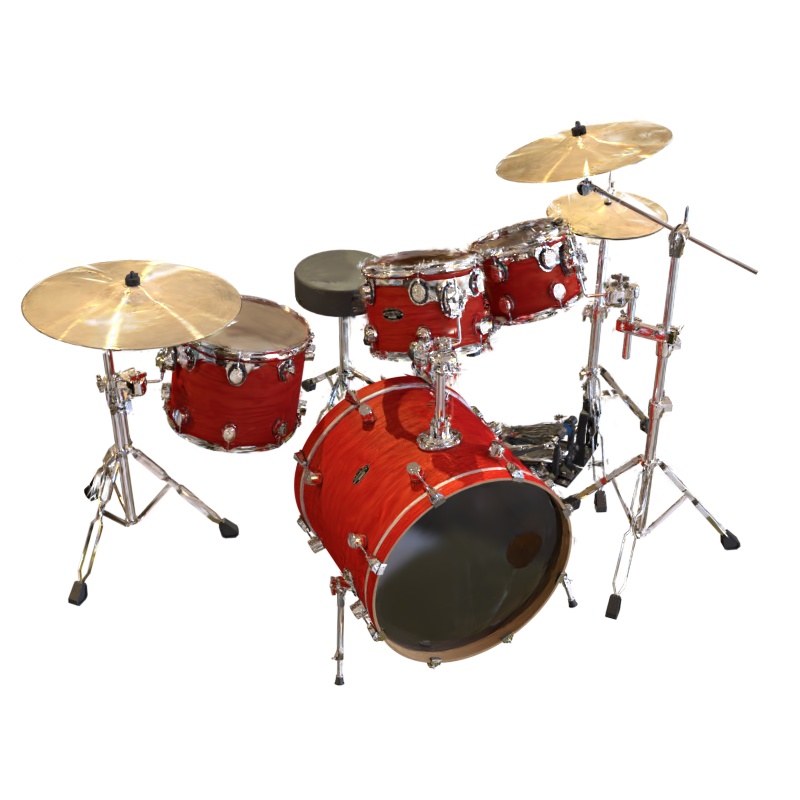}
        & 
        \includegraphics[width=0.08\textwidth]{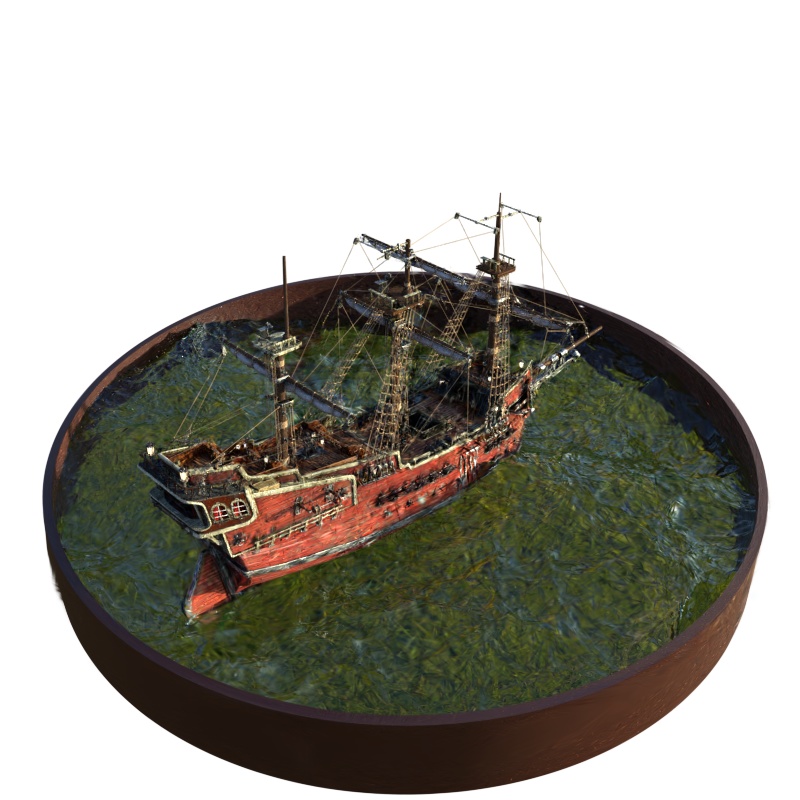} & 
        \includegraphics[width=0.08\textwidth]{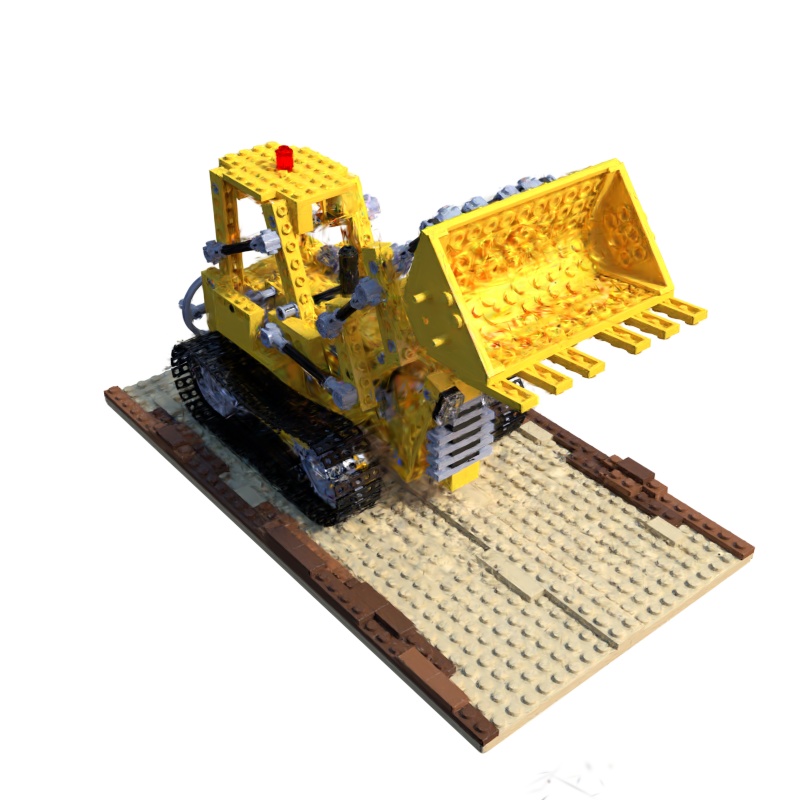}  \\
        \rotatebox[origin=l]{90}{WarpRF} &
        \includegraphics[width=0.08\textwidth]{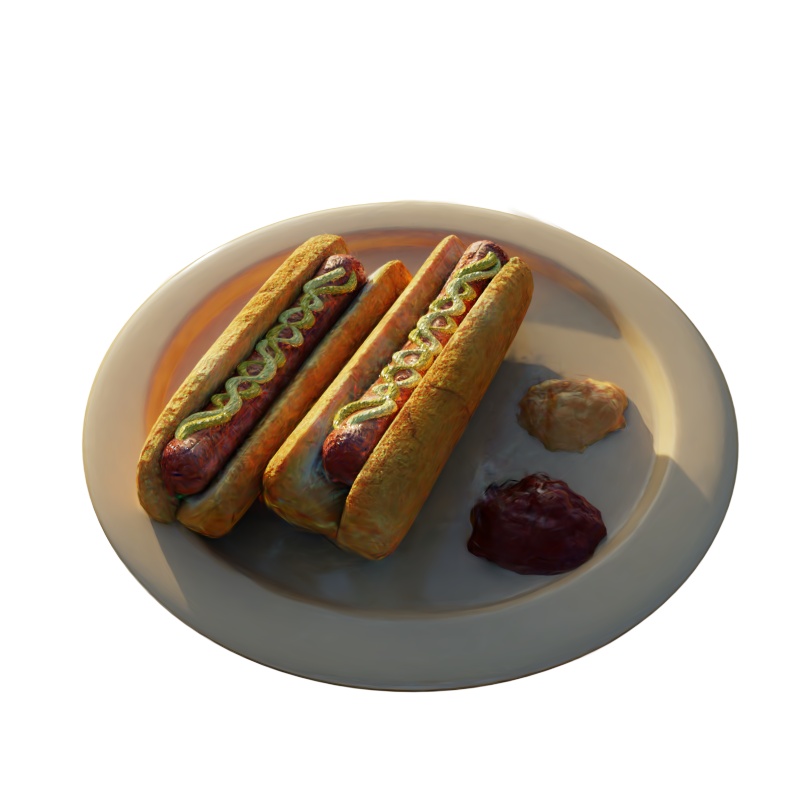} & 
        \includegraphics[width=0.08\textwidth]{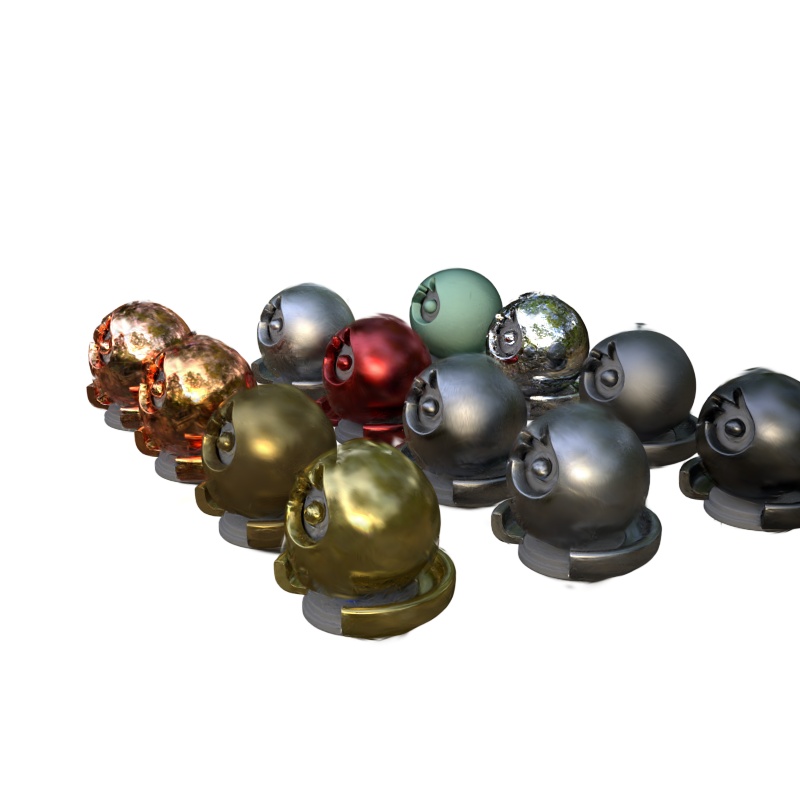} 
        & 
        \includegraphics[width=0.08\textwidth]{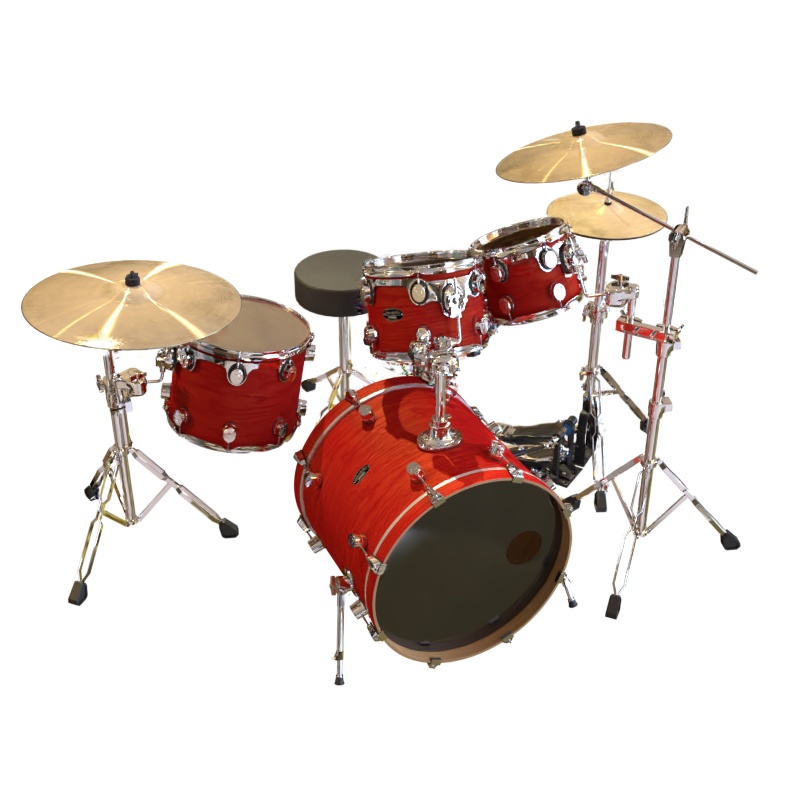}
        & 
        \includegraphics[width=0.08\textwidth]{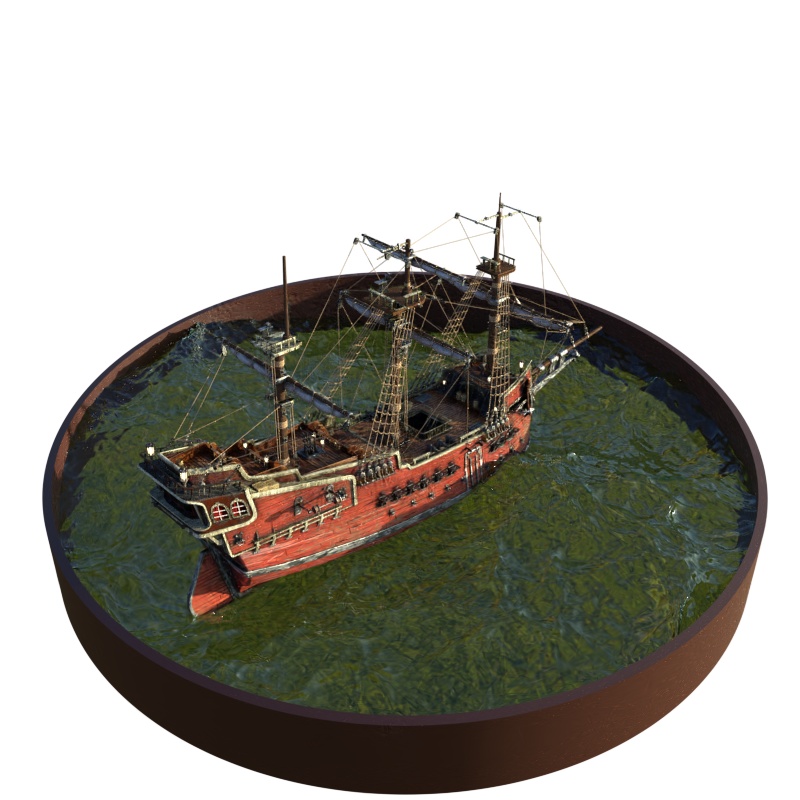} & 
        \includegraphics[width=0.08\textwidth]{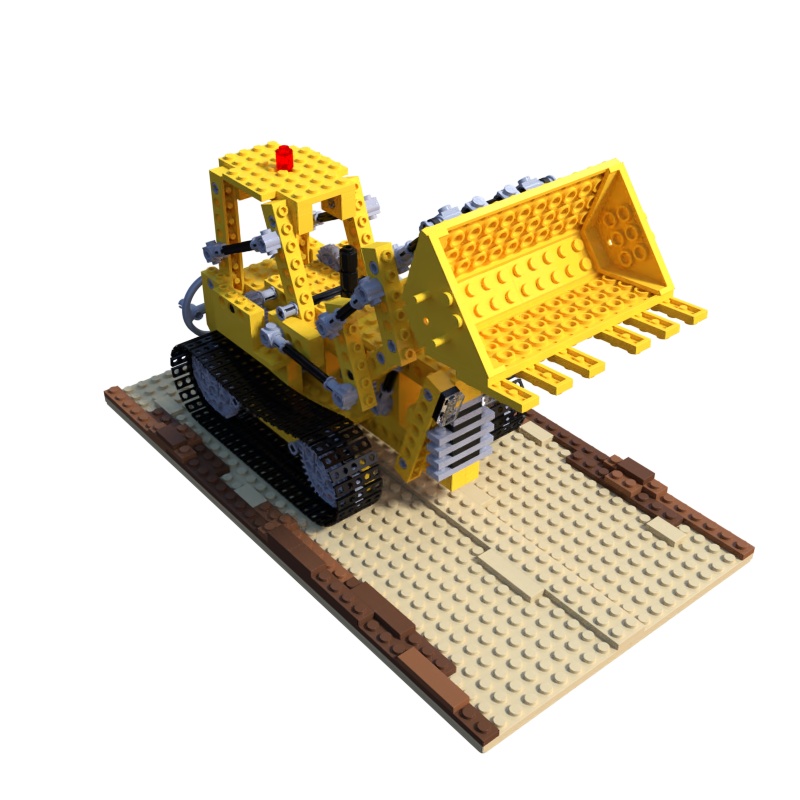}  \\
        \rotatebox[origin=l]{90}{\quad GT} &
        \includegraphics[width=0.08\textwidth]{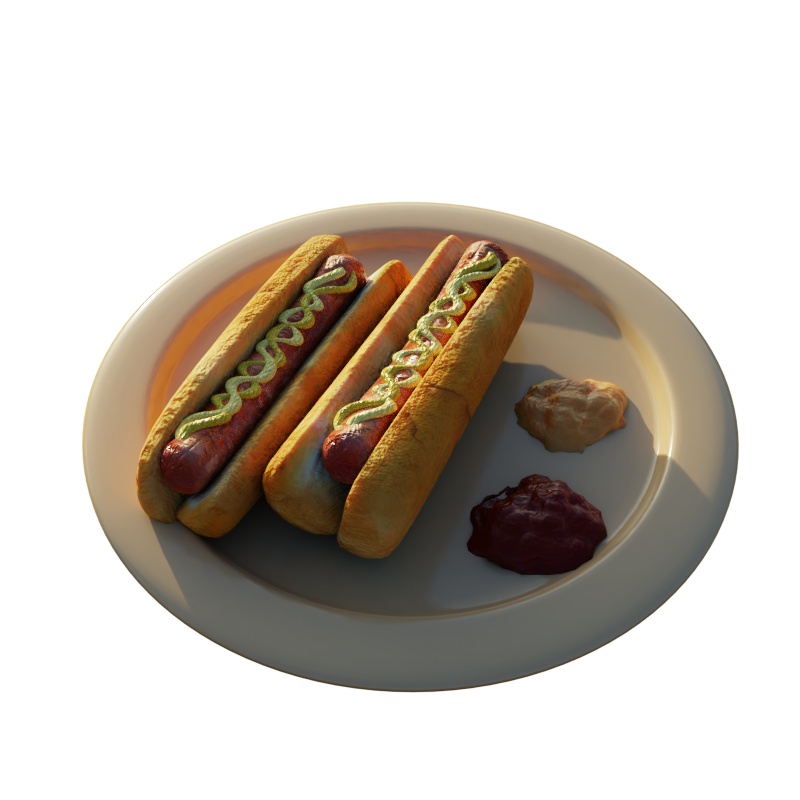} & 
        \includegraphics[width=0.08\textwidth]{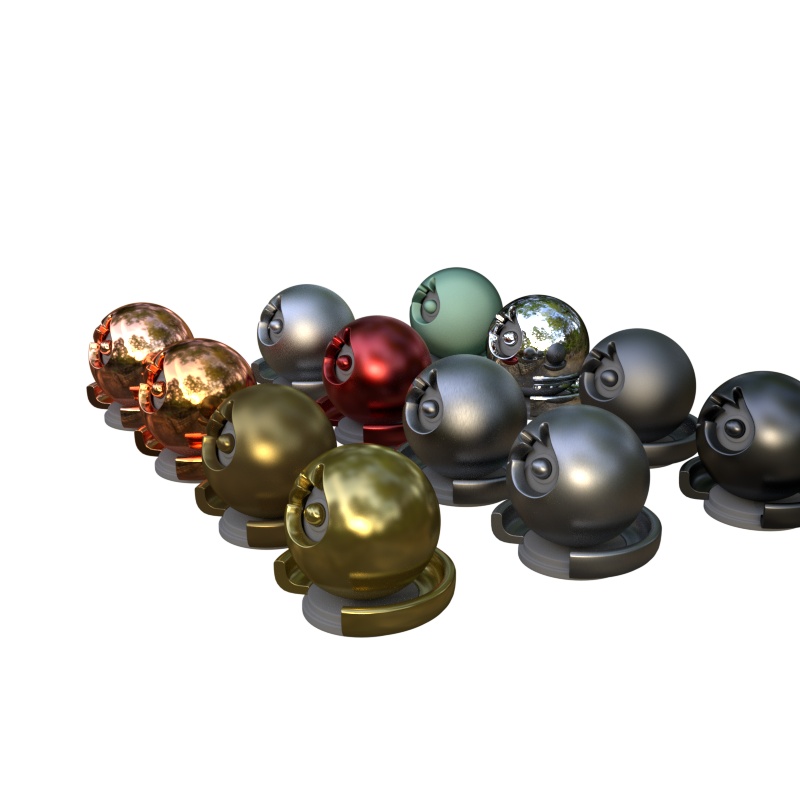} 
        & 
        \includegraphics[width=0.08\textwidth]{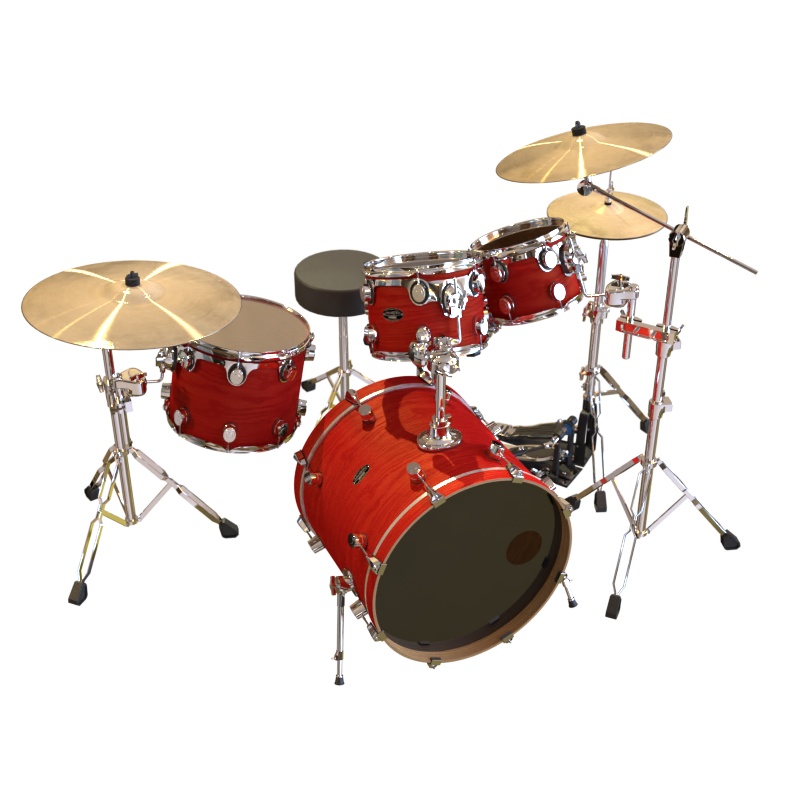}
        & 
        \includegraphics[width=0.08\textwidth]{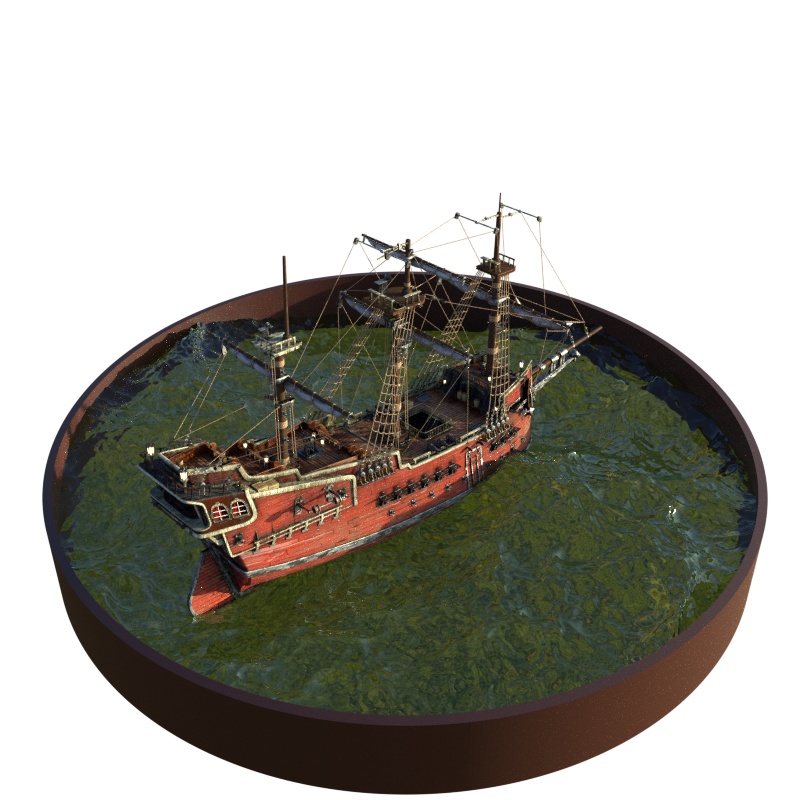} & 
        \includegraphics[width=0.08\textwidth]{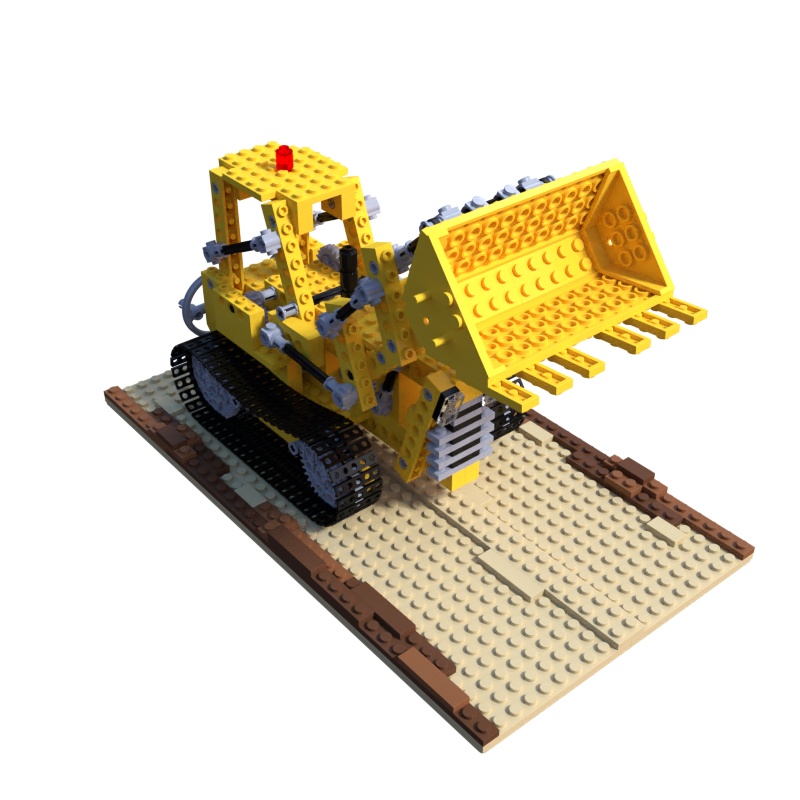}  \\
        
    \end{tabular}
    \caption{\textbf{Qualitative results of active camera selection on NeRF Synthetic dataset with 10 views.} From top to bottom, we collect rendered results by 3DGS trained with to active view selection being performed by Manifold, FIsherRF and WarpRF, followed by the real images.}
    \label{fig:qual_synth10}
\end{figure}

\boldparagraph{3DGS} We adopt the same experimental setup as in \cite{Jiang2024FisherRF} and utilize the widely adopted Mip-NeRF360 \cite{barron2022mip} and NeRF Synthetic \cite{mildenhall2020nerf} datasets. The Mip-NeRF360 dataset consists of nine real-world scenes, while the NeRF Synthetic dataset includes eight synthetically rendered scenes.
For both datasets, we use all training views as the candidate pool. Mip-NeRF360 images are resized to a width of 1600 pixels, whereas NeRF Synthetic images are used at their original $800\times800$ resolution. Training begins with four uniformly selected cameras, adding one every 100 epochs until 20 views are selected. The model is trained for 30\,000 iterations in total. To prevent training degeneration, opacity is reset each time a new view is added.
Following \cite{Jiang2024FisherRF}, we also experiment with an alternative setting, beginning with two training views and incrementally adding a camera every 100 epochs until reaching 10 views. 

\begin{figure*}[t]
    \centering
    \renewcommand{\tabcolsep}{1pt}
    \begin{tabular}{rcccccc}        
    \rotatebox[origin=l]{90}{Manifold \cite{lyu2024manifold}} & 
        \includegraphics[width=0.19\textwidth, height=0.127\textwidth]{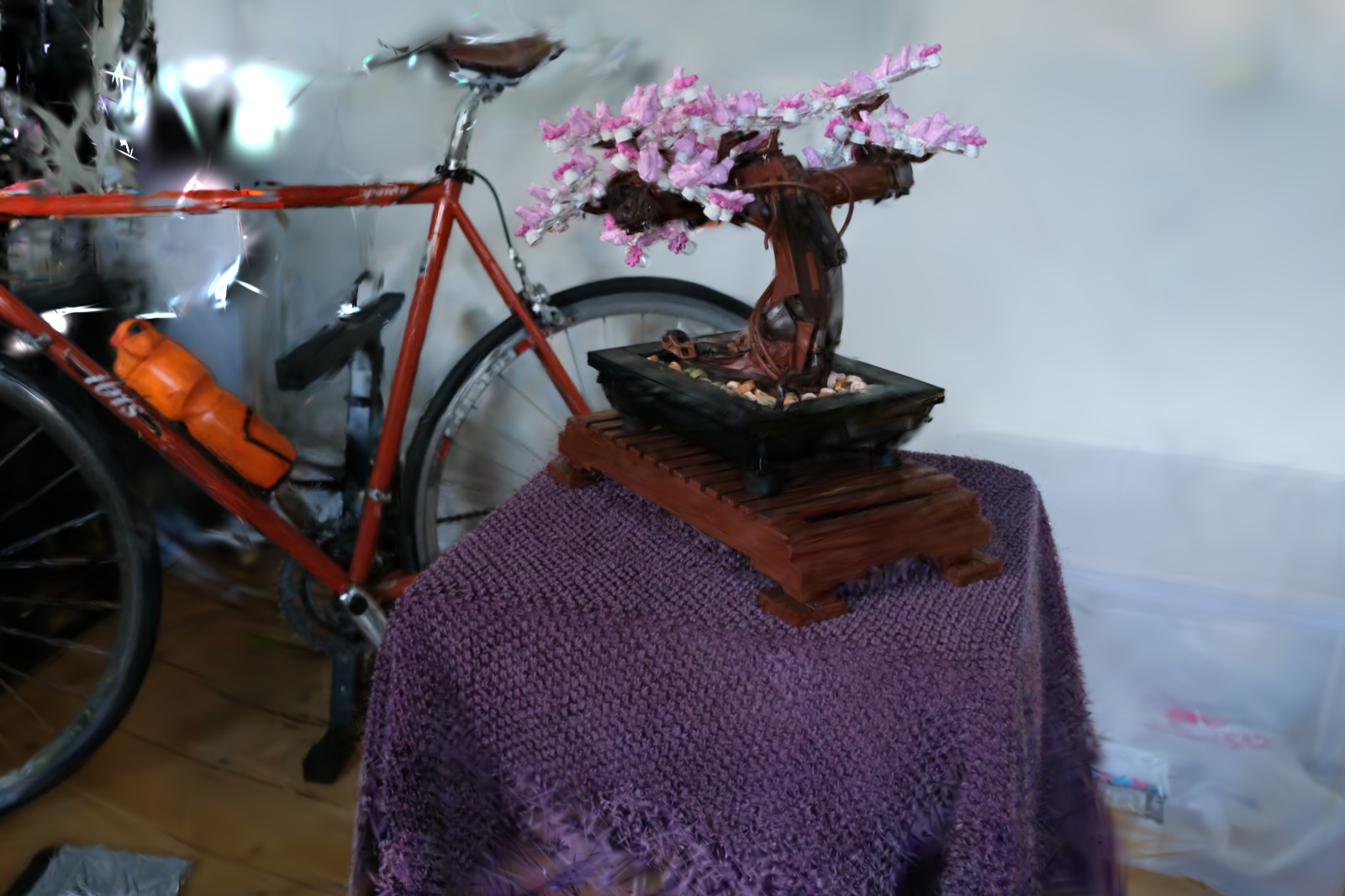} & 
        \includegraphics[width=0.19\textwidth, height=0.127\textwidth]{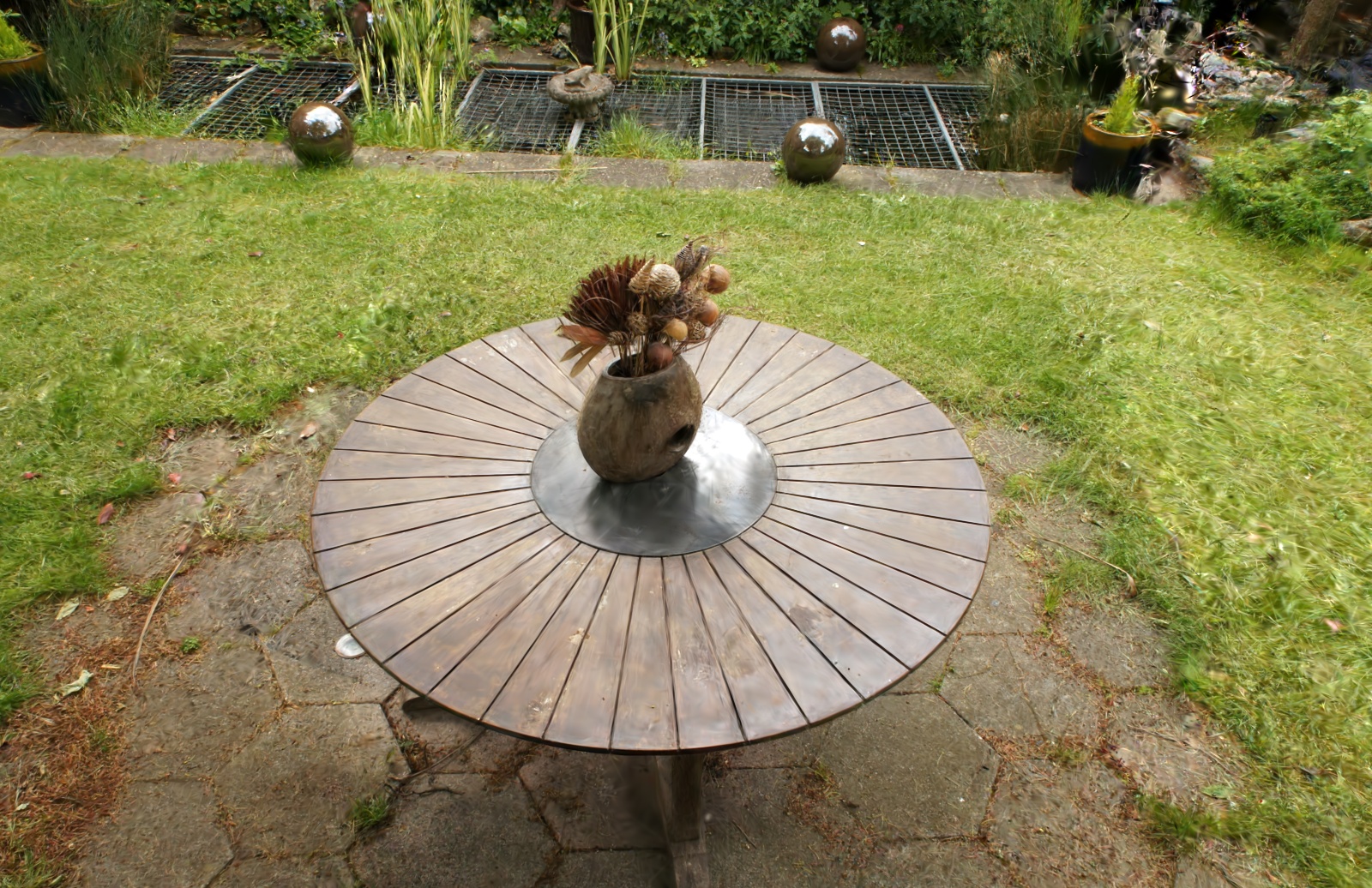} 
        & 
        \includegraphics[width=0.19\textwidth, height=0.127\textwidth]{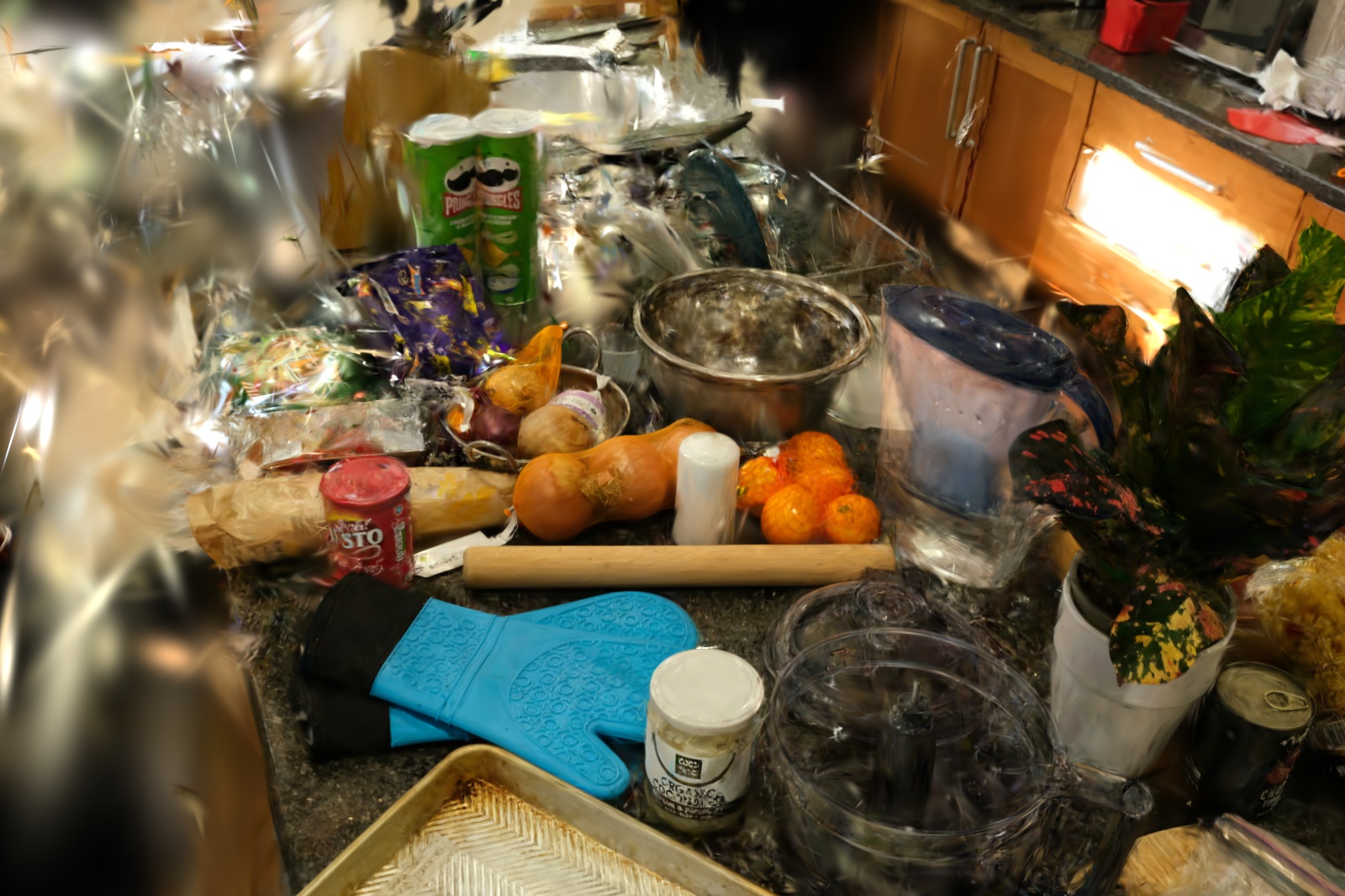}
        & 
        \includegraphics[width=0.19\textwidth, height=0.127\textwidth]{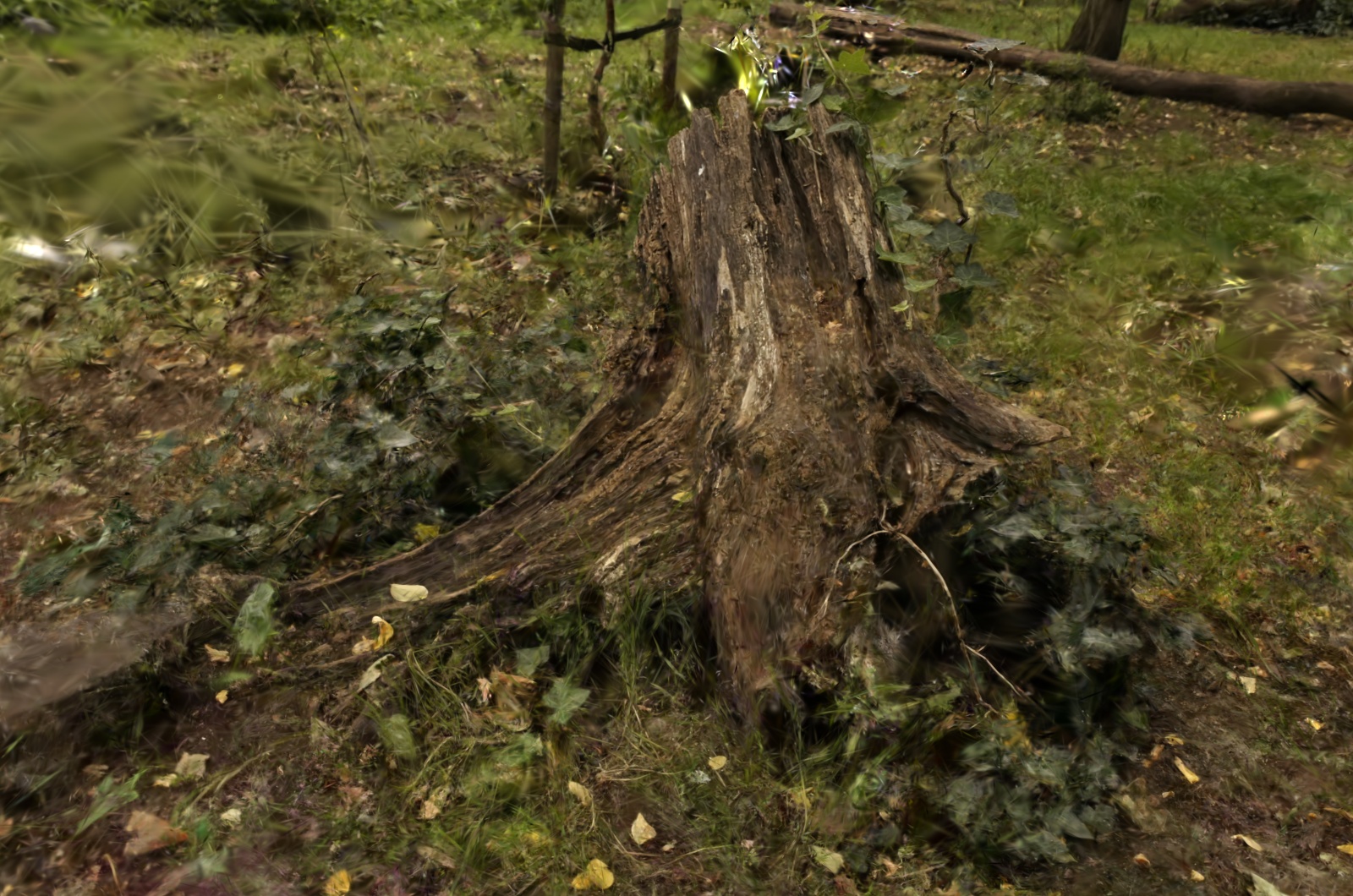} & 
        \includegraphics[width=0.19\textwidth, height=0.127\textwidth]{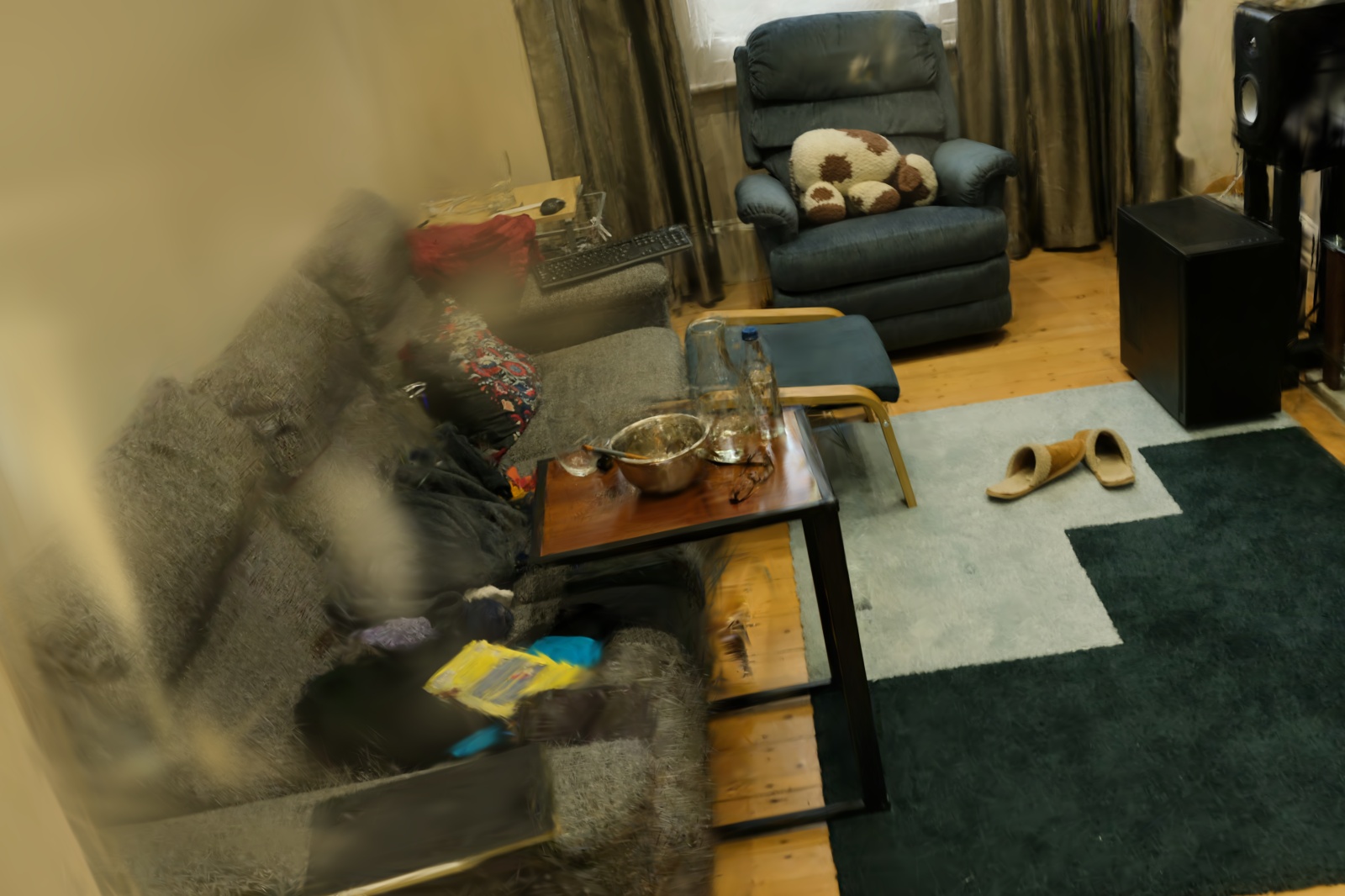}  \\
        \rotatebox[origin=l]{90}{FisherRF \cite{Jiang2024FisherRF}} &
        \includegraphics[width=0.19\textwidth, height=0.127\textwidth]{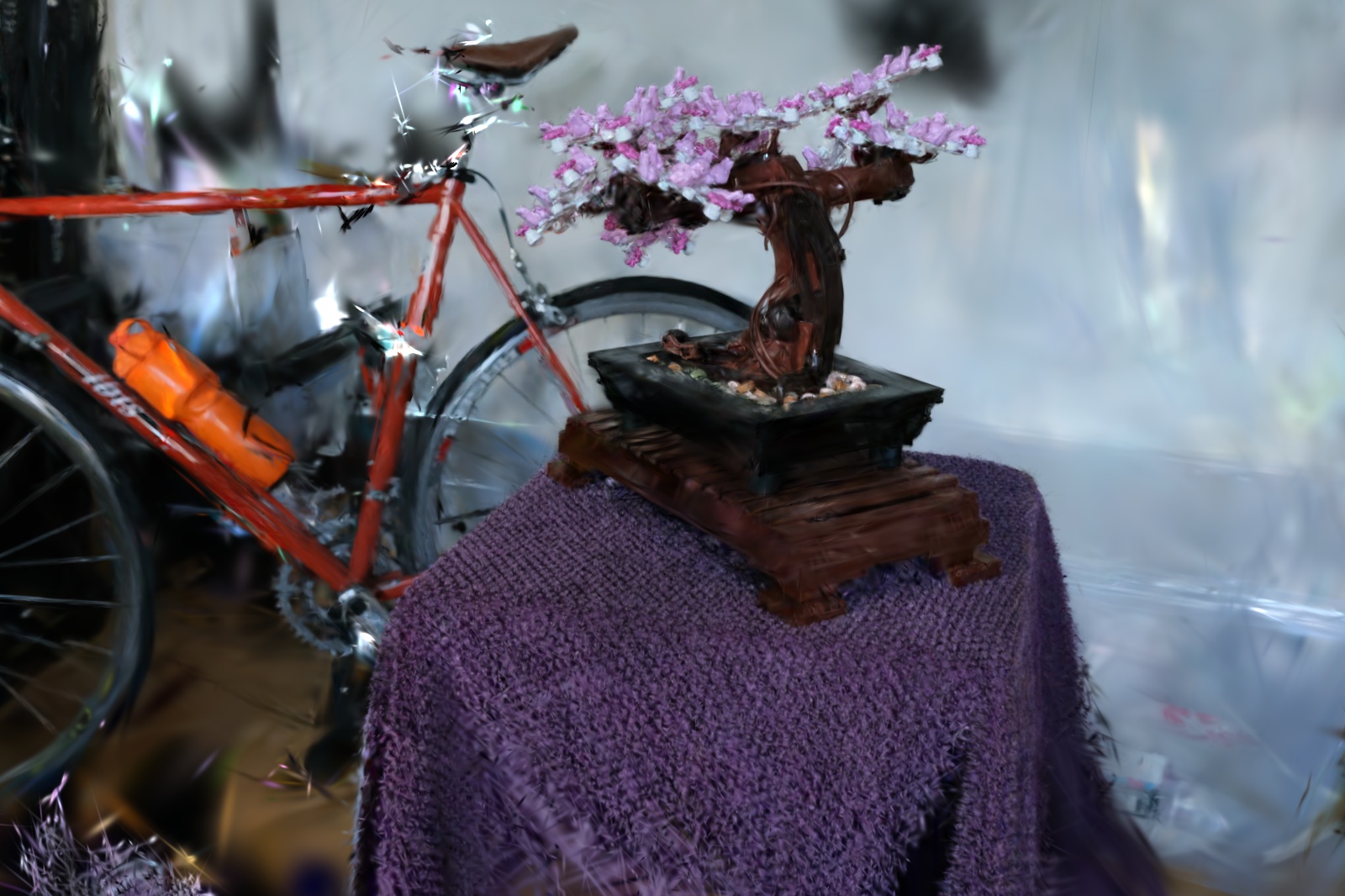} & 
        \includegraphics[width=0.19\textwidth, height=0.127\textwidth]{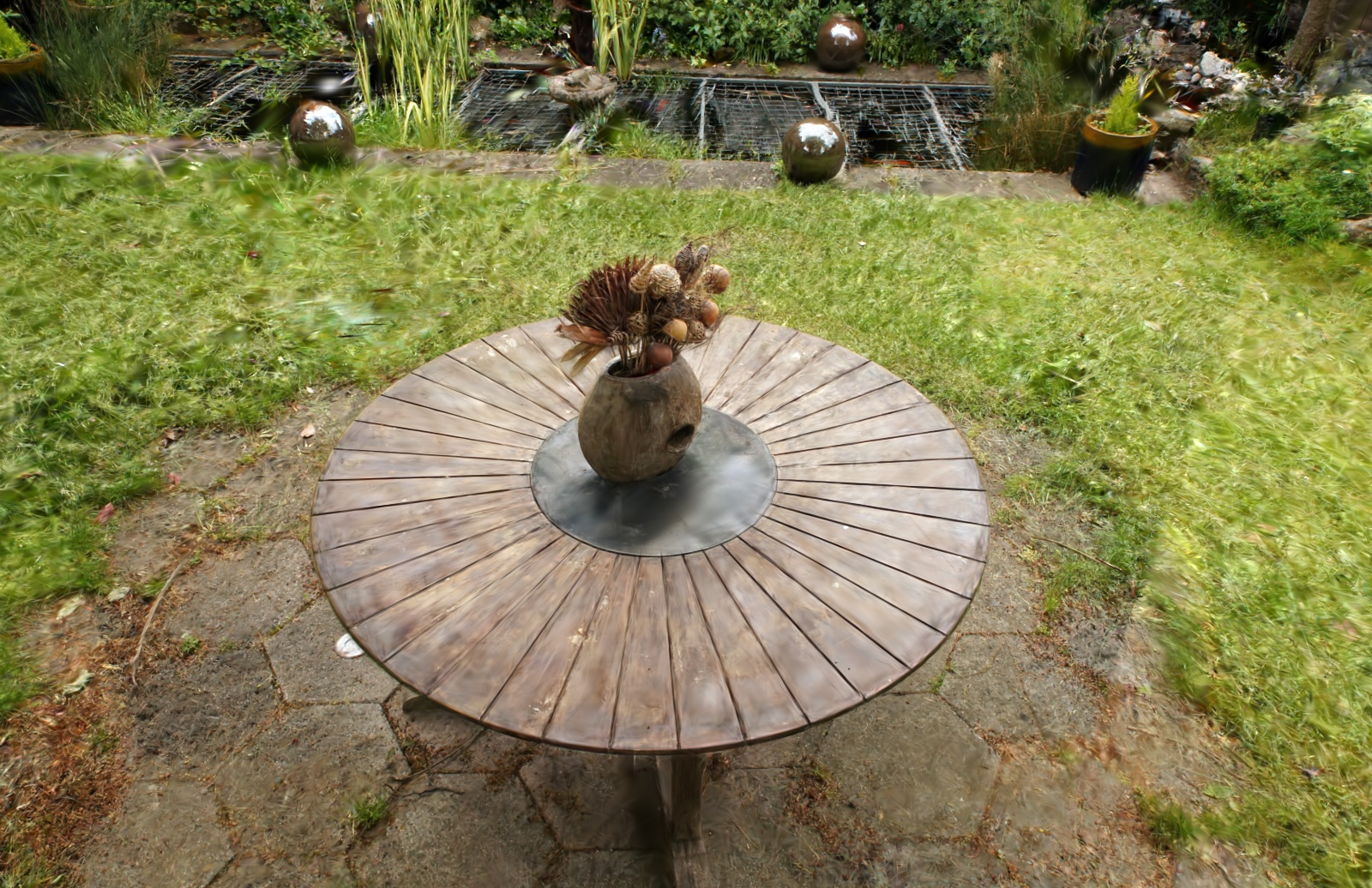} 
        & 
        \includegraphics[width=0.19\textwidth, height=0.127\textwidth]{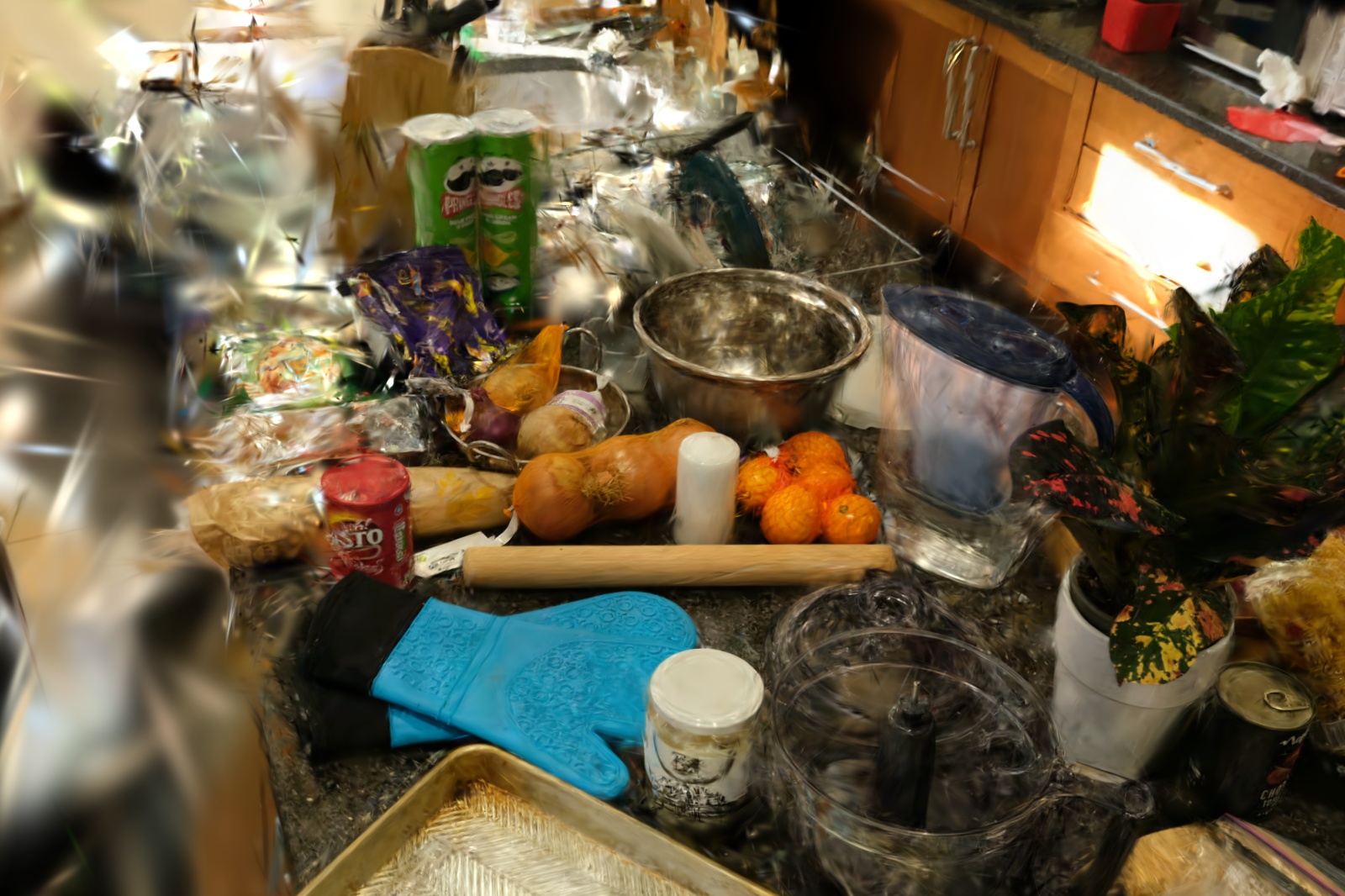}
        & 
        \includegraphics[width=0.19\textwidth, height=0.127\textwidth]{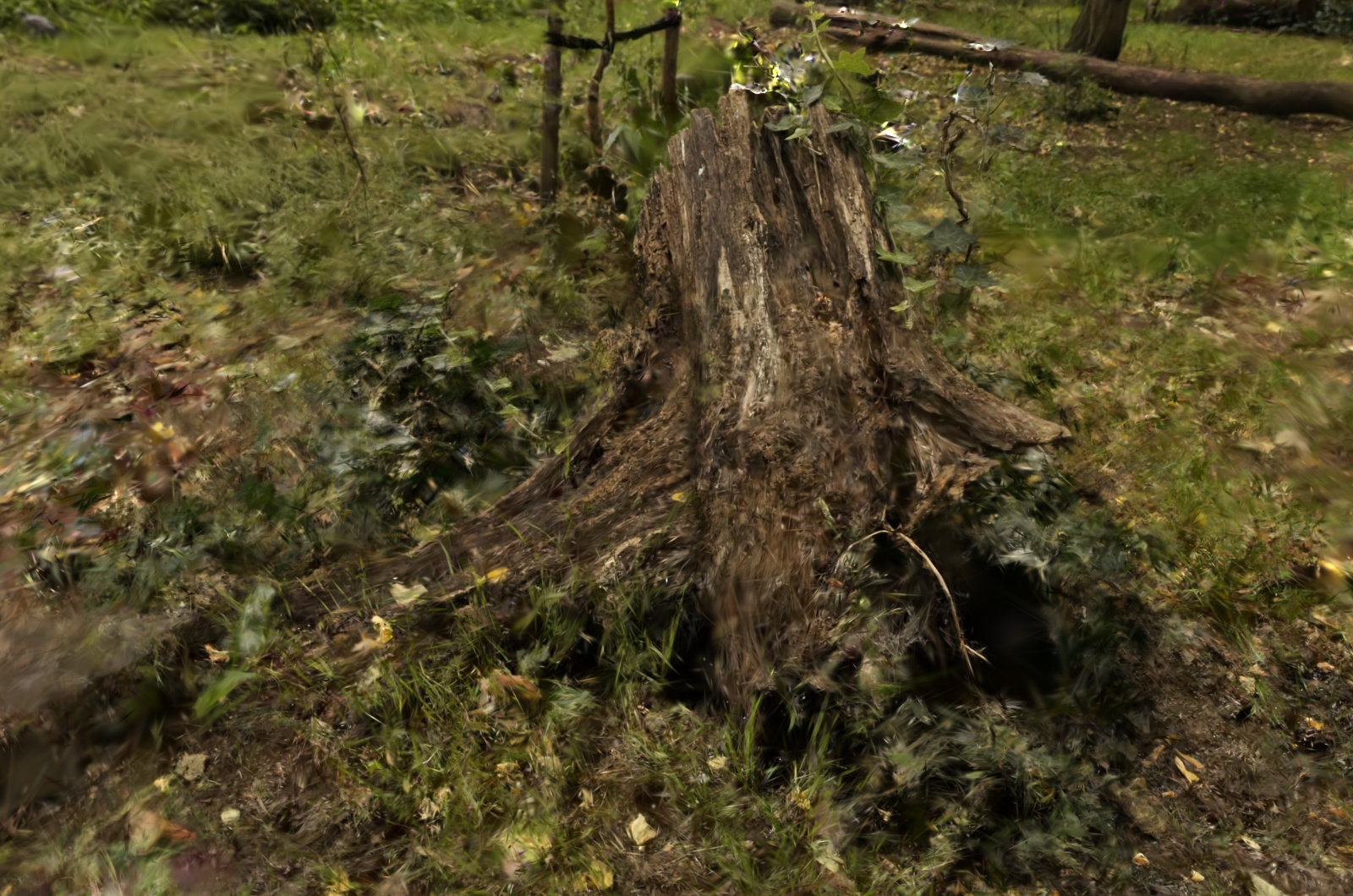} & 
        \includegraphics[width=0.19\textwidth, height=0.127\textwidth]{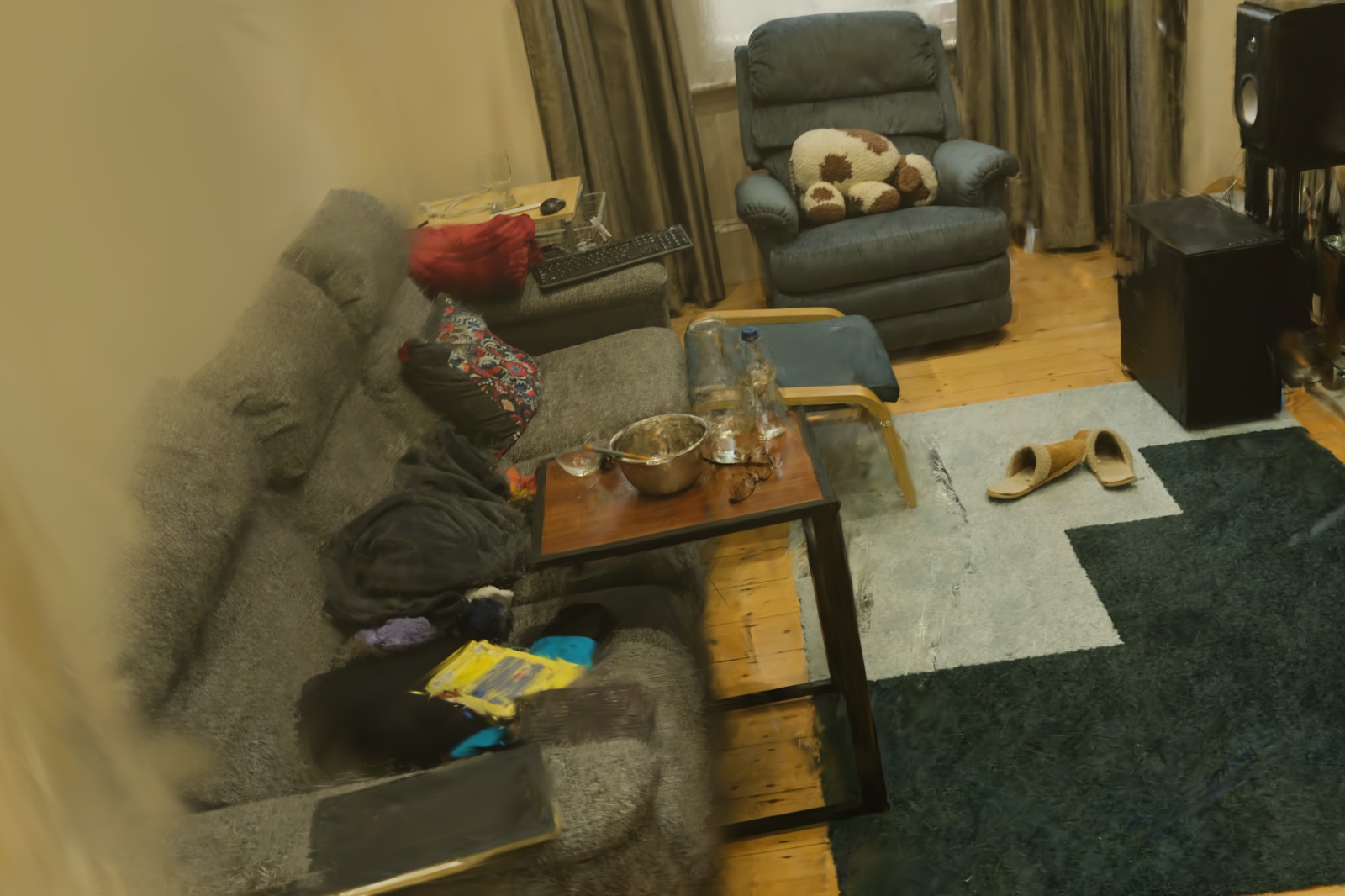}  \\
        \rotatebox[origin=l]{90}{\quad  WarpRF} &
        \includegraphics[width=0.19\textwidth, height=0.127\textwidth]{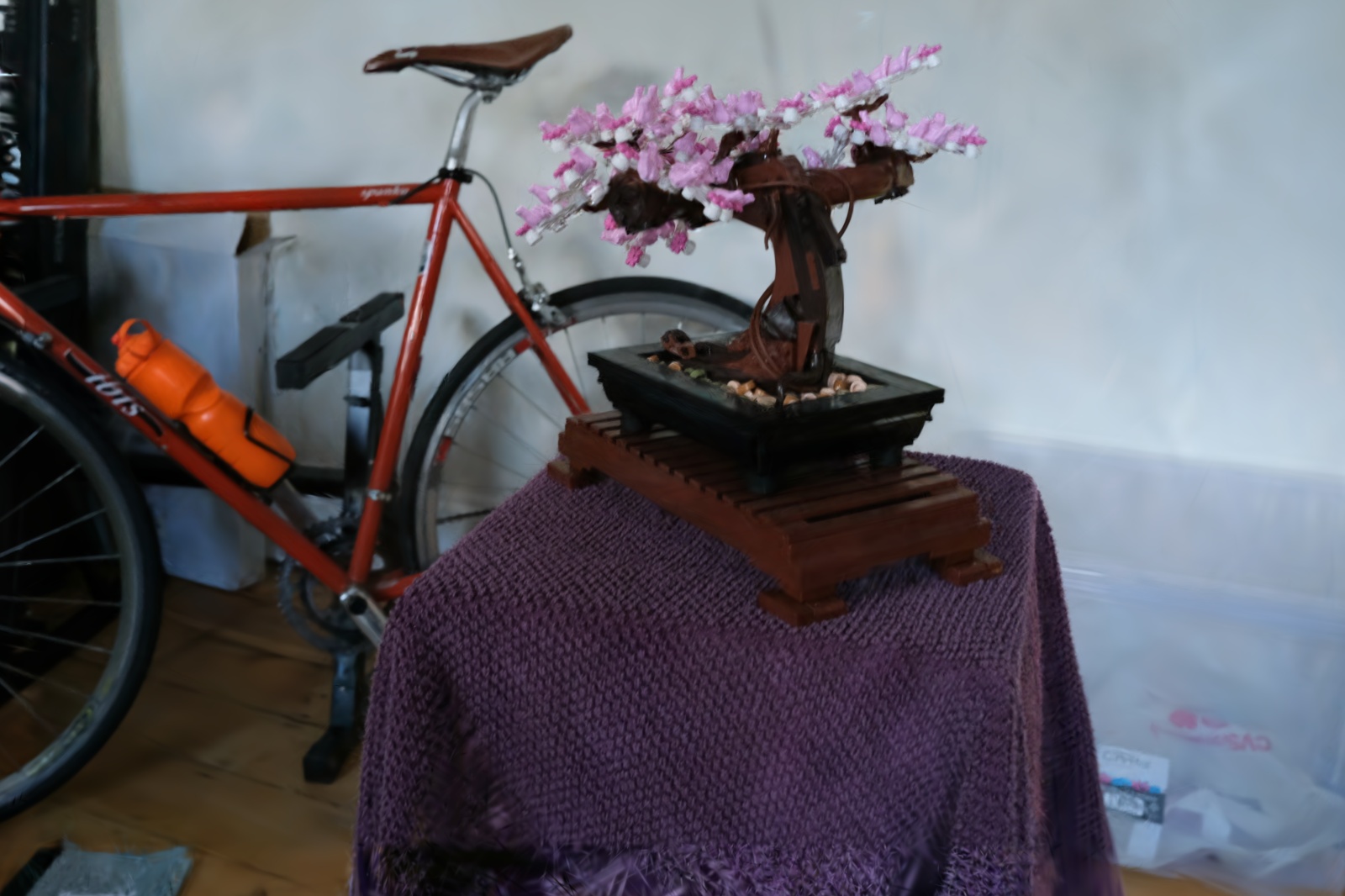} & 
        \includegraphics[width=0.19\textwidth, height=0.127\textwidth]{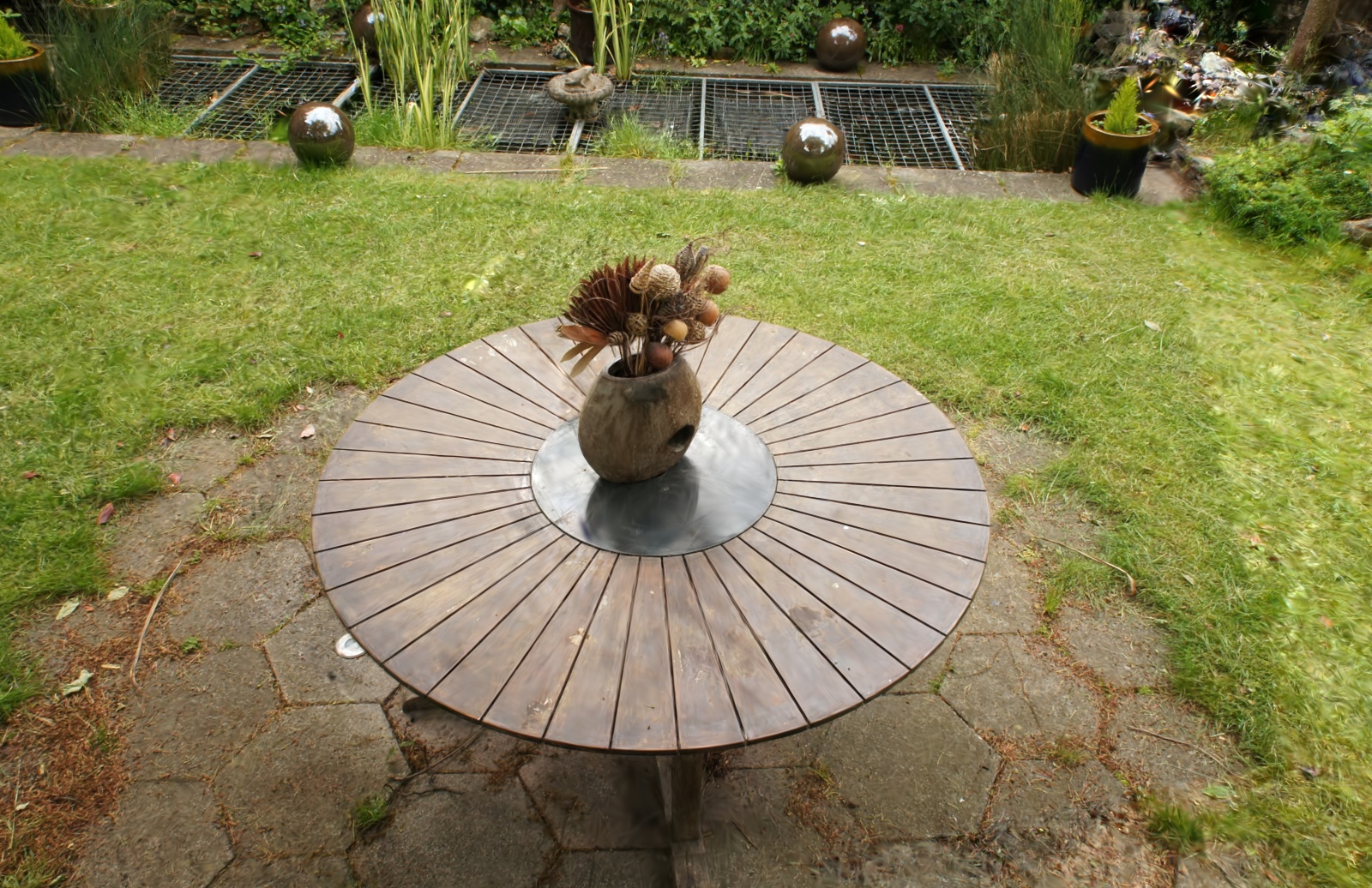} 
        & 
        \includegraphics[width=0.19\textwidth, height=0.127\textwidth]{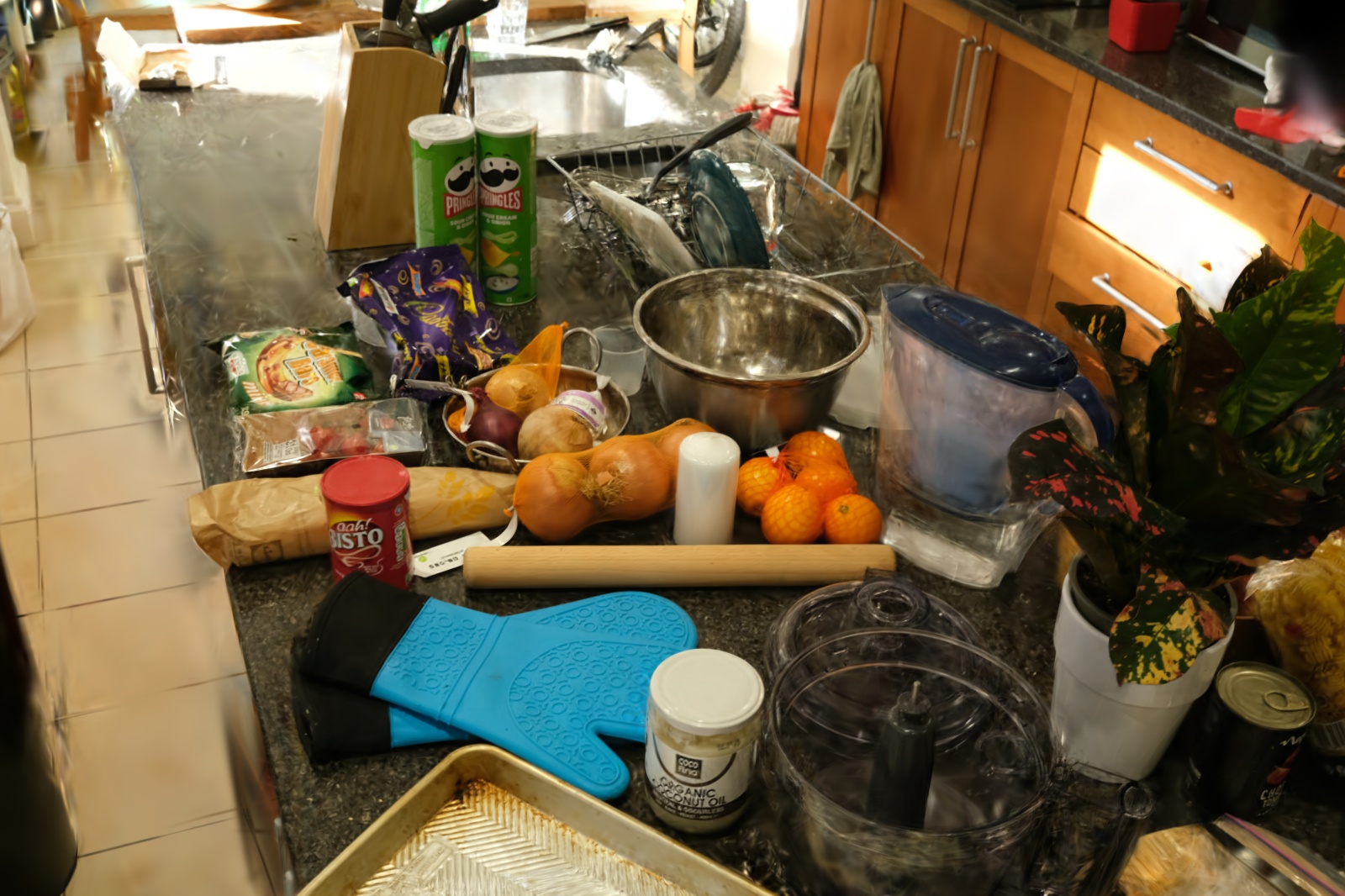}
        & 
        \includegraphics[width=0.19\textwidth, height=0.127\textwidth]{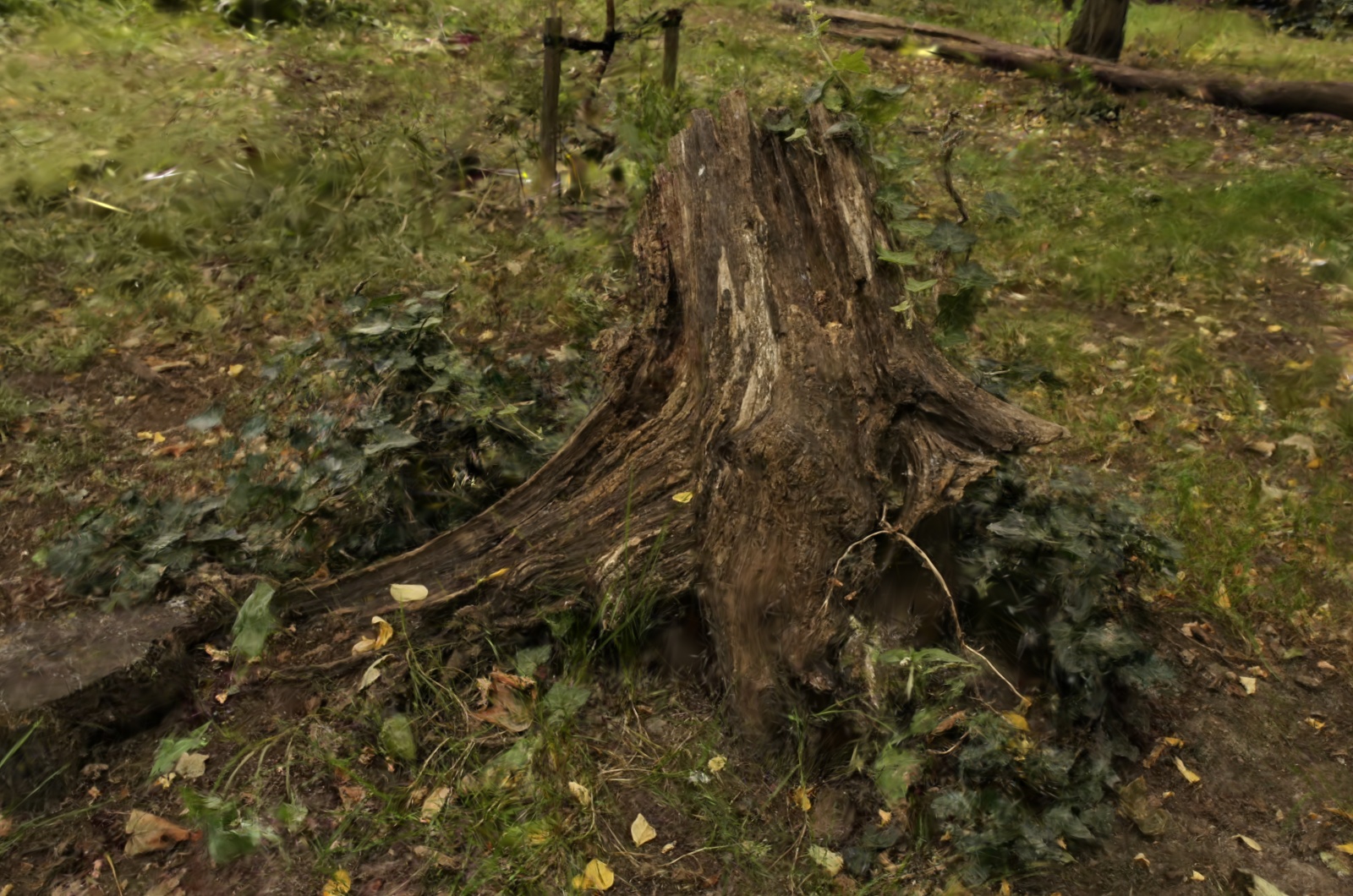} & 
        \includegraphics[width=0.19\textwidth, height=0.127\textwidth]{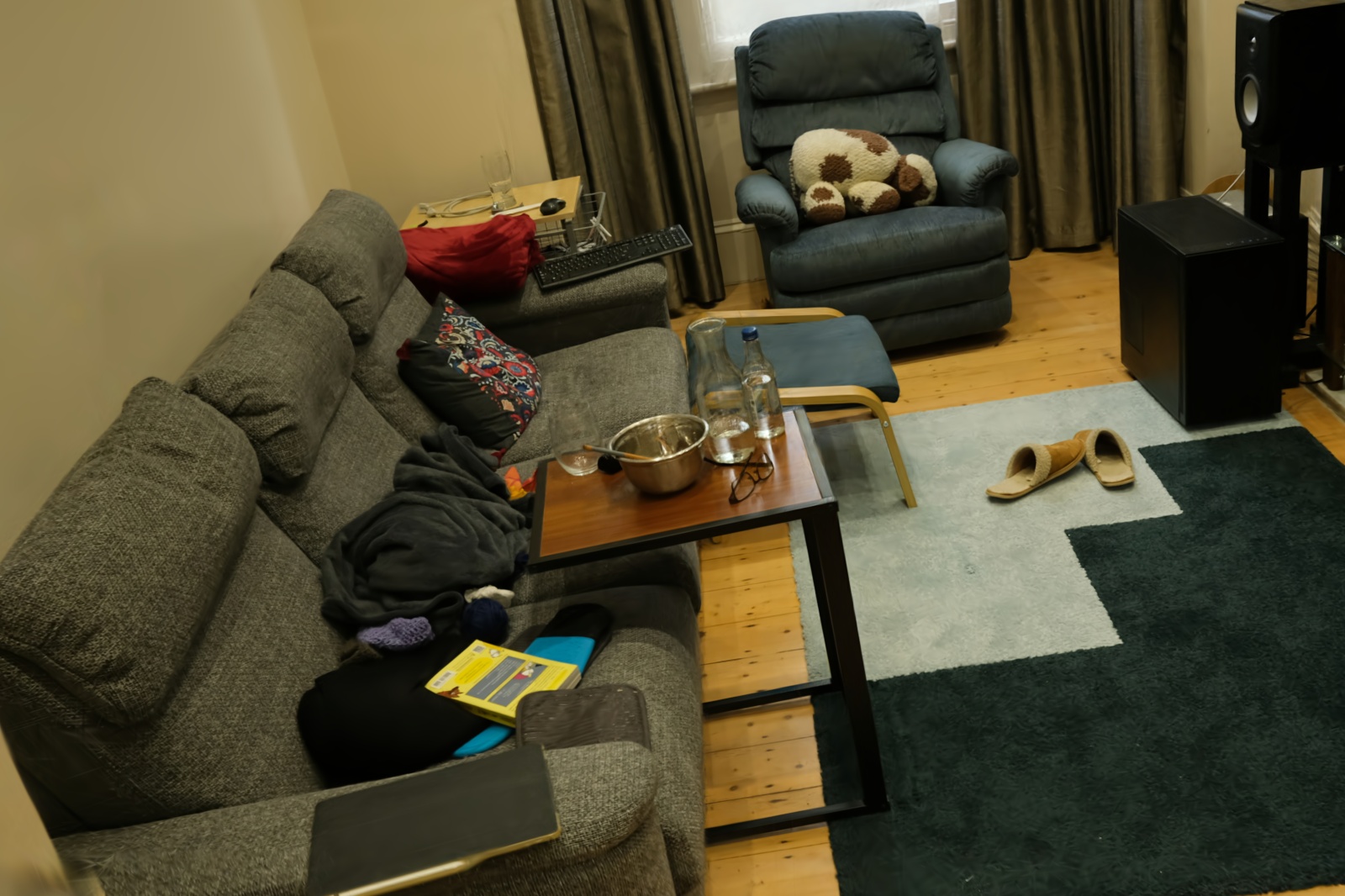}  \\
        \rotatebox[origin=l]{90}{\quad \quad  GT} &
        \includegraphics[width=0.19\textwidth, height=0.127\textwidth]{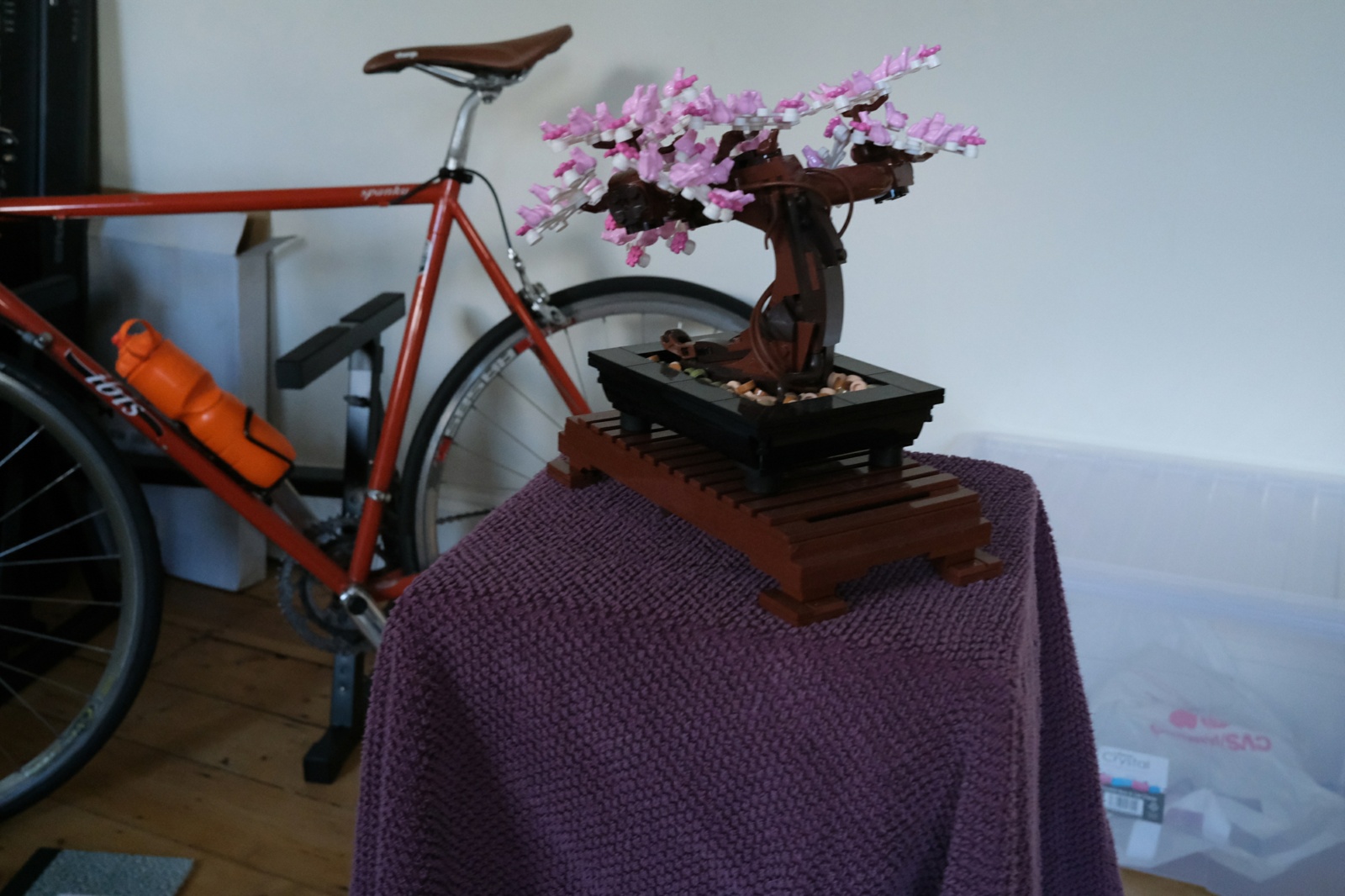} & 
        \includegraphics[width=0.19\textwidth, height=0.127\textwidth]{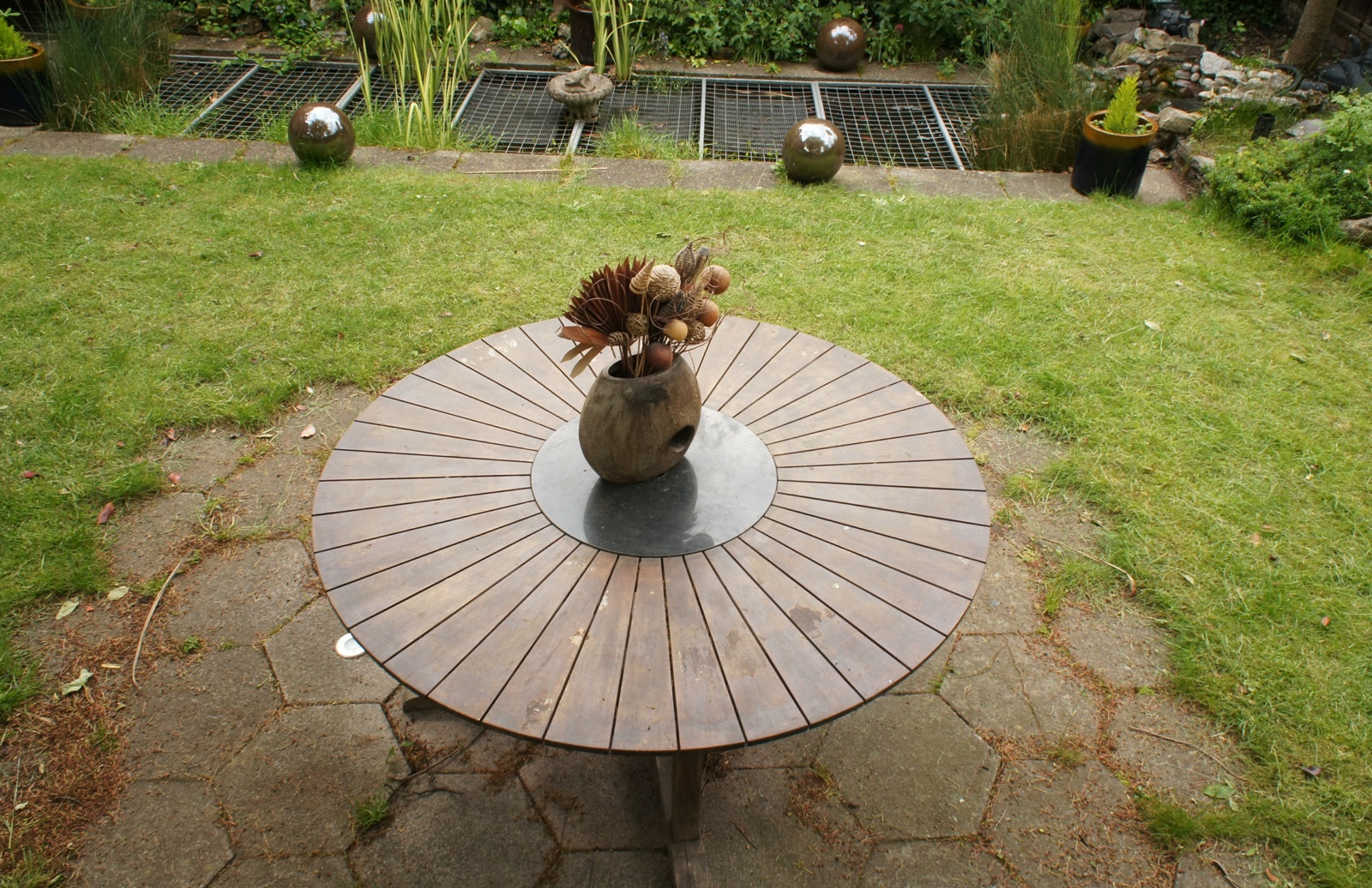} 
        & 
        \includegraphics[width=0.19\textwidth, height=0.127\textwidth]{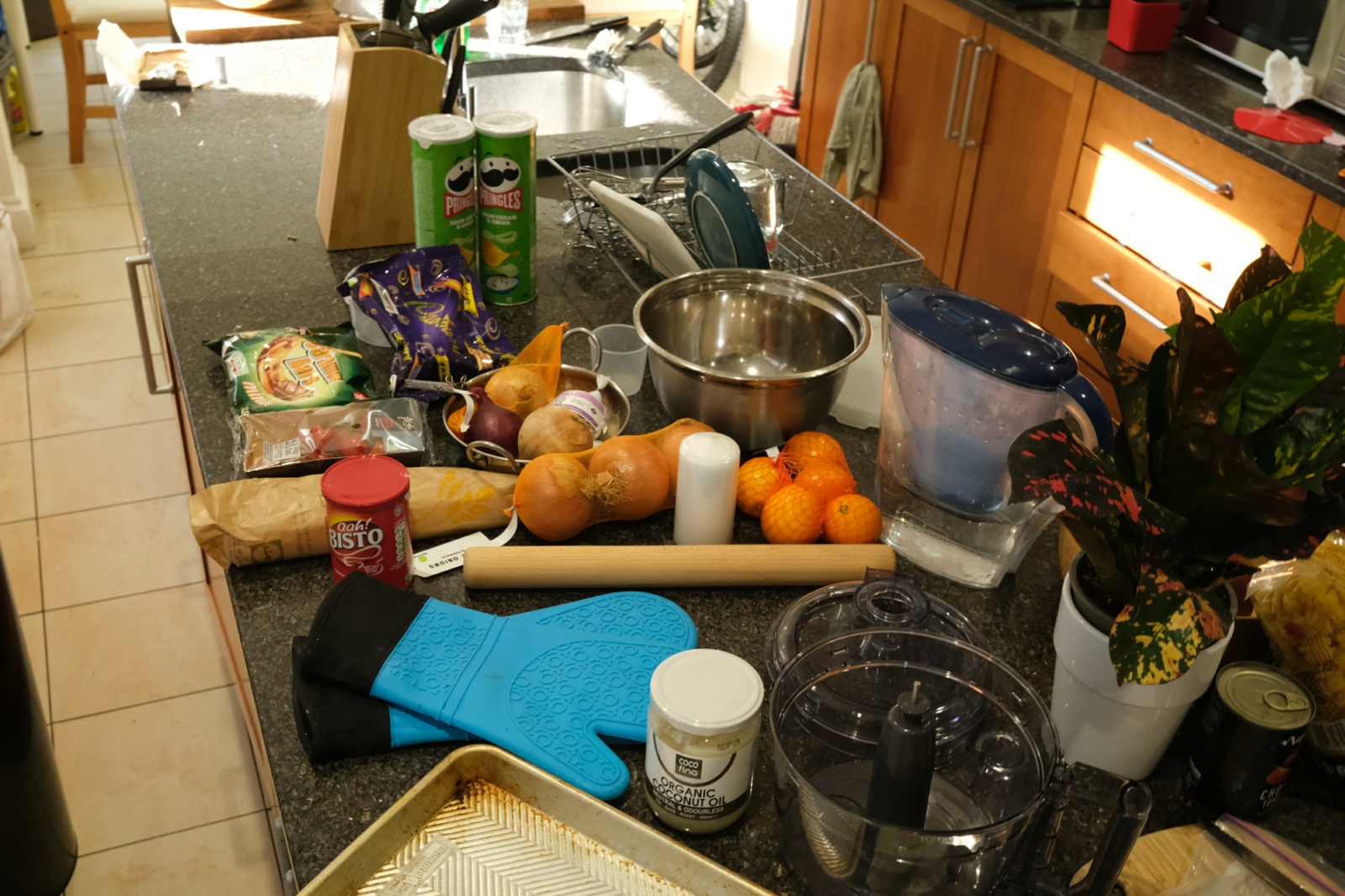}
        & 
        \includegraphics[width=0.19\textwidth, height=0.127\textwidth]{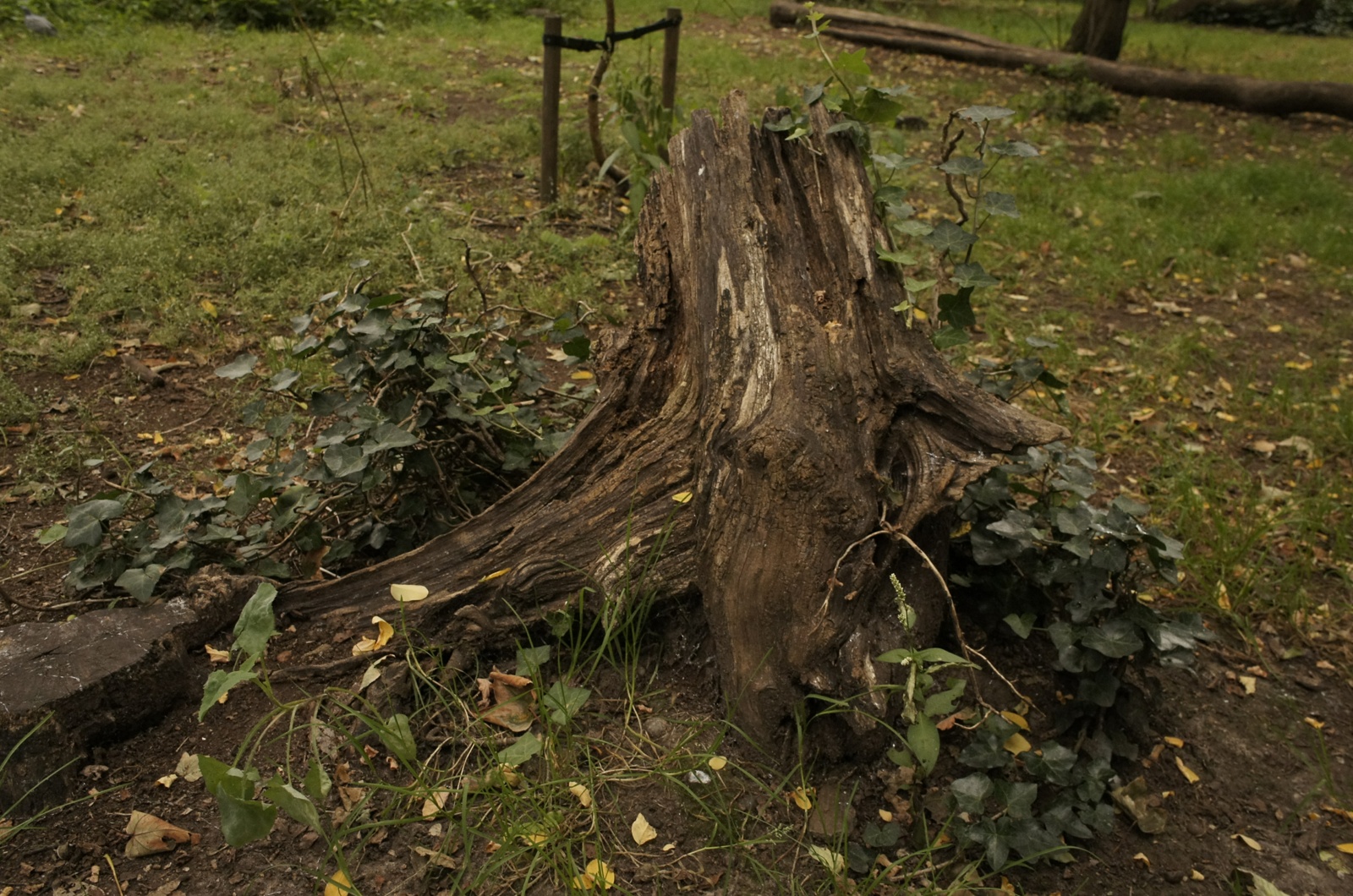} & 
        \includegraphics[width=0.19\textwidth, height=0.127\textwidth]{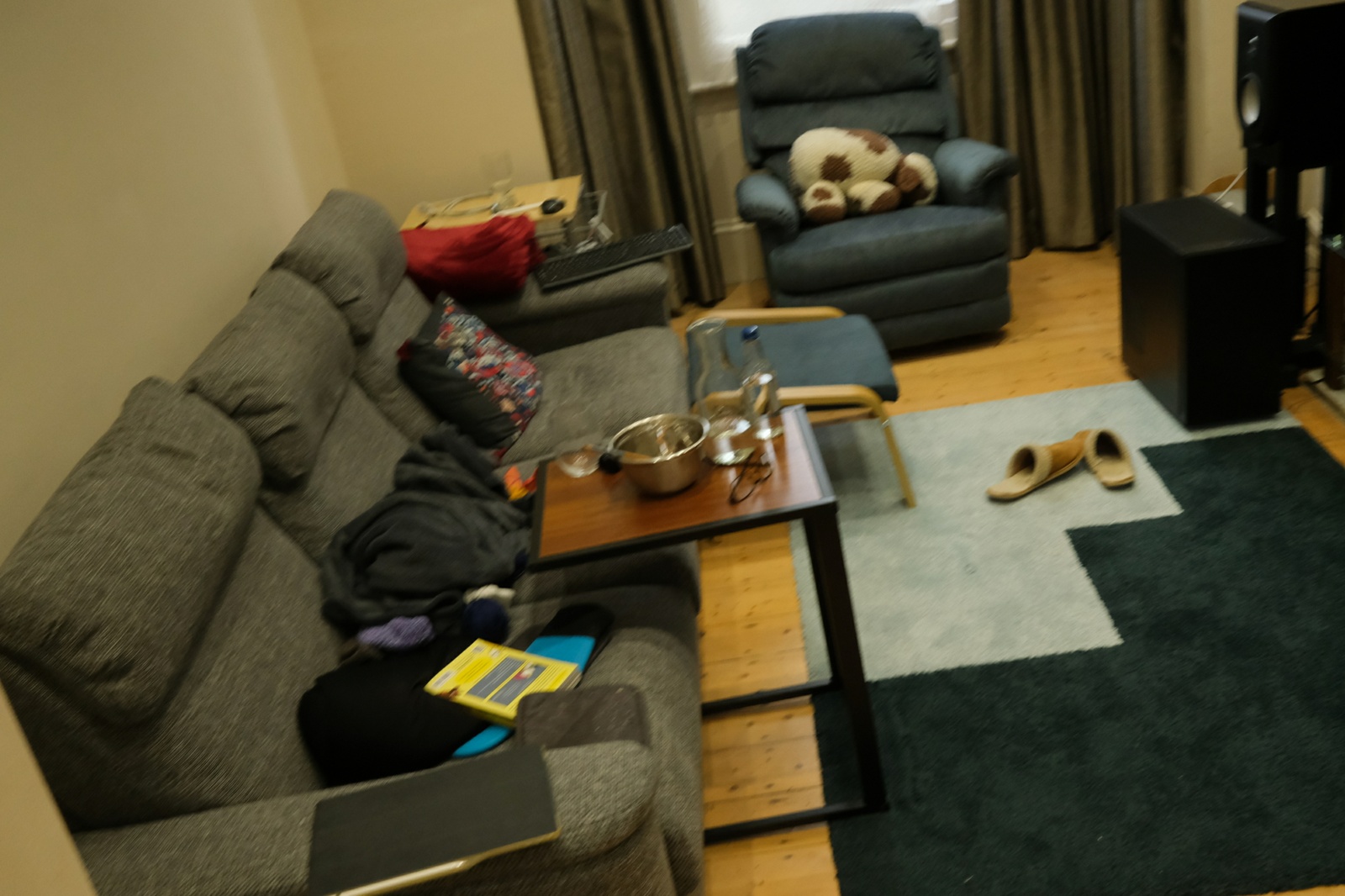}  \\
        
    \end{tabular}\vspace{-0.3cm}
    \caption{\textbf{Qualitative results of active camera selection on MipNeRF360 dataset.}  From top to bottom, we collect rendered results by 3DGS trained with to active view selection being performed by Manifold, FIsherRF and WarpRF, followed by the real images.}
    \label{fig:qual_mipnerf}
\end{figure*}

Quantitative results for active view selection are provided in Tables \ref{tab:active_mip}, \ref{tab:active_nerf_20} and \ref{tab:active_nerf_10}, respectively. We report the results as the average of three runs. We can see that WarpRF consistently outperforms the previous methods in both the real-world and synthetic scenes. Qualitative results are shown in Figures \ref{fig:qual_synth10} and \ref{fig:qual_mipnerf}.

\boldparagraph{SVRaster} We apply WarpRF to perform active view selection with SVRaster, marking the first attempt in this sense. In this experiment, we evaluate performance on the NeRF Synthetic and Tanks and Temples datasets. For NeRF Synthetic, we start with four initial views and iteratively select an additional view every 5\,000 iterations until reaching 20 views, after which we train the model for 20\,000 iterations. For the Tanks and Temples dataset, this process continues until 40 views are selected. Since no existing methods quantify uncertainty for SVRaster, we compare our approach against two baselines: one that selects cameras randomly and another that selects the camera farthest from the current training views at each step. The results are presented in Tables \ref{tab:active_nerf_20} (bottom) and \ref{tab:active_tandt} for NeRF Synthetic and Tanks and Temples, respectively. WarpRF allows for consisently more effective active view selection over both baselines.

\subsection{Active Mapping}
\label{sec:active_mapping}

To conclude, we apply WarpRF uncertainty modelling to perform active mapping for surface reconstruction, with both NeRF and 3DGS frameworks. 

\boldparagraph{NeRF} We follow a setup similar to that of \cite{xue2024nvf}. We begin with four initial views. At each stage, we train NeRF for 500 iterations, then sample candidate poses, calculate the uncertainties for the candidate views, select the one with the highest uncertainty and add it to the training views. This process continues until the NeRF model has been trained on 20 views.
Since our uncertainty definition is differentiable with respect to camera poses, we can also employ a gradient-based optimization strategy for finding the best views. Following a similar approach to \cite{xue2024nvf}, we first identify the top three poses with the highest uncertainty and then refine them through gradient-based optimization to maximize uncertainty.

We compare our method against NVF \cite{xue2024nvf} and WD \cite{lee2022uncertainty} methods on the `Lego' and `Hotdog' scenes, for which the authors provide code ready to be run. For evaluation, we assess the reconstructed meshes using Accuracy (Acc.), Completion (Comp.), and Completion Ratio (CR.), as proposed in \cite{sucar2021imap}, and evaluate the rendered images using standard image quality metrics.
The average results of multiple runs are reported in Tab. \ref{tab:mapping_nerf} where (Opt) refers to the gradient-based optimization strategy. As a selection strategy, WarpRF achieves the best results across all metrics in the `Lego' scene except for LPIPS, where we obtain very similar results to NVF. In the `Hotdog' scene, we achieve superior results in two out of three mesh reconstruction metrics and perform on par with NVF in RGB rendering metrics. With gradient-based optimization, our WarpRF outperforms NVF across all metrics in the `Lego' scene, and maintains superiority in two out of three mesh reconstruction metrics for the `Hotdog' scene while remaining competitive with NVF in RGB rendering metrics.

\begin{table*}[]
\centering
\scalebox{1.1}{
\begin{tabular}{cl|llllll}
\toprule
Scene                   & Method                                & Acc $\downarrow$    & Comp $\downarrow$  & CR $\uparrow$   & PSNR $\uparrow$  & SSIM $\uparrow$ & LPIPS $\downarrow$ \\ \midrule
\multirow{5}{*}{Lego}   & NeRF + WD \cite{lee2022uncertainty}                              & 0.0162 & 0.0229 & 0.376 & 23.402 & 0.878 & 0.099 \\
                        & NeRF + NVF \cite{xue2024nvf}                             & 0.0173 & 0.0216 & 0.375 & 23.543 & 0.878 & \fst 0.093 \\
                        & \bf NeRF + WarpRF (ours)                             & \fst 0.0161 & \fst 0.0212 & \fst 0.383 & \fst 23.898 & \fst 0.883 & 0.094 \\
                        \cline{2-8} 
                        & \multicolumn{1}{l|}{NeRF + NVF\textit{-opt} \cite{xue2024nvf}}  & 0.0161 & 0.0233 & 0.368 & 23.585 & 0.881 & 0.095 \\
                        & \multicolumn{1}{l|}{\bf NeRF + WarpRF\textit{-opt} (ours)} & \fst 0.0154 & \fst 0.0224 & \fst 0.374 & \fst 23.852 & \fst 0.881 & \fst 0.095 \\
                        
                        \midrule
\multirow{5}{*}{Hotdog} & NeRF + WD \cite{lee2022uncertainty}                              & 0.0259 & 0.0533 & 0.301 & 23.859 & 0.903 & 0.139 \\
                        & NeRF + NVF \cite{xue2024nvf}                              & 0.0253 & \fst 0.0496 & 0.315 & \fst 24.652 & \fst 0.913 & \fst 0.122 \\
                        & \bf NeRF + WarpRF (ours)                             & \fst 0.0228 & 0.0538 & \fst 0.315 &  24.465 &  0.912 &  0.126 \\ \cline{2-8} 
                        & \multicolumn{1}{l|}{NeRF + NVF\textit{-opt} \cite{xue2024nvf} } & 0.0232 & \fst 0.0499 & 0.305 & \fst 25.015 & \fst 0.914 & 0.123 \\
                        & \multicolumn{1}{l|}{\bf NeRF + WarpRF\textit{-opt} (ours) } & \fst 0.0217 & 0.0509 & \fst 0.321 & 24.635 & 0.913 & \fst 0.123 \\ \toprule
\end{tabular}}\vspace{-0.2cm}
    \caption{\textbf{Quantitative evaluation of active mapping with NeRF.} \textit{-opt} refers to gradient-based optimization strategy.}
\label{tab:mapping_nerf}
\end{table*}

\boldparagraph{3DGS} 
We employ PGSR \cite{chen2024pgsr} for the 3DGS-based evaluation. Our experiments are conducted on the Tanks and Temples \cite{knapitsch2017tanks} dataset. We begin with four uniformly selected views, and incrementally add a new frame every $K$ iterations for the first 20\,000 iterations, until a total of 40 views are selected. Training then continues for a total of 30,000 iterations. We compare WarpRF with FisherRF \cite{Jiang2024FisherRF} and report the results in Tab.  \ref{tab:mesh_tnt}. Once again, our method consistently outperforms FisherRF in every scene. Qualitative results are shown in Fig. \ref{fig:qual_tnt}.

\begin{table}[]
\centering
\renewcommand{\tabcolsep}{10pt}
\begin{tabular}{l|cc}
\toprule
Scene       & FisherRF \cite{Jiang2024FisherRF} &    \bf WarpRF (ours) \\ \midrule
Barn        & 0.266         & \fst 0.322     \\
Caterpillar & 0.241         & \fst 0.263     \\
Courthouse  & 0.065         & \fst 0.095     \\
Ignatitus   & 0.622         & \fst 0.692     \\
Meetingroom & 0.128         & \fst 0.134     \\
Truck       & 0.416         & \fst 0.453     \\ \midrule
Mean        & 0.290         & \fst 0.326    \\ \bottomrule
\end{tabular}\vspace{-0.2cm}
\caption{\textbf{Quantitative evaluation of active mapping with 3DGS (F1 score $\uparrow$).} Comparison between 3DGS variants trained thorugh active mapping being performed by FisherRF and WarpRF.}
\label{tab:mesh_tnt}
\end{table}

        

\begin{figure}[t]
    \centering
    \renewcommand{\tabcolsep}{1pt}
    \begin{tabular}{cc}        
    FisherRF \cite{Jiang2024FisherRF} & WarpRF \\ 
        \includegraphics[width=0.25\textwidth]{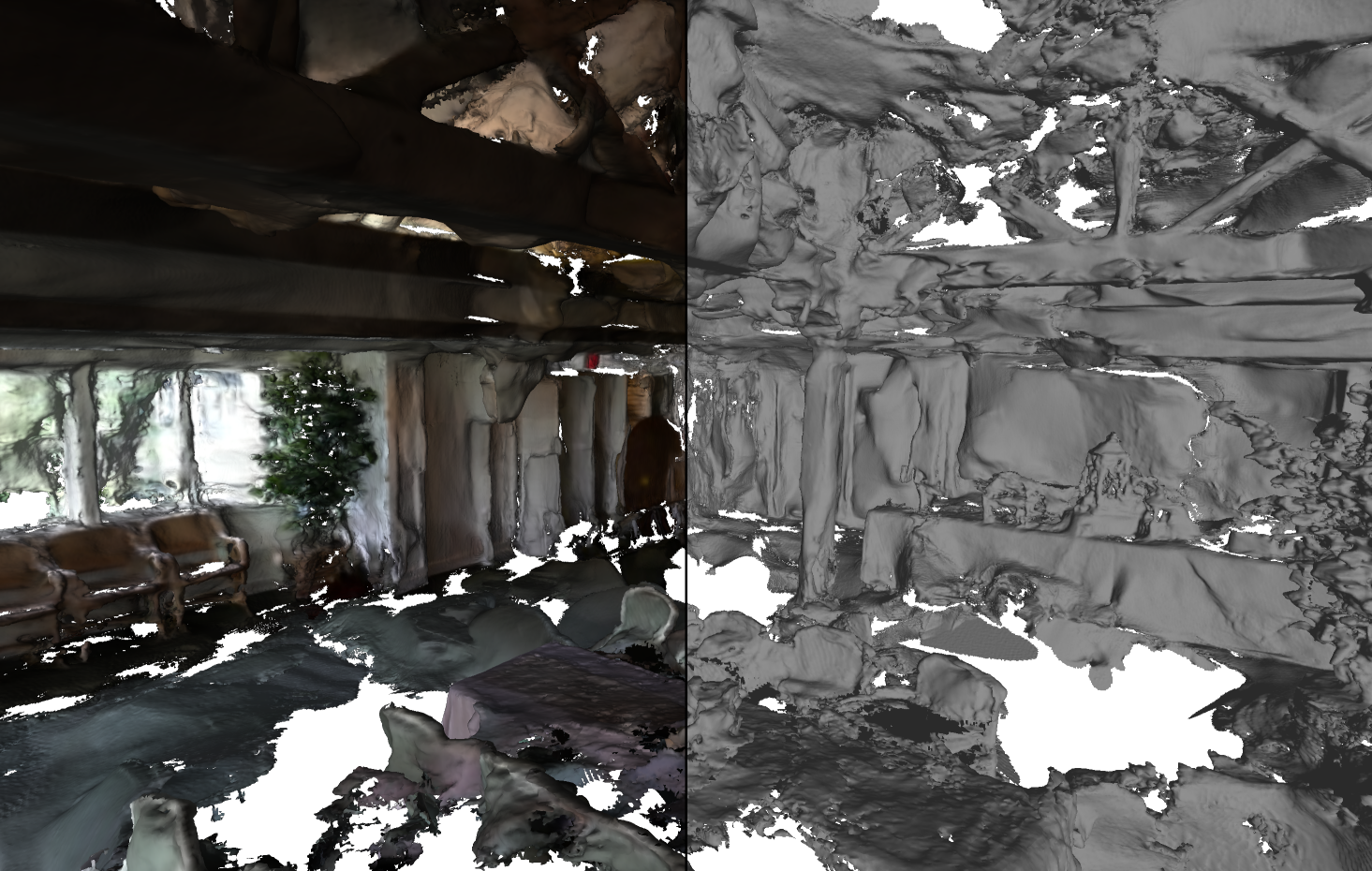} & 
        \includegraphics[width=0.25\textwidth]{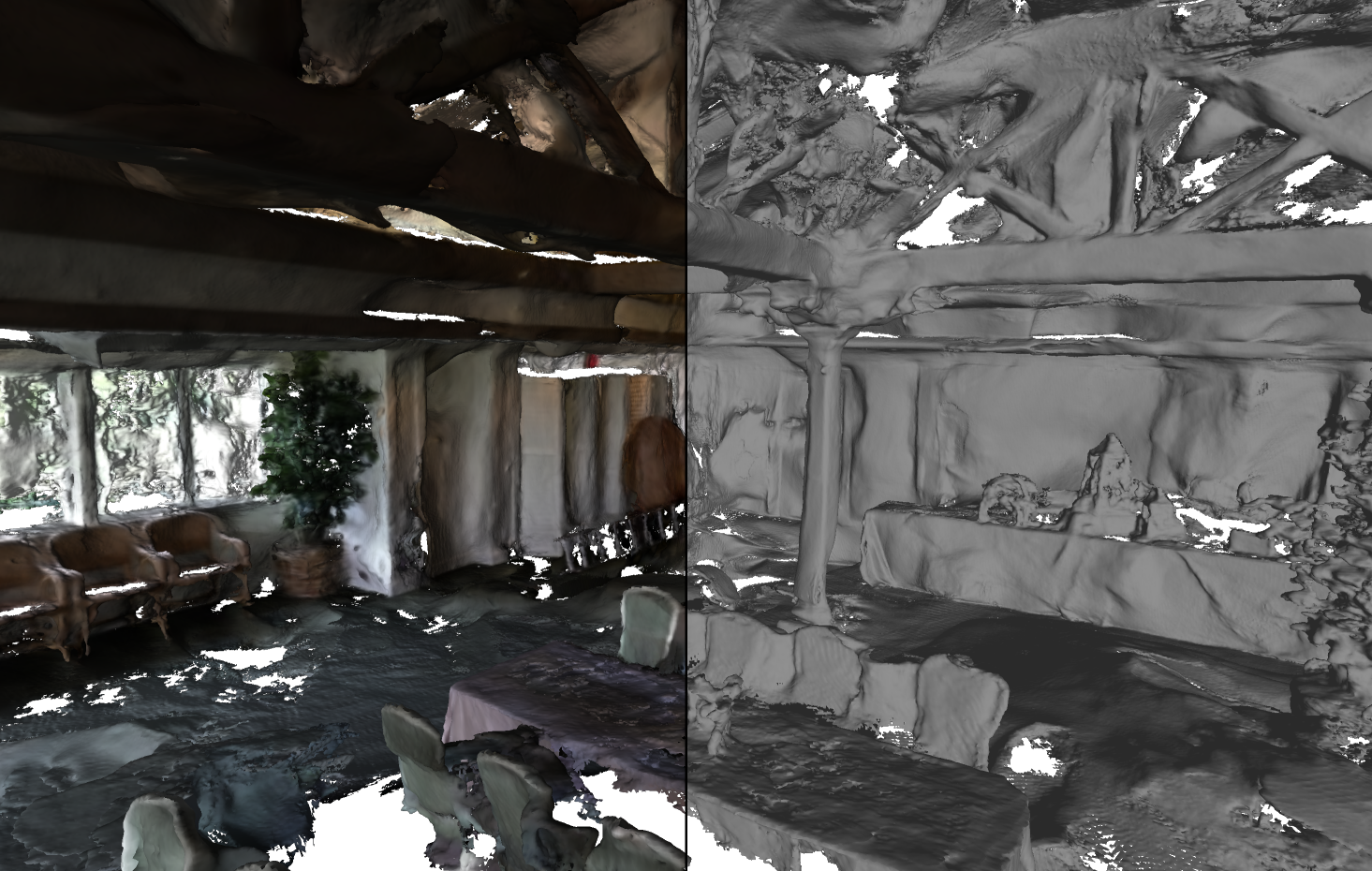} 
        \\
        \includegraphics[width=0.25\textwidth]{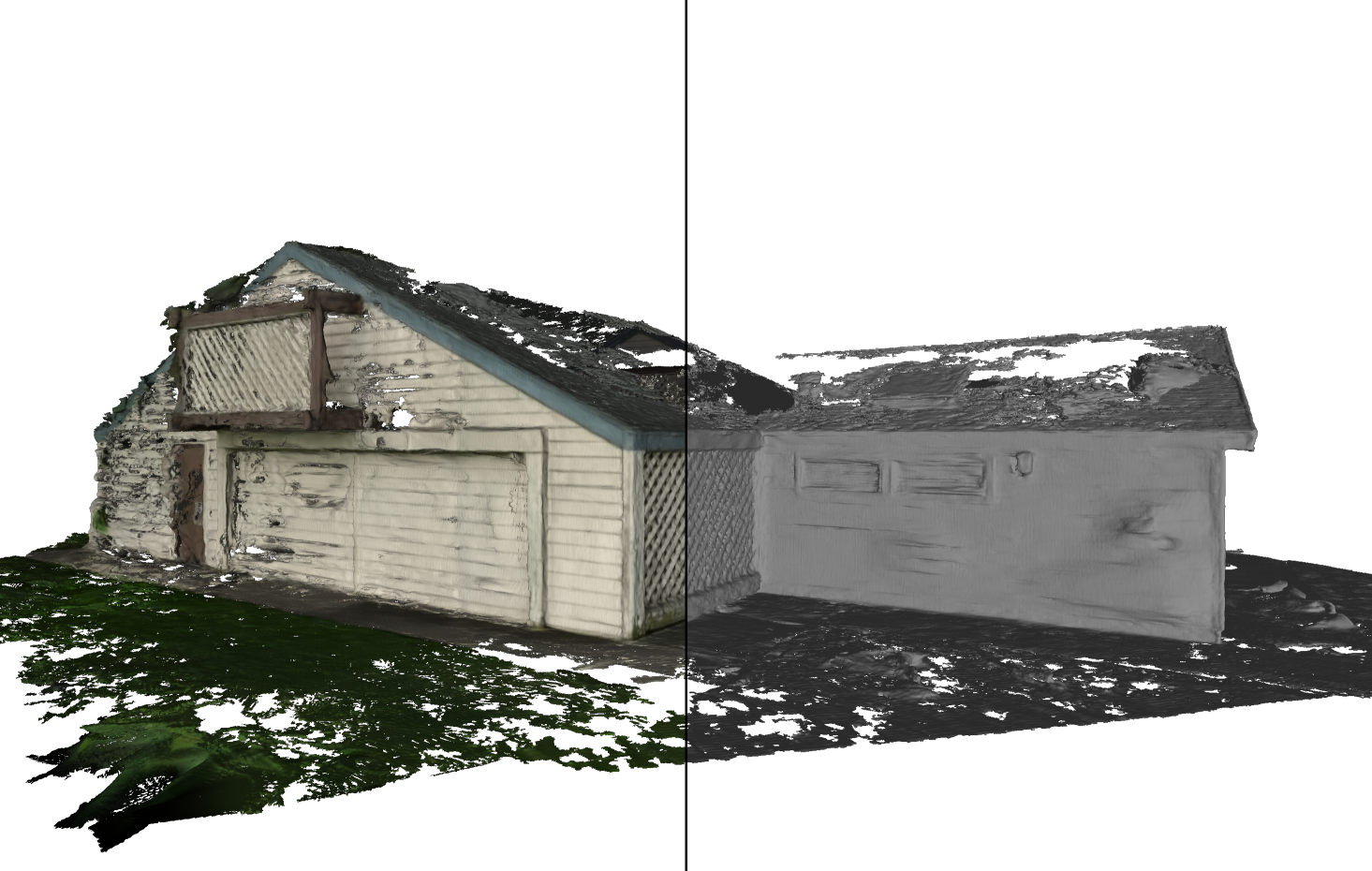} & 
        \includegraphics[width=0.25\textwidth]{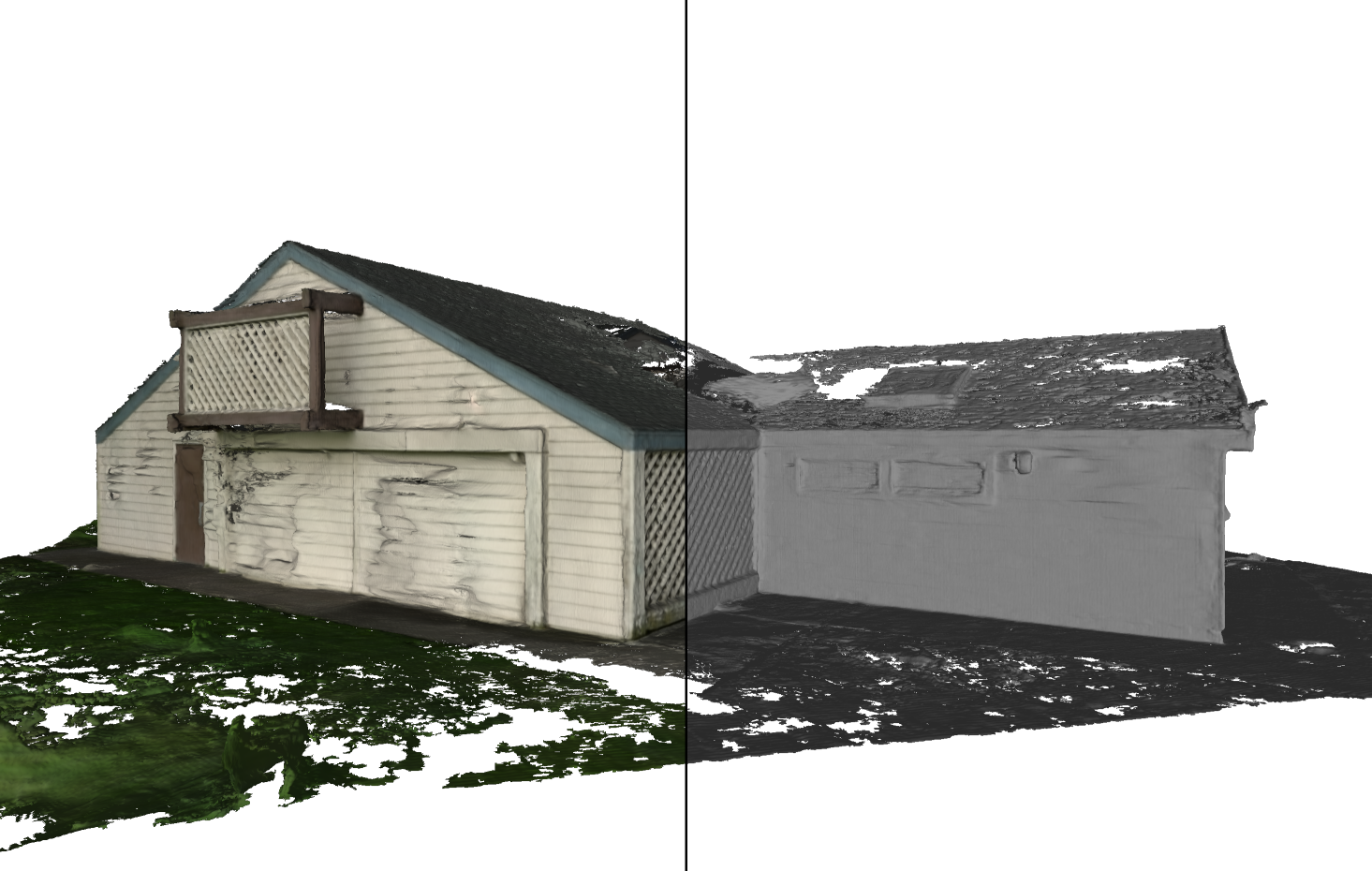}
        \\
        \includegraphics[width=0.25\textwidth]{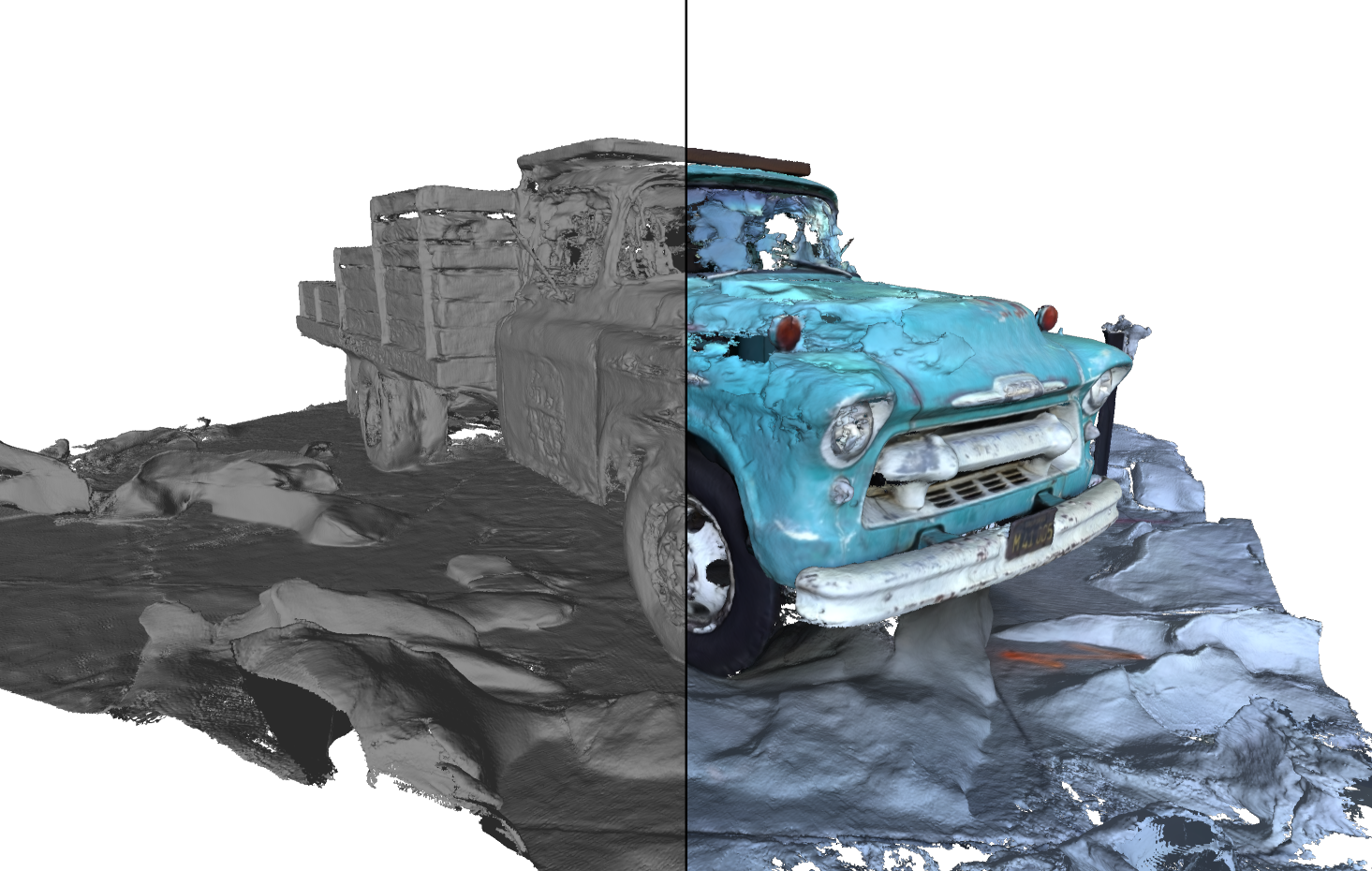}
        &
        \includegraphics[width=0.25\textwidth]{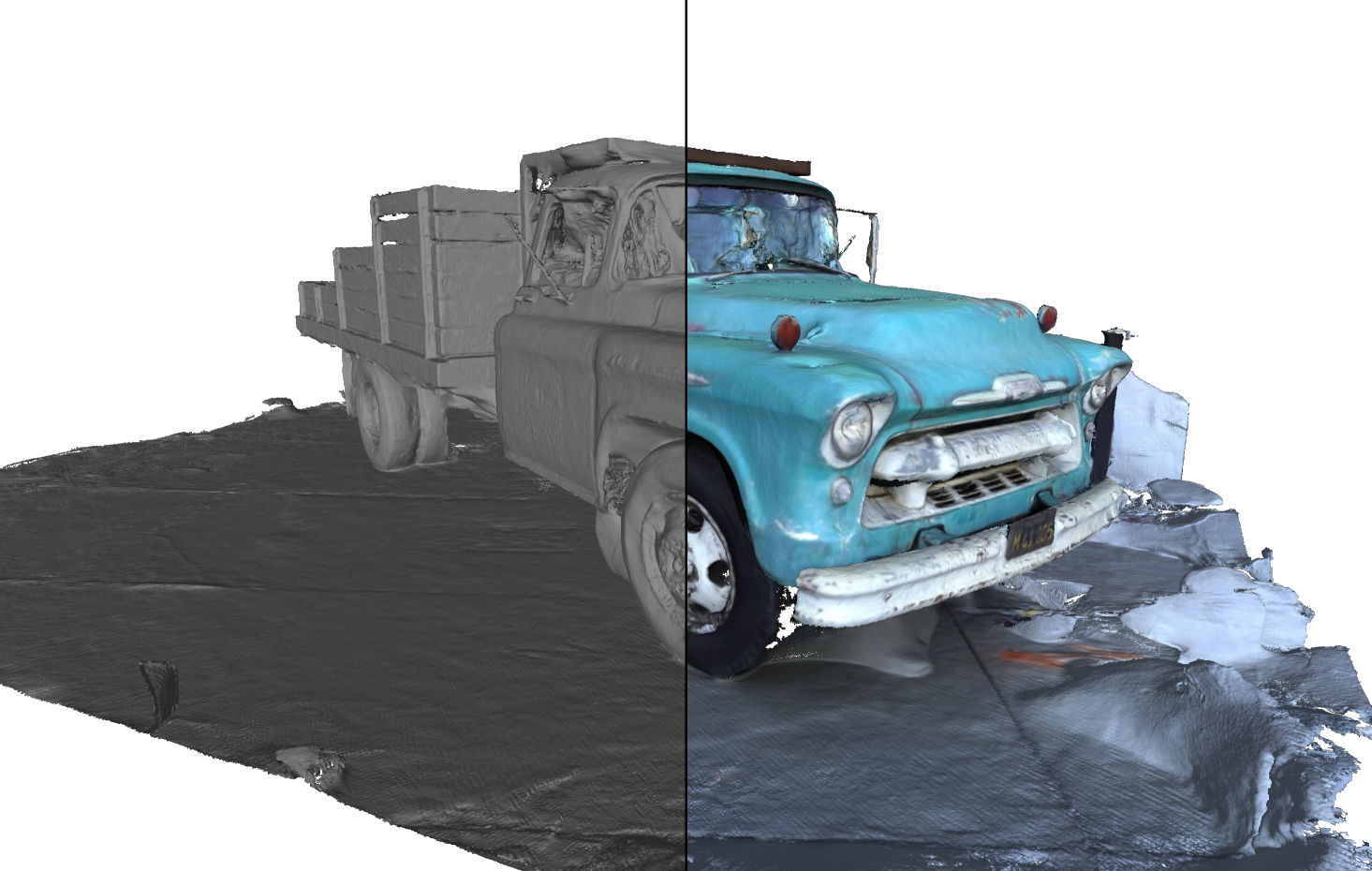}
        \\
        
        \includegraphics[width=0.25\textwidth]{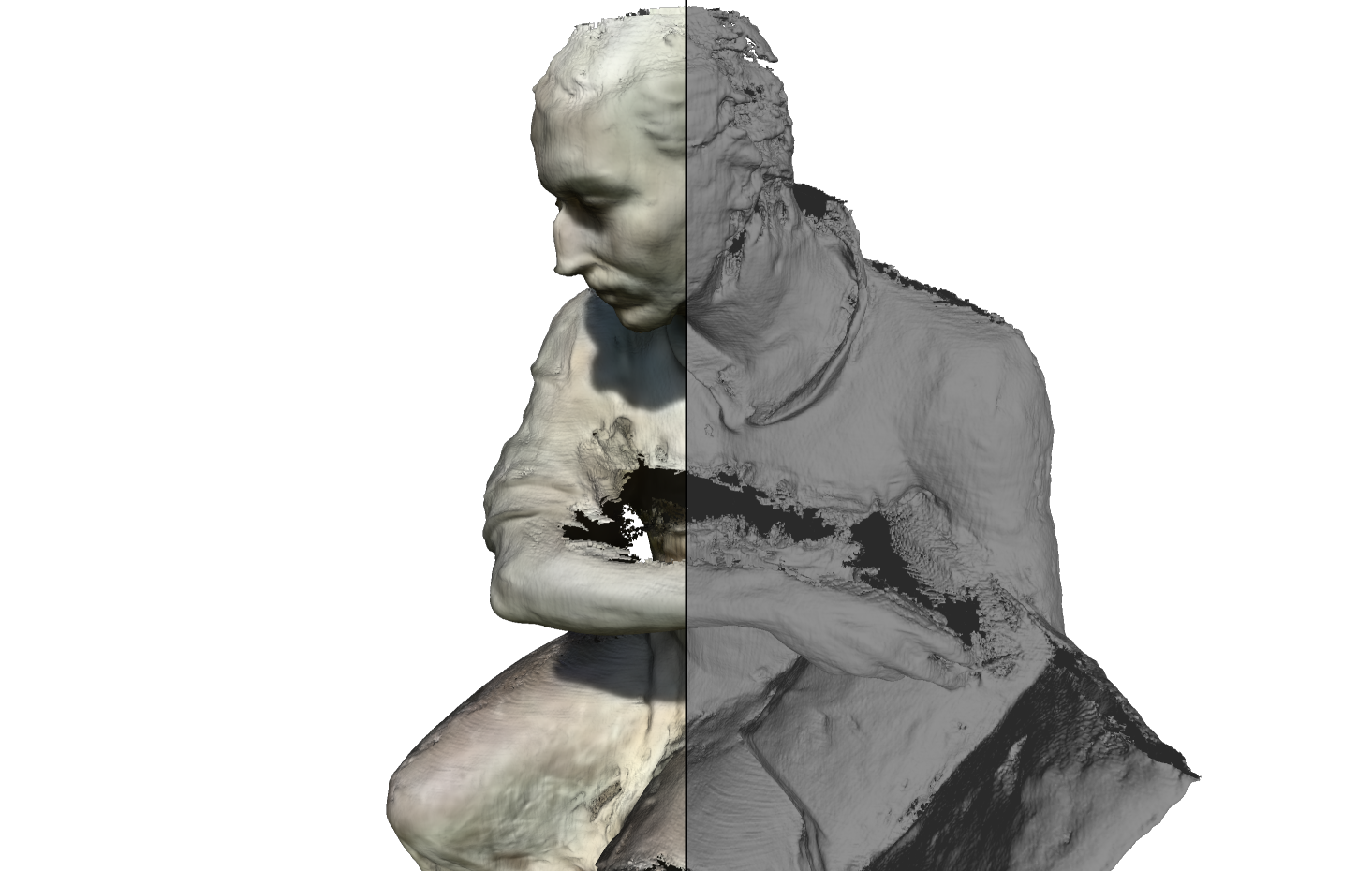} 
        & 
        \includegraphics[width=0.25\textwidth]{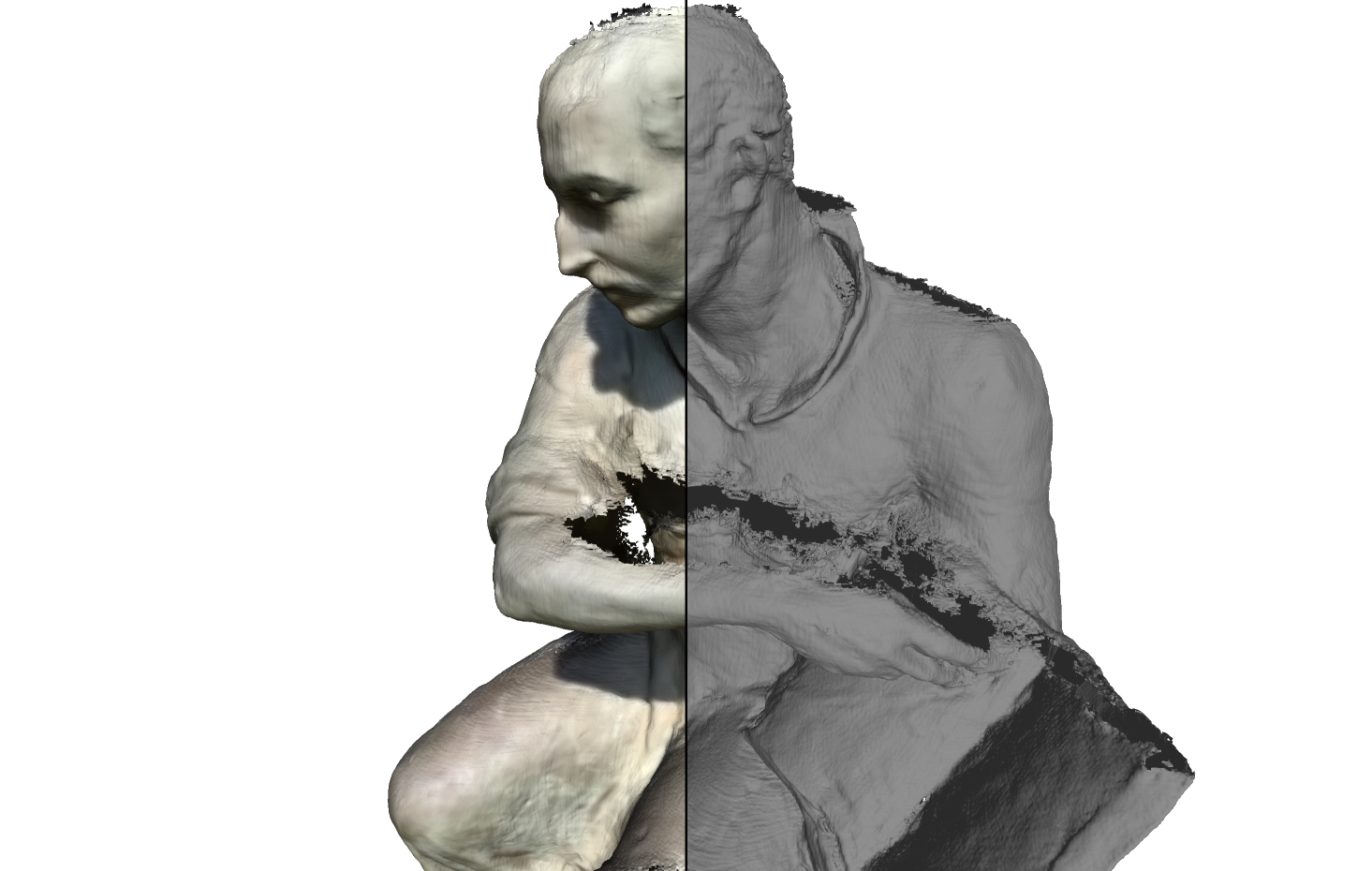} 
     \\
    \end{tabular}
    \caption{\textbf{Reconstructed meshes using active mapping on Tanks and Temples dataset.} We show meshes reconstructed by 3DGS trained through active mapping being performed by FisherRF and WarpRF.}
    \label{fig:qual_tnt}
\end{figure}

\section{Conclusion}
\label{sec:conc}

In this paper, we have presented WarpRF, a training-free approach to quantify uncertainty in any radiance field implementation. Built upon principles of multi-view geometry, our method relies on warping to retrieve reliable renderings from known viewpoints and measure their consistency with those rendered out of a novel, unseen viewpoint. 
WarpRF is simple, elegant and does not require any training; it can be applied on top of pre-trained radiance fields independently of their internal structure, whether based on NeRF, 3DGS or more advanced models.
Through our exhaustive evaluation, we have demonstrated that WarpRF achieves state-of-the-art performance on uncertainty quantification and downstream tasks, such as active view selection and active mapping, at the expense of more intrusive solutions tailored for the specific radiance field implementations. 

\boldparagraph{Acknowledgements}
Sadra Safadoust was supported by KUIS AI Fellowship and UNVEST R\&D Center. This project is funded by the European Union (ERC, ENSURE, 101116486) with additional compute support from
Leonardo Booster (EuroHPC Joint Undertaking, EHPC-AI2024A01-060). Views and opinions expressed are however those of the author(s) only and do not necessarily reflect those of the European Union or the European Research
Council. Neither the European Union nor the granting authority can be held responsible for them.

{
    \small
    \bibliographystyle{ieeenat_fullname}
    \bibliography{main}

\begin{thebibliography}{41}
\providecommand{\natexlab}[1]{#1}
\providecommand{\url}[1]{\texttt{#1}}
\expandafter\ifx\csname urlstyle\endcsname\relax
  \providecommand{\doi}[1]{doi: #1}\else
  \providecommand{\doi}{doi: \begingroup \urlstyle{rm}\Url}\fi

\bibitem[Barron et~al.(2022)Barron, Mildenhall, Verbin, Srinivasan, and Hedman]{barron2022mip}
Jonathan~T Barron, Ben Mildenhall, Dor Verbin, Pratul~P Srinivasan, and Peter Hedman.
\newblock Mip-nerf 360: Unbounded anti-aliased neural radiance fields.
\newblock In \emph{CVPR}, pages 5470--5479, 2022.

\bibitem[Chen et~al.(2024)Chen, Li, Ye, Wang, Xie, Zhai, Wang, Liu, Bao, and Zhang]{chen2024pgsr}
Danpeng Chen, Hai Li, Weicai Ye, Yifan Wang, Weijian Xie, Shangjin Zhai, Nan Wang, Haomin Liu, Hujun Bao, and Guofeng Zhang.
\newblock Pgsr: Planar-based gaussian splatting for efficient and high-fidelity surface reconstruction.
\newblock \emph{arXiv preprint arXiv:2406.06521}, 2024.

\bibitem[Deng et~al.(2022)Deng, Liu, Zhu, and Ramanan]{kangle2021dsnerf}
Kangle Deng, Andrew Liu, Jun-Yan Zhu, and Deva Ramanan.
\newblock Depth-supervised {NeRF}: Fewer views and faster training for free.
\newblock In \emph{Proceedings of the IEEE/CVF Conference on Computer Vision and Pattern Recognition (CVPR)}, 2022.

\bibitem[Furukawa and Ponce(2009)]{furukawa2009accurate}
Yasutaka Furukawa and Jean Ponce.
\newblock Accurate, dense, and robust multiview stereopsis.
\newblock \emph{IEEE TPAMI}, 32\penalty0 (8):\penalty0 1362--1376, 2009.

\bibitem[Gawlikowski et~al.(2023)Gawlikowski, Tassi, Ali, Lee, Humt, Feng, Kruspe, Triebel, Jung, Roscher, et~al.]{gawlikowski2023survey}
Jakob Gawlikowski, Cedrique Rovile~Njieutcheu Tassi, Mohsin Ali, Jongseok Lee, Matthias Humt, Jianxiang Feng, Anna Kruspe, Rudolph Triebel, Peter Jung, Ribana Roscher, et~al.
\newblock A survey of uncertainty in deep neural networks.
\newblock \emph{Artificial Intelligence Review}, 56\penalty0 (Suppl 1):\penalty0 1513--1589, 2023.

\bibitem[Godard et~al.(2019)Godard, {Mac Aodha}, Firman, and Brostow]{monodepth2}
Cl{\'{e}}ment Godard, Oisin {Mac Aodha}, Michael Firman, and Gabriel~J. Brostow.
\newblock Digging into self-supervised monocular depth prediction.
\newblock In \emph{ICCV}, 2019.

\bibitem[Goli et~al.(2024)Goli, Reading, Sellán, Jacobson, and Tagliasacchi]{goli2023}
Lily Goli, Cody Reading, Silvia Sellán, Alec Jacobson, and Andrea Tagliasacchi.
\newblock {Bayes' Rays}: Uncertainty quantification in neural radiance fields.
\newblock In \emph{CVPR}, 2024.

\bibitem[Gu{\'e}don and Lepetit(2024)]{sugar}
Antoine Gu{\'e}don and Vincent Lepetit.
\newblock Sugar: Surface-aligned gaussian splatting for efficient 3d mesh reconstruction and high-quality mesh rendering.
\newblock In \emph{CVPR}, 2024.

\bibitem[Gu{\'e}don et~al.(2022)Gu{\'e}don, Monasse, and Lepetit]{guedon2022scone}
Antoine Gu{\'e}don, Pascal Monasse, and Vincent Lepetit.
\newblock Scone: Surface coverage optimization in unknown environments by volumetric integration.
\newblock In \emph{NeurIPS}, pages 20731--20743, 2022.

\bibitem[Gu{\'e}don et~al.(2023)Gu{\'e}don, Monnier, Monasse, and Lepetit]{guedon2023macarons}
Antoine Gu{\'e}don, Tom Monnier, Pascal Monasse, and Vincent Lepetit.
\newblock Macarons: Mapping and coverage anticipation with rgb online self-supervision.
\newblock In \emph{CVPR}, pages 940--951, 2023.

\bibitem[Hu et~al.(2024)Hu, Chen, Feng, Li, Yang, Bao, Zhang, and Cui]{hu2024cg}
Jiarui Hu, Xianhao Chen, Boyin Feng, Guanglin Li, Liangjing Yang, Hujun Bao, Guofeng Zhang, and Zhaopeng Cui.
\newblock Cg-slam: Efficient dense rgb-d slam in a consistent uncertainty-aware 3d gaussian field.
\newblock In \emph{ECCV}, pages 93--112. Springer, 2024.

\bibitem[Jiang et~al.(2024)Jiang, Lei, and Daniilidis]{Jiang2024FisherRF}
Wen Jiang, Boshu Lei, and Kostas Daniilidis.
\newblock Fisherrf: Active view selection and mapping with radiance fields using fisher information.
\newblock In \emph{ECCV}, pages 422--440, 2024.

\bibitem[Kendall and Gal(2017)]{kendall2017uncertainties}
Alex Kendall and Yarin Gal.
\newblock What uncertainties do we need in bayesian deep learning for computer vision?
\newblock In \emph{NeurIPS}, 2017.

\bibitem[Kerbl et~al.(2023)Kerbl, Kopanas, Leimk{\"u}hler, and Drettakis]{kerbl3Dgaussians}
Bernhard Kerbl, Georgios Kopanas, Thomas Leimk{\"u}hler, and George Drettakis.
\newblock 3d gaussian splatting for real-time radiance field rendering.
\newblock \emph{ACM Transactions on Graphics}, 42\penalty0 (4), 2023.

\bibitem[Kim et~al.(2025)Kim, Lim, and Han]{kim20254d}
Mijeong Kim, Jongwoo Lim, and Bohyung Han.
\newblock 4d gaussian splatting in the wild with uncertainty-aware regularization.
\newblock \emph{Advances in Neural Information Processing Systems}, 37:\penalty0 129209--129226, 2025.

\bibitem[Klasson et~al.(2024)Klasson, Mereu, Kannala, and Solin]{klasson2024sources}
Marcus Klasson, Riccardo Mereu, Juho Kannala, and Arno Solin.
\newblock Sources of uncertainty in 3d scene reconstruction.
\newblock In \emph{ECCV}. Springer, 2024.

\bibitem[Knapitsch et~al.(2017)Knapitsch, Park, Zhou, and Koltun]{knapitsch2017tanks}
Arno Knapitsch, Jaesik Park, Qian-Yi Zhou, and Vladlen Koltun.
\newblock Tanks and temples: Benchmarking large-scale scene reconstruction.
\newblock \emph{ACM Transactions on Graphics (ToG)}, 36\penalty0 (4):\penalty0 1--13, 2017.

\bibitem[Lee et~al.(2022)Lee, Chen, Wang, Liniger, Kumar, and Yu]{lee2022uncertainty}
Soomin Lee, Le Chen, Jiahao Wang, Alexander Liniger, Suryansh Kumar, and Fisher Yu.
\newblock Uncertainty guided policy for active robotic 3d reconstruction using neural radiance fields.
\newblock \emph{IEEE Robotics and Automation Letters}, 7\penalty0 (4):\penalty0 12070--12077, 2022.

\bibitem[Lyu et~al.(2024)Lyu, Tewari, Habermann, Saito, Zollh{\"o}fer, Leimk{\"u}ehler, and Theobalt]{lyu2024manifold}
Linjie Lyu, Ayush Tewari, Marc Habermann, Shunsuke Saito, Michael Zollh{\"o}fer, Thomas Leimk{\"u}ehler, and Christian Theobalt.
\newblock Manifold sampling for differentiable uncertainty in radiance fields.
\newblock In \emph{SIGGRAPH Asia Conference Proceedings}, 2024.

\bibitem[Mildenhall et~al.(2020)Mildenhall, Srinivasan, Tancik, Barron, Ramamoorthi, and Ng]{mildenhall2020nerf}
Ben Mildenhall, Pratul~P. Srinivasan, Matthew Tancik, Jonathan~T. Barron, Ravi Ramamoorthi, and Ren Ng.
\newblock Nerf: Representing scenes as neural radiance fields for view synthesis.
\newblock In \emph{ECCV}, 2020.

\bibitem[Pan et~al.(2022)Pan, Lai, Song, and Huang]{pan2022activenerf}
Xuran Pan, Zihang Lai, Shiji Song, and Gao Huang.
\newblock Activenerf: Learning where to see with uncertainty estimation.
\newblock In \emph{ECCV}, pages 230--246. Springer, 2022.

\bibitem[Poggi et~al.(2022)Poggi, Kim, Tosi, Aleotti, Min, Sohn, and Mattoccia]{poggi2022confidence}
M Poggi, S Kim, F Tosi, F Aleotti, D Min, K Sohn, and S Mattoccia.
\newblock On the confidence of stereo matching in a deep-learning era: a quantitative evaluation.
\newblock \emph{IEEE TPAMI}, 44\penalty0 (9):\penalty0 5293--5313, 2022.

\bibitem[Ran et~al.(2023)Ran, Zeng, He, Chen, Li, Chen, Lee, and Ye]{Ran2023neurar}
Yunlong Ran, Jing Zeng, Shibo He, Jiming Chen, Lincheng Li, Yingfeng Chen, Gimhee Lee, and Qi Ye.
\newblock Neurar: Neural uncertainty for autonomous 3d reconstruction with implicit neural representations.
\newblock \emph{IEEE Robotics and Automation Letters}, 8\penalty0 (2):\penalty0 1125–1132, 2023.

\bibitem[Ren et~al.(2024)Ren, Zhu, Sun, Chen, Pollefeys, and Peng]{ren2024nerf}
Weining Ren, Zihan Zhu, Boyang Sun, Jiaqi Chen, Marc Pollefeys, and Songyou Peng.
\newblock Nerf on-the-go: Exploiting uncertainty for distractor-free nerfs in the wild.
\newblock In \emph{CVPR}, pages 8931--8940, 2024.

\bibitem[Roessle et~al.(2022)Roessle, Barron, Mildenhall, Srinivasan, and Nie{\ss}ner]{roessle2022depthpriorsnerf}
Barbara Roessle, Jonathan~T. Barron, Ben Mildenhall, Pratul~P. Srinivasan, and Matthias Nie{\ss}ner.
\newblock Dense depth priors for neural radiance fields from sparse input views.
\newblock In \emph{Proceedings of the IEEE/CVF Conference on Computer Vision and Pattern Recognition (CVPR)}, 2022.

\bibitem[Safadoust et~al.(2024)Safadoust, Tosi, G{\"u}ney, and Poggi]{safadoust2024BMVC}
Sadra Safadoust, Fabio Tosi, Fatma G{\"u}ney, and Matteo Poggi.
\newblock Self-evolving depth-supervised 3d gaussian splatting from rendered stereo pairs.
\newblock In \emph{BMVC}, 2024.

\bibitem[Schops et~al.(2017)Schops, Schonberger, Galliani, Sattler, Schindler, Pollefeys, and Geiger]{schops2017eth3d}
Thomas Schops, Johannes~L Schonberger, Silvano Galliani, Torsten Sattler, Konrad Schindler, Marc Pollefeys, and Andreas Geiger.
\newblock A multi-view stereo benchmark with high-resolution images and multi-camera videos.
\newblock In \emph{CVPR}, pages 3260--3269, 2017.

\bibitem[Shen et~al.(2021)Shen, Ruiz, Agudo, and Moreno{-}Noguer]{shen2021snerf}
Jianxiong Shen, Adria Ruiz, Antonio Agudo, and Francesc Moreno{-}Noguer.
\newblock Stochastic neural radiance fields: Quantifying uncertainty in implicit 3d representations.
\newblock \emph{CoRR}, abs/2109.02123, 2021.

\bibitem[Shen et~al.(2022)Shen, Agudo, Moreno-Noguer, and Ruiz]{CF-NeRF}
Jianxiong Shen, Antonio Agudo, Francesc Moreno-Noguer, and Adria Ruiz.
\newblock Conditional-flow nerf: Accurate 3d modelling with reliable uncertainty quantification.
\newblock In \emph{ECCV}, 2022.

\bibitem[Shen et~al.(2024)Shen, Ren, Ruiz, and Moreno-Noguer]{shen2024estimating}
Jianxiong Shen, Ruijie Ren, Adria Ruiz, and Francesc Moreno-Noguer.
\newblock Estimating 3d uncertainty field: Quantifying uncertainty for neural radiance fields.
\newblock In \emph{2024 IEEE International Conference on Robotics and Automation (ICRA)}, pages 2375--2381. IEEE, 2024.

\bibitem[Sucar et~al.(2021)Sucar, Liu, Ortiz, and Davison]{sucar2021imap}
Edgar Sucar, Shikun Liu, Joseph Ortiz, and Andrew~J Davison.
\newblock imap: Implicit mapping and positioning in real-time.
\newblock In \emph{ICCV}, pages 6229--6238, 2021.

\bibitem[Sun et~al.(2024)Sun, Choe, Loop, Ma, and Wang]{Sun2024SVR}
Cheng Sun, Jaesung Choe, Charles Loop, Wei-Chiu Ma, and Yu-Chiang~Frank Wang.
\newblock Sparse voxels rasterization: Real-time high-fidelity radiance field rendering.
\newblock \emph{ArXiv}, abs/2412.04459, 2024.

\bibitem[Sünderhauf et~al.(2023)Sünderhauf, Abou-Chakra, and Miller]{sünderhauf2022densityaware}
Niko Sünderhauf, Jad Abou-Chakra, and Dimity Miller.
\newblock Density-aware nerf ensembles: Quantifying predictive uncertainty in neural radiance fields.
\newblock In \emph{ICRA}, 2023.

\bibitem[Turkulainen et~al.(2025)Turkulainen, Ren, Melekhov, Seiskari, Rahtu, and Kannala]{turkulainen2024dnsplatter}
Matias Turkulainen, Xuqian Ren, Iaroslav Melekhov, Otto Seiskari, Esa Rahtu, and Juho Kannala.
\newblock Dn-splatter: Depth and normal priors for gaussian splatting and meshing.
\newblock In \emph{WACV}, 2025.

\bibitem[Wang et~al.(2021)Wang, Liu, Liu, Theobalt, Komura, and Wang]{neus}
Peng Wang, Lingjie Liu, Yuan Liu, Christian Theobalt, Taku Komura, and Wenping Wang.
\newblock Neus: Learning neural implicit surfaces by volume rendering for multi-view reconstruction.
\newblock In \emph{NeurIPS}, 2021.

\bibitem[Xue et~al.(2024)Xue, Dill, Mathur, Dellaert, Tsiotras, and Xu]{xue2024nvf}
Shangjie Xue, Jesse Dill, Pranay Mathur, Frank Dellaert, Panagiotis Tsiotras, and Danfei Xu.
\newblock Neural visibility field for uncertainty-driven active mapping.
\newblock In \emph{CVPR}, pages 18122--18132, 2024.

\bibitem[Yan et~al.(2023{\natexlab{a}})Yan, Liu, Quan, Chen, and Fu]{yan2023activeIO}
Dongyu Yan, Jianheng Liu, Fengyu Quan, Haoyao Chen, and Mengmeng Fu.
\newblock Active implicit object reconstruction using uncertainty-guided next-best-view optimization.
\newblock \emph{IEEE Robotics and Automation Letters}, pages 1--8, 2023{\natexlab{a}}.

\bibitem[Yan et~al.(2023{\natexlab{b}})Yan, Yang, and Zha]{yan2023active-neural-mapping}
Zike Yan, Haoxiang Yang, and Hongbin Zha.
\newblock Active neural mapping.
\newblock In \emph{ICCV}, 2023{\natexlab{b}}.

\bibitem[Yeshwanth et~al.(2023)Yeshwanth, Liu, Nie{\ss}ner, and Dai]{yeshwanth2023scannet++}
Chandan Yeshwanth, Yueh-Cheng Liu, Matthias Nie{\ss}ner, and Angela Dai.
\newblock Scannet++: A high-fidelity dataset of 3d indoor scenes.
\newblock In \emph{ICCV}, pages 12--22, 2023.

\bibitem[Y{\"u}cer et~al.(2016)Y{\"u}cer, Sorkine-Hornung, Wang, and Sorkine-Hornung]{yucer2016eLF}
Kaan Y{\"u}cer, Alexander Sorkine-Hornung, Oliver Wang, and Olga Sorkine-Hornung.
\newblock Efficient 3d object segmentation from densely sampled light fields with applications to 3d reconstruction.
\newblock \emph{ACM Transactions on Graphics (TOG)}, 35\penalty0 (3):\penalty0 1--15, 2016.

\bibitem[Zhan et~al.(2022)Zhan, Zheng, Xu, Reid, and Rezatofighi]{zhan2022activermap}
Huangying Zhan, Jiyang Zheng, Yi Xu, Ian Reid, and Hamid Rezatofighi.
\newblock Activermap: Radiance field for active mapping and planning, 2022.

\end{thebibliography}
}

\clearpage
\onecolumn
{\Large \bf \centering{
WarpRF: Multi-View Consistency for Training-Free Uncertainty Quantification} \\
and Applications in Radiance Fields \par }
\vspace{0.5cm}
{\Large \centering{
Supplementary Material \par }}
\vspace{1cm}
\def\thesection {\Alph{section}}
\setcounter{section}{0}
\renewcommand{\theHsection}{Supplement.\thesection}

This document provides additional material for our work ``WarpRF: Multi-View Consistency 
for Training-Free Uncertainty Quantification and Applications in Radiance Fields''. We provide results of uncertainty quantification for each scene in Sec. \ref{supp_sec:perscene_ause}, then report results of active view selection for each scene in Sec. \ref{supp_sec:perscene_active}.
\vspace{1cm}

\section{Per-Scene Uncertainty Quantifications}
\label{supp_sec:perscene_ause}
Uncertainty quantification experiments are performed on the ETH3D and ScanNet++ datasets. For the ETH3D dataset, we use all scenes of the high-resolution set, having 14 to 76 images. We train 3DGS/NeRF on every other image and use the rest to evaluate uncertainty.
For ScanNet++, we select the `27dd4da69e', `3864514494', `5eb31827b7', `8b5caf3398' `8d563fc2cc', and `b20a261fdf' scenes. Using the first 40 images of each scene, we train the radiance field on every other image and use the remaining ones for AUSE evaluation.

Tables \ref{supp_tab:ause_scannet} and \ref{supp_tab:ause_eth3d} present per-scene AUSE evaluation results for each scene in ScanNet++ and ETH3D, respectively.

\begin{table*}[b]
\centering
\begin{tabular}{l|cccccc}
\toprule
Method & 27dd4da69e & 3864514494 & 5eb31827b7 & 8b5caf3398 & 8d563fc2cc & b20a261fdf \\ \midrule
 NeRF + BayesRays &      0.456 &      0.407 &      0.489 &      0.498 &      0.368 &      0.412 \\
 NeRF + WarpRF (ours) &      0.487 &      0.365 &      0.485 &      0.418 &      0.345 &      0.437 \\ \midrule
 3DGS + Manifold &      0.569 &      0.474 &      0.514 &      0.466 &      0.505 &      0.533 \\
 3DGS + FisherRF &      0.379 &      0.358 &      0.293 &      0.297 &      0.414 &      0.386 \\
 3DGS + WarpRF (ours) &      0.364 &      0.300 &      0.291 &      0.331 &      0.308 &      0.430 \\ \bottomrule
\end{tabular}
\caption{\textbf{AUSE evaluation $\downarrow$ on ScanNet++ scenes.}}
\label{supp_tab:ause_scannet}
\end{table*}
\begin{table*}[]
\centering
\begin{adjustbox}{max width=\textwidth}
\begin{tabular}{l|ccccccccccccc}
\toprule
Method & courtyard & delivery area & electro & facade & kicker & meadow & office & pipes & playground & relief & relief 2 & terrace & terrains \\ \midrule
NeRF + BayesRays &      0.133 &      0.422 &      0.253 &      0.245 &      0.267 &      0.194 &      0.443 &      0.328 &      0.185 &      0.300 &      0.269 &      0.214 &      0.366  \\
 NeRF + WarpRF (ours) &      0.130 &      0.333 &      0.394 &      0.205 &      0.288 &      0.104 &      0.294 &      0.195 &      0.229 &      0.217 &      0.271 &      0.292 &      0.301  \\ \midrule
 3DGS + Manifold &      0.539 &      0.524 &      0.514 &      0.509 &      0.532 &      0.474 &      0.531 &      0.432 &      0.606 &      0.533 &      0.550 &      0.520 &      0.524  \\
 3DGS + FisherRF &      0.214 &      0.192 &      0.257 &      0.320 &      0.185 &      0.389 &      0.225 &      0.424 &      0.351 &      0.336 &      0.403 &      0.322 &      0.212  \\
 3DGS + WarpRF (ours) &      0.196 &      0.173 &      0.227 &      0.211 &      0.221 &      0.253 &      0.192 &      0.243 &      0.352 &      0.193 &      0.264 &      0.243 &      0.180  \\ \bottomrule
\end{tabular}
\end{adjustbox}
\caption{\textbf{AUSE evaluation $\downarrow$ on ETH3D scenes.}}
\label{supp_tab:ause_eth3d}
\end{table*}

\section{Per-Scene Active View Selection Results}
\label{supp_sec:perscene_active}
Tables \ref{supp_tab:mip_psnr} to \ref{supp_tab:mip_lpips} present the per-scene quantitative results of active view selection on the Mip-NeRF360 dataset, evaluated using PSNR, SSIM, and LPIPS metrics. Similarly, Tables \ref{supp_tab:nerf20_psnr} to \ref{supp_tab:nerf20_lpips} report results for NeRF Synthetic scenes with 20 views, while Tables \ref{supp_tab:nerf10_psnr} to \ref{supp_tab:nerf10_lpips} provide results for 10-view settings. 
Finally, the results for Tanks and Temples dataset are provided in Tab. \ref{supp_tab:tandt}.
\begin{table*}[]
\centering
\begin{tabular}{l|ccccccccc}
\toprule
Method       & bicycle  &  bonsai  & counter  & flowers  &  garden  & kitchen  &   room   &  stump   & treehill\\ \midrule
3DGS + Manifold  &   18.345 &  22.597 &   22.259 &   15.394 &   22.704 &   24.002 &   22.987 &   20.369 &   17.225 \\
3DGS + FisherRF  &   18.006 &  22.597 &   21.682 &   14.847 &   22.017 &   23.559 &   22.552 &   20.058 &   17.341 \\
3DGS + WarpRF (ours)      &   17.850 &  23.577 &   22.194 &   15.430 &   23.184 &   23.848 &   22.792 &   20.349 &   17.212 \\ \bottomrule
\end{tabular}
\caption{\textbf{PSNR $\uparrow$ results for active view selection on Mip-NeRF360}. }
\label{supp_tab:mip_psnr}
\end{table*}

\begin{table*}[]
\centering
\begin{tabular}{l|ccccccccc}
\toprule
Method       & bicycle  &  bonsai  & counter  & flowers  &  garden  & kitchen  &   room   &  stump   & treehill\\ \midrule
3DGS + Manifold  &    0.415 &   0.809 &    0.776 &    0.321 &    0.681 &    0.836 &    0.796 &    0.476 &    0.436 \\
3DGS + FisherRF  &    0.386 &   0.808 &    0.763 &    0.315 &    0.643 &    0.805 &    0.783 &    0.454 &    0.457 \\
3DGS + WarpRF (ours)      &    0.376 &   0.830 &    0.775 &    0.333 &    0.706 &    0.829 &    0.808 &    0.480 &    0.418 \\ \bottomrule
\end{tabular}
\caption{\textbf{SSIM $\uparrow$ results for active view selection on Mip-NeRF360}. }
\label{supp_tab:mip_ssim}
\end{table*}

\begin{table*}[]
\centering
\begin{tabular}{l|ccccccccc}
\toprule
Method       & bicycle  &  bonsai  & counter  & flowers  &  garden  & kitchen  &   room   &  stump   & treehill\\ \midrule
3DGS + Manifold  &    0.436 &   0.294 &    0.286 &    0.492 &    0.236 &    0.203 &    0.303 &    0.424 &    0.456 \\
3DGS + FisherRF  &    0.447 &   0.295 &    0.300 &    0.499 &    0.261 &    0.232 &    0.318 &    0.440 &    0.444 \\
3DGS + WarpRF (ours)      &    0.460 &   0.281 &    0.290 &    0.488 &    0.225 &    0.209 &    0.296 &    0.424 &    0.467 \\ \bottomrule
\end{tabular}
\caption{\textbf{LPIPS $\downarrow$ results for active view selection on Mip-NeRF360}. }
\label{supp_tab:mip_lpips}
\end{table*}

\begin{table*}[]
\centering
\begin{tabular}{l|cccccccc}
\toprule
Method       &  chair   &  drums   &  ficus   &  hotdog  &   lego   & materials &   mic    &   ship  \\ \midrule
3DGS + Manifold&   30.920 &  23.228 &   30.734 &   29.464 &   28.864 &   25.887 &   30.387 &   25.699 \\
3DGS + FisherRF  &   32.277 &  23.697 &   30.411 &   33.636 &   30.423 &   25.689 &   31.343 &   26.620 \\
3DGS + WarpRF (ours)      &   32.150 &  23.638 &   30.731 &   33.344 &   30.691 &   26.254 &   30.571 &   26.772 \\ \midrule
SVRaster + Random & 29.760 & 21.059 & 22.605 & 29.676 & 27.856 & 24.063 & 30.279 & 25.759 \\
SVRaster + Farthest & 30.394 & 21.462 & 22.736 & 30.613 & 27.722 & 22.849 & 30.620 & 26.040 \\
SVRaster + WarpRF (ours) & 30.519 & 22.156 & 24.034 & 31.454 & 28.536 & 25.283 & 30.973 & 26.325 \\ \bottomrule
\end{tabular}
\caption{\textbf{PSNR $\uparrow$ results for active view selection on NeRF Synthetic with 20 views}. }
\label{supp_tab:nerf20_psnr}
\end{table*}

\begin{table*}[]
\centering
\begin{tabular}{l|cccccccc}
\toprule
Method       &  chair   &  drums   &  ficus   &  hotdog  &   lego   & materials &   mic    &   ship  \\ \midrule
3DGS + Manifold&    0.965 &   0.919 &    0.972 &    0.951 &    0.931 &    0.926 &    0.976 &    0.841 \\
3DGS + FisherRF  &    0.974 &   0.927 &    0.972 &    0.969 &    0.951 &    0.923 &    0.980 &    0.844 \\
3DGS + WarpRF (ours)      &    0.975 &   0.926 &    0.972 &    0.971 &    0.955 &    0.931 &    0.977 &    0.846 \\ \midrule
SVRaster + Random & 0.949 & 0.871 & 0.902 & 0.933 & 0.921 & 0.882 & 0.973 & 0.810 \\ 
SVRaster + Farthest & 0.959 & 0.886 & 0.908 & 0.947 & 0.930 & 0.872 & 0.976 & 0.816 \\
SVRaster + WarpRF (ours) & 0.959 & 0.892 & 0.926 & 0.950 & 0.935 & 0.905 & 0.976 & 0.819 \\ \bottomrule
\end{tabular}
\caption{\textbf{SSIM $\uparrow$ results for active view selection on NeRF Synthetic with 20 views}. }
\label{supp_tab:nerf20_ssim}
\end{table*}

\begin{table*}[]
\centering
\begin{tabular}{l|cccccccc}
\toprule
Method       &  chair   &  drums   &  ficus   &  hotdog  &   lego   & materials &   mic    &   ship  \\ \midrule
3DGS + Manifold&    0.029 &   0.059 &    0.022 &    0.057 &    0.060 &    0.057 &    0.017 &    0.154 \\
3DGS + FisherRF  &    0.021 &   0.054 &    0.023 &    0.038 &    0.043 &    0.061 &    0.014 &    0.145 \\
3DGS + WarpRF (ours)      &    0.021 &   0.054 &    0.022 &    0.033 &    0.041 &    0.055 &    0.016 &    0.143 \\ \midrule
SVRaster + Random  & 0.052 & 0.124 & 0.082 & 0.085 & 0.069 & 0.100 & 0.023 & 0.167 \\
SVRaster + Farthest & 0.041 & 0.106 & 0.071 & 0.068 & 0.062 & 0.107 & 0.021 & 0.157 \\
SVRaster + WarpRF (ours) & 0.042 & 0.101 & 0.059 & 0.065 & 0.057 & 0.085 & 0.020 & 0.155 \\ \bottomrule
\end{tabular}
\caption{\textbf{LPIPS $\downarrow$ results for active view selection on NeRF Synthetic with 20 views}. }
\label{supp_tab:nerf20_lpips}
\end{table*}
\begin{table*}[]
\centering
\begin{tabular}{l|cccccccc}
\toprule
Method       &  chair   &  drums   &  ficus   &  hotdog  &   lego   & materials &   mic    &   ship  \\ \midrule
3DGS + Manifold&   27.006 &  21.220 &   26.152 &   25.240 &   24.330 &   21.898 &   26.825 &   20.600 \\
3DGS + FisherRF  &   26.766 &  20.975 &   25.817 &   29.456 &   25.162 &   21.474 &   28.427 &   21.468 \\
3DGS + WarpRF (ours)      &   27.875 &  21.269 &   26.859 &   27.971 &   24.842 &   21.682 &   28.446 &   20.714 \\ \bottomrule
\end{tabular}
\caption{\textbf{PSNR $\uparrow$ results for active view selection on NeRF Synthetic with 10 views}. }
\label{supp_tab:nerf10_psnr}
\end{table*}

\begin{table*}[]
\centering
\begin{tabular}{l|cccccccc}
\toprule
Method       &  chair   &  drums   &  ficus   &  hotdog  &   lego   & materials &   mic    &   ship  \\ \midrule
3DGS + Manifold&    0.929 &   0.887 &    0.943 &    0.916 &    0.873 &    0.879 &    0.957 &    0.746 \\
3DGS + FisherRF  &    0.928 &   0.886 &    0.941 &    0.940 &    0.882 &    0.870 &    0.966 &    0.757 \\
3DGS + WarpRF (ours)      &    0.940 &   0.891 &    0.949 &    0.939 &    0.891 &    0.871 &    0.966 &    0.753 \\ \bottomrule
\end{tabular}
\caption{\textbf{SSIM $\uparrow$ results for active view selection on NeRF Synthetic with 10 views}. }
\label{supp_tab:nerf10_ssim}
\end{table*}

\begin{table*}[]
\centering
\begin{tabular}{l|cccccccc}
\toprule
Method       &  chair   &  drums   &  ficus   &  hotdog  &   lego   & materials &   mic    &   ship  \\ \midrule
3DGS + Manifold&    0.057 &   0.085 &    0.047 &    0.098 &    0.108 &    0.105 &    0.035 &    0.217 \\
3DGS + FisherRF  &    0.056 &   0.090 &    0.048 &    0.065 &    0.099 &    0.111 &    0.025 &    0.199 \\
3DGS + WarpRF (ours)      &    0.048 &   0.084 &    0.043 &    0.066 &    0.096 &    0.108 &    0.025 &    0.204 \\ \bottomrule
\end{tabular}
\caption{\textbf{LPIPS $\downarrow$ results for active view selection on NeRF Synthetic with 10 views}. }
\label{supp_tab:nerf10_lpips}
\end{table*}

\begin{table*}[]
\centering
\begin{tabular}{l|cc|cc|cc}
\toprule
                    & \multicolumn{2}{c|}{PSNR $\uparrow$} & \multicolumn{2}{c|}{SSIM $\uparrow$} & \multicolumn{2}{c}{LPIPS $\downarrow$} \\
Method              & Train       & Truck       & Train       & Truck       & Train       & Truck       \\ \hline
SVRaster + Random   & 13.449      & 19.818      & 0.437       & 0.710       & 0.477       & 0.239       \\
SVRaster + Farthest & 14.594      & 20.353      & 0.482       & 0.741       & 0.434       & 0.218       \\
SVRaster + WarpRF (ours)     & 14.803      & 20.746      & 0.493       & 0.756       & 0.432       & 0.204       \\ \hline
\end{tabular}
\caption{\textbf{Active view selection results on Tanks and Temples dataset}. }
\label{supp_tab:tandt}
\end{table*}   

\end{document}